\documentclass[10pt]{article}
\usepackage[a4paper, top=1.5cm, bottom=2.5cm, left=2.5cm, right=2.5cm]{geometry}
\usepackage{amsmath}
\usepackage{accents}  
\usepackage{amsfonts}
\usepackage{amsthm}
\usepackage{bm}
\usepackage{cite}
\usepackage{hyperref}
\usepackage{ifthen}
\usepackage{mathtools}
\usepackage{svg}
\usepackage{tikz}

\newboolean{showauthorinfo}
\setboolean{showauthorinfo}{true}
\newboolean{showdate}
\setboolean{showdate}{true}

\newtheorem{assumption}{Assumption}

\newtheorem{remark}{Remark}
\theoremstyle{definition}

\theoremstyle{property}
\newtheorem{property}{Property}

\newcommand*\circled[1]{\textcircled{\raisebox{-0.9pt}{#1}}}

\newcommand{\norm}[1]{\lVert#1\rVert}
\newcommand{\lrnorm}[1]{\left\lVert#1\right\rVert}
\newcommand{\ubar}[1]{\underaccent{\bar}{#1}}

\def\p{{\bm{p}}}
\def\tp{{\tilde{\bm{p}}}}
\def\hp{{\hat{\bm{p}}}}

\def\x{{\bm{x}}}
\def\dx{{\dot{\bm{x}}}}
\def\tx{{\tilde{\bm{x}}}}
\def\hx{{\hat{\bm{x}}}}
\def\dhx{{\dot{\hat{\bm{x}}}}}

\def\bu{{\bm{u}}}
\def\tu{{\tilde{\bm{u}}}}
\def\y{{\bm{y}}}

\def\z{{\bm{z}}}
\def\dz{{\dot{\bm{z}}}}
\def\tz{{\tilde{\bm{z}}}}
\def\dtz{{\dot{\tilde{\bm{z}}}}}
\def\r{{\bm{r}}}
\def\xr{{\bm{x}^\mathrm{r}}}
\def\dxr{{\dot{\bm{x}}^\mathrm{r}}}
\def\ur{{\bm{u}^\mathrm{r}}}
\def\bv{{\bm{v}}}
\def\tv{{\tilde{\bm{v}}}}
\def\ds{{\dot{s}}}
\def\dts{{\dot{\tilde{s}}}}
\def\bs{{\bar{s}}}
\def\w{{\bm{w}}}
\def\bw{{\bar{w}}}
\def\bwo{{\bar{w}^\mathrm{o}}}
\def\bmeta{{\bm{\eta}}}

\def\rec{^\mathrm{rec}}
\def\ubw{{\ubar{\bm{w}}}}
\def\bbw{{\bar{\bm{w}}}}
\def\b{^\mathrm{b}}
\def\wb{{\bm{w}\b}}
\def\ubeta{{\ubar{\bmeta}}}
\def\bbeta{{\bar{\bmeta}}}

\def\tr{^\mathrm{r}}
\def\bdelta{{\bm{\delta}}}
\def\dbdelta{{\dot{\bdelta}}}
\def\tbdelta{{\tilde{\bdelta}}}
\def\beps{{\bm{\epsilon}}}
\def\dbeps{{\dot{\beps}}}
\def\bmzeta{{\bm{\zeta}}}

\def\fwarg{{f^\mathrm{w}(\x,\bu,\w)}}

\def\Ts{{T^\mathrm{s}}}
\def\0t{{0|t}}
\def\tt{{\tau|t}}
\def\tpt{{t+\tau}}
\def\ct{{\cdot|t}}

\def\Tst{{\Ts|t}}
\def\Tt{{T|t}}
\def\tpT{{t+T}}
\def\Tpt{{T+\tau}}
\def\tpTst{{t+\Ts+\tau}}
\def\tpTt{{t+T+\tau}}
\def\tpTsT{{t+\Ts+T}}
\def\tTs{{t+\Ts}}
\def\Tstau{{\Ts+\tau}}
\def\TsT{{\Ts+T}}
\def\ttT{{\tau|\tTs}}
\def\tTt{{\Tstau|t}}
\def\TTt{{\TsT|t}}
\def\TtT{{T|\tTs}}

\def\gj{{g_j}}
\def\gjs{{g_{j}^\mathrm{s}}}
\def\gjrs{{g_{j}^\mathrm{r,s}}}
\def\Ljs{{L_{j}^\mathrm{s}}}
\def\Ljrs{{L_{j}^\mathrm{r,s}}}
\def\ljs{{l_{j}^\mathrm{s}}}
\def\ljrs{{l_{j}^\mathrm{r,s}}}
\def\cj{{c_j}}
\def\cjs{{c_{j}^\mathrm{s}}}
\def\cjcs{{c_{j}^\mathrm{c,s}}}
\def\cjso{{c_{j}^\mathrm{s,o}}}
\def\go{{g^\mathrm{o}}}
\def\Lo{{L^\mathrm{o}}}
\def\lo{{l^\mathrm{o}}}
\def\gjo{{g_{j}^\mathrm{o}}}
\def\gjtto{{g_{j,\tt}^\mathrm{o}}}

\def\gjttTo{{g_{j,\ttT}^\mathrm{o}}}
\def\gro{{g^\mathrm{r,o}}}
\def\gjro{{g_{j}^\mathrm{r,o}}}
\def\gjttro{{g_{j,\tt}^\mathrm{r,o}}}
\def\Ljo{{L_j^\mathrm{o}}}
\def\ljo{{l_j^\mathrm{o}}}
\def\Ljtto{{L_{j,\tt}^\mathrm{o}}}
\def\ljtto{{l_{j,\tt}^\mathrm{o}}}
\def\co{{c^\mathrm{o}}}
\def\cco{{c^\mathrm{c,o}}}

\def\nx{{n^\mathrm{x}}}
\def\np{{n^\mathrm{p}}}
\def\nuu{{n^\mathrm{u}}}
\def\nxtu{{\nx\times\nuu}}
\def\nutx{{\nuu\times\nx}}
\def\ny{{n^\mathrm{y}}}
\def\nw{{n^\mathrm{w}}}
\def\neta{{n^\mathrm{\eta}}}

\def\ns{{n^\mathrm{s}}}
\def\no{{n^\mathrm{o}}}

\def\Inx{{I^{\nx}}}
\def\Inp{{I^{\np}}}
\def\Inu{{I^{\nuu}}}
\def\R{{\mathbb{R}}}
\def\Rg0{{\mathbb{R}_{> 0}}}
\def\Rgeq0{{\mathbb{R}_{\geq 0}}}

\def\Rx{{\mathbb{R}^{\nx}}}
\def\Ru{{\mathbb{R}^{\nuu}}}
\def\Rux{{\mathbb{R}^{\nuu+\nx}}}

\def\Rxtx{{\mathbb{R}^{\nx\times\nx}}}
\def\Rutx{{\mathbb{R}^{\nuu\times\nx}}}
\def\Rp{{\mathbb{R}^{\np}}}

\def\Ry{{\mathbb{R}^{\ny}}}
\def\N{{\mathbb{N}}}
\def\Nn{{\mathbb{N}_{[1,n]}}}
\def\Nw{{\mathbb{N}_{[1,\nw]}}}
\def\Rw{{\mathbb{R}^{\nw}}}
\def\Neta{{\mathbb{N}_{[1,\neta]}}}
\def\Reta{{\mathbb{R}^{\neta}}}
\def\RC{{\mathbb{R}^{\ny\times\nx}}}
\def\RF{{\mathbb{R}^{\ny\times\neta}}}

\def\RL{{\mathbb{R}^{\nx\times\ny}}}
\def\Ns{{\mathbb{N}_{[1,\ns]}}}
\def\No{{\mathbb{N}_{[1,\no]}}}

\def\L1{{\mathcal{L}_1}}
\def\Linf{{\mathcal{L}_\infty}}
\def\F{{\mathcal{F}}}
\def\J{{\mathcal{J}}}
\def\K{{\mathcal{K}}}
\def\U{{\mathcal{U}}}
\def\bU{{\bar{\U}}}
\def\W{{\mathcal{W}}}
\def\bW{{\bar{\W}}}

\def\H{{\mathcal{H}}}
\def\bH{{\bar{H}}}
\def\bHone{{\bar{H}_1}}
\def\X{{\mathcal{X}}}
\def\Xf{{\mathcal{X}^\mathrm{f}}}
\def\bX{{\bar{\X}}}

\def\Z{{\mathcal{Z}}}
\def\bZ{{\bar{\mathcal{Z}}}}

\def\XtX{{\X^2}}
\def\XtZ{{\X \times \Z}}
\def\XtbX{{\X \times \bX}}
\def\XtbZ{{\X \times \bZ}}

\def\xu{{\x,\bu}}
\def\ux{{\bu,\x}}
\def\xz{{\x,\z}}
\def\hxz{{\hx,\z}}

\def\zv{{\z,\bv}}
\def\zz{{\z,\z}}
\def\xzv{{\x,\z,\bv}}

\def\hxzv{{\hx,\z,\bv}}
\def\kappaxzv{{\kappa(\xzv)}}
\def\kappahxzv{{\kappa(\hxzv)}}
\def\kappadef{{\XtZ \rightarrow \Ru}}

\def\kappaf{{\kappa^\mathrm{f}}}
\def\kappafxr{{\kappa^\mathrm{f}(\x,\r)}}
\def\kappafhxr{{\kappa^\mathrm{f}(\hx,\r)}}
\def\kappad{{\kappa^\delta}}
\def\kappadxz{{\kappad(\xz)}}
\def\kappadhxz{{\kappad(\hxz)}}
\def\kappaddef{{\XtX \rightarrow \Ru}}
\def\bkappa{{\bar{\kappa}^\delta}}

\def\dmax{{d_\mathrm{max}}}
\def\Xmin{{X_{\mathrm{min}}}}

\def\classKinf{{\mathrm{class\text{-}}\K_\infty}}

\def\Qeps{{Q^\epsilon}}

\def\Jf{{\J^\mathrm{f}}}
\def\Js{{\J^\mathrm{s}}}
\def\Ts{{T^\mathrm{s}}}
\def\tops{^\mathrm{s}}
\def\topf{^\mathrm{f}}
\def\Ljsx{{L^\mathrm{\Js,x}}}
\def\Ljsu{{L^\mathrm{\Js,u}}}
\def\Ljsxu{{L^\mathrm{\Js,xu}}}
\def\Ljf{{L^\mathrm{\Jf}}}
\def\Ljfx{{L^\mathrm{\Jf,x}}}

\def\balphasf{{\bar{\alpha}^\mathrm{s,f}}}
\def\balphasfone{{\bar{\alpha}^\mathrm{s,f,1}}}

\def\ts{{\tilde{s}}}
\def\cx{{\bm{c}^\mathrm{x}}}
\def\cxs{{\bm{c}^\mathrm{x}(s)}}
\def\cxss{{\frac{\partial \cxs}{\partial s}}}

\def\g{{\bm{\gamma}}}

\def\gx{{\g^\mathrm{x}}}
\def\gxs{{\g^\mathrm{x}(s)}}

\def\gxts{{\g^\mathrm{x}(\ts)}}
\def\gxss{{\frac{\partial \gxs}{\partial s}}}
\def\gxtss{{\frac{\partial \gxts}{\partial \ts}}}

\def\dgx{{\dot{\g}^\mathrm{x}}}
\def\dgxs{{\dot{\g}^\mathrm{x}(s)}}

\def\gxssshort{{\g_s^\mathrm{x}}}
\def\gxssshorttop{{\g_s^{\mathrm{x}^\top}}}
\def\dgxssshort{{\dot{\g}_s^\mathrm{x}}}
\def\dgxssshorttop{{\dot{\g}_s^{\mathrm{x}^\top}}}

\def\gu{{\g^\mathrm{u}}}
\def\gus{{\g^\mathrm{u}(s)}}

\def\Vd{{V^\delta}}
\def\Vdxz{{V^\delta(\xz)}}
\def\dVdxz{{\frac{d}{dt}V^\delta(\xz)}}
\def\sVdxz{{\sqrt{\Vdxz}}}
\def\Vdhxz{{V^\delta(\hxz)}}
\def\dVdhxz{{\frac{d}{dt}V^\delta(\hxz)}}
\def\sVdhxz{{\sqrt{\Vdhxz}}}
\def\Vdxhx{{V^\delta(\x,\hx)}}
\def\dVdxhx{{\frac{d}{dt}V^\delta(\x,\hx)}}
\def\sVdxhx{{\sqrt{\Vdxhx}}}
\def\Vddef{{\XtX \rightarrow \Rgeq0}}
\def\cdl{{c^{\delta\mathrm{,l}}}}
\def\cdu{{c^{\delta\mathrm{,u}}}}
\def\Pd{{P^\delta}}
\def\Kd{{K^\delta}}
\def\Xd{{X^\delta}}
\def\Yd{{Y^\delta}}
\def\dPd{{\dot{P}^\delta}}
\def\Aarg{{A(\bmzeta)}}
\def\Barg{{B(\bmzeta)}}
\def\Pdarg{{P^\delta(\x)}}
\def\Kdarg{{K^\delta(\x)}}

\def\Ydarg{{Y^\delta(\x)}}

\def\Pdcs{{P^\delta(\cxs)}}
\def\Pdg{{P^\delta(\gx)}}

\def\bPdg{{\bar{P}^\delta(\gx)}}
\def\dPdg{{\dot{P}^\delta(\gx)}}

\def\Ydg{{Y^\delta(\gx)}}

\def\Kdg{{K^\delta(\gx)}}
\def\Kdgs{{K^\delta(\gxs)}}

\def\Xdg{{X^\delta(\gx)}}

\def\dXdg{{\dot{X}^\delta(\gx)}}

\def\Ag{{A(\gx,\gu)}}
\def\Bg{{B(\gx,\gu)}}
\def\Ags{{A(\gxs,\gus)}}
\def\Bgs{{B(\gxs,\gus)}}

\def\Acl{{A^\mathrm{cl}}}
\def\Aclarg{{\Acl(\gx,\gu)}}
\def\Aclargs{{\Acl(\gxs,\gus)}}
\def\Acltop{{A^{\mathrm{cl}^\top}}}

\def\oplim{{\operatorname{lim}}}
\def\opmin{{\operatorname{min}}}
\def\opmax{{\operatorname{max}}}
\def\opsqrt{{\operatorname{sqrt}}}
\def\opvert{{\operatorname{vert}}}

\def\lambdamin{{\lambda_\mathrm{min}}}
\def\lambdamax{{\lambda_\mathrm{max}}}
\def\lambdadelta{{\lambda^\delta}}
\def\lambdaeps{{\lambda^\epsilon}}
\def\lambdadeltaeps{{\lambda^{\delta,\epsilon}}}

\begin{document}

\begin{center}
{\large Technical Report}

\vspace{1em}

{\LARGE \bfseries A Step-by-step Guide on Nonlinear Model Predictive Control for Safe Mobile Robot Navigation}

\vspace{1em}

\ifthenelse{\boolean{showauthorinfo}}{
    {\large Dennis Benders\footnotemark[1], Laura Ferranti\footnotemark[1], and Johannes K\"{o}hler\footnotemark[2]}
}{}

\vspace{1em}

\ifthenelse{\boolean{showdate}}{
    {August 9, 2025}
}{}

\vspace{1em}
\end{center}

\ifthenelse{\boolean{showauthorinfo}}{
    \footnotetext[1]{Dennis Benders and Laura Ferranti are with the department of Cognitive Robotics, Delft University of Technology, 2628 CD Delft, The Netherlands (email: \{d.benders, l.ferranti\}@tudelft.nl). Laura Ferranti received support from the Dutch Science Foundation NWO-TTW Foundation within the Veni project HARMONIA (nr. 18165).}
    \footnotetext[2]{Johannes K\"{o}hler is with the Institute for Dynamic Systems and Control, ETH Z\"{u}rich, CH-8092, Switzerland (email: jkoehle@ethz.ch). Johannes K\"ohler was supported by the Swiss National Science Foundation under NCCR Automation (grant agreement 51NF40 180545).}
}{}


\section{Introduction}
Designing a model predictive control (MPC) scheme that enables a mobile robot to safely navigate through an obstacle-filled environment is a complicated yet essential task in robotics. In this technical report, safety refers to ensuring that the robot respects state and input constraints while avoiding collisions with obstacles despite the presence of disturbances and measurement noise. This report offers a step-by-step approach to implementing nonlinear model predictive Control (NMPC) schemes addressing these safety requirements. Numerous books and survey papers provide comprehensive overviews of linear MPC (LMPC) \cite{bemporad2007robust,kouvaritakis2016model}, NMPC \cite{rawlings2017model,allgower2004nonlinear,mayne2014model,grune2017nonlinear,saltik2018outlook}, and their applications in various domains, including robotics \cite{nascimento2018nonholonomic,nguyen2021model,shi2021advanced,wei2022mpc}. This report does not aim to replicate those exhaustive reviews. Instead, it focuses specifically on NMPC as a foundation for safe mobile robot navigation. The goal is to provide a practical and accessible path from theoretical concepts to mathematical proofs and implementation, emphasizing safety and performance guarantees. It is intended for researchers, robotics engineers, and practitioners seeking to bridge the gap between theoretical NMPC formulations and real-world robotic applications.

\subsection{Overview}
To implement an NMPC (hereafter referred to simply as MPC) scheme on a mobile robot, several key components must be addressed:

\begin{itemize}
    \item \textbf{System modeling}: define the robot's dynamics, its environment, and task objectives (Section~\ref{sec:robot_env}).
    \item \textbf{Reference tracking}: formulate the MPC problem for trajectory tracking (Section~\ref{sec:tmpc}). This formulation is related to the ones described in \cite{faulwasser2013optimization,kohler2020nonlinear,benders2025embedded}.
    \item \textbf{Robustness to disturbances}: modify the MPC formulation to ensure safety under disturbances (Section~\ref{sec:rmpc}). This formulation is related to the ones described in \cite{kohler2018novel,soloperto2019collision,kohler2020computationally,zhao2022tube,sasfi2023robust}.
    \item \textbf{Robustness to model uncertainty}: develop a robust output-feedback MPC scheme to handle model uncertainty consisting of both disturbances and measurement noise (Section~\ref{sec:rompc}). This formulation is related to the ones described in \cite{kogel2017robust,kohler2019simple,kohler2021robust,chou2022safe,brown2025robust}.
\end{itemize}

Each MPC scheme is introduced in a structured manner. The formulation is presented first, followed by a discussion of its desirable properties. Detailed mathematical proofs are then provided to demonstrate how these properties are satisfied. These proofs rely on specific assumptions, which are shown to hold under the proposed designs.

\subsection{Notation}
Throughout this guide, vectors are denoted in bold and matrices in capital letters. The sets $\R_{[a,b)}$ and $\N_{[a,b)}$ represent real and natural numbers, respectively, ranging from $a$ (inclusive) to $b$ (exclusive). The interior of a set $A$ is indicated with $\operatorname{int}(A)$. The Euclidean norm for vectors and the induced norm for matrices are denoted by $\norm{\cdot}$, while the quadratic norm with respect to a positive (semi-)definite matrix $Q$ is written as $\norm{\x}_Q = \x^\top Q \x$, where $Q \succ 0$ or $Q \succeq 0$. Subscripts indicate time steps, indices, or both, and superscripts denote the quantity to which a variable belongs. For example, $\gjttro(\p_{\tt})$ refers to the $j$th obstacle avoidance constraint for the reference trajectory, expressed in terms of the position states $\p_{\tt}$ at prediction stage $\tau$ of the MPC problem solved at time $t$. The predicted state-input trajectory at time $t$ is denoted by $(\x_{\tt}, \bu_{\tt})$, with the optimal solution written as $(\x_{\tt}^*, \bu_{\tt}^*)$ and a candidate solution as $(\tx_{\tt}, \tu_{\tt})$.

\section{Mobile robot, environment, and task description}\label{sec:robot_env}
The nominal robot system dynamics are given by
\begin{equation}\label{eq:xdot_nom}
    \dx = f(\x, \bu),
\end{equation}
with state $\x \in \Rx$, control input $\bu \in \Ru$, and Lipschitz continuous nonlinear function $f(\xu)$. The system is subject to polytopic input and state constraints, also called system constraints, described as
\begin{equation}\label{eq:con_sys}
    \Z \coloneqq \U \times \X = \left\{(\bu,\x) \in \Rux | \gjs(\x,\bu) \leq 0, \ j \in \Ns\right\},
\end{equation}
with $\gjs(\x,\bu) = \Ljs \begin{bmatrix}\bu\\\x\end{bmatrix} - \ljs, \Ljs \in \mathbb{R}^{1 \times \ns}, \ljs \in \R, j \in \Ns$. Note that each input and state constraint has a lower and upper bound. Therefore, there exist $\ns = 2(n^\mathrm{u}+n^\mathrm{x})$ system constraints in total.

The mobile robot operates in an environment with obstacles. To ensure the optimization of collision-free trajectories by the MPC, we need to define the obstacle avoidance constraints. In this guide, the obstacle avoidance constraints are assumed to be constructed as polytopic inner approximations of the possibly non-convex collision-free space. Since the trajectory of the robot progresses over time, these constraints should be defined as a function of time $t$ and prediction stage $\tau$ in the MPC horizon:
\begin{equation}\label{eq:con_obs}
    \F_{\tt} \coloneqq \left\{\p_{\tt} \in \Rp\ \middle|\ \gjtto(\p_{\tt}) \leq 0,\ j \in \No\ \tau \in [0,T],\ t \geq 0\right\},
\end{equation}
with position $\p = M\x \in \Rp$, $\gjtto(\p_{\tt}) = \Ljtto \p_{\tt} - \ljtto, \Ljtto \in \mathbb{R}^{1 \times n^\mathrm{p}}$, $\ljtto \in \mathbb{R},$ $j \in \No$. Without loss of generality, we set $\norm{\Ljtto} = 1, j \in \No$ by scaling the constraints. For notational convenience, let's denote the obstacle avoidance constraints by $\go(\p) = \Lo \p - \lo$ if we write about general constraints properties and $\gjo(\p) = \Ljo \p - \ljo,\ j\in\No$ if we write constraints properties that need to hold for all constraints explicitly.

\begin{remark}
    Obstacle avoidance constraints \eqref{eq:con_obs} form convex polytopic sets that change over the prediction time $\tau$ at time $t$. A method to compute a single convex polytope is described in \cite{liu2017planning}. Based on this, \cite{benders2025embedded} leverages a sequence of polytopic obstacle avoidance constraints, which are used in this work. A similar way to represent the obstacle avoidance constraints is recently proposed in \cite{arrizabalaga2024differentiable}, in which the constraints are continuously parameterized and differentiable off-centered ellipsoids based on polynomial path segments. This allows them to be more flexible at the cost of increased computation time compared to \eqref{eq:con_obs}.
\end{remark}

The task of the mobile robot is to track a reference trajectory that satisfies the following property:
\begin{property}\label{property:r}
    Reference trajectory
    \begin{equation}\label{eq:r}
        \r = [\ur^\top \xr^\top]^\top:
    \end{equation}
    \begin{enumerate}
        \item is dynamically feasible:
        \begin{equation}
            \dxr_{\tpt} = f(\x_{\tpt}\tr,\bu_{\tpt}\tr);
        \end{equation}
        \item satisfies a tightened set of the system constraints:
        \begin{equation}
            \r_{\tpt} \in \bZ,
        \end{equation}
        where $\bZ$ is defined as
        \begin{equation}\label{eq:r_sys}
            \bZ \coloneqq \bU \times \bX = \left\{(\ux) \in \Rux\ \middle|\ \gjrs(\xu) \leq 0,\ j \in \Ns\right\} \subseteq \operatorname{int}(\Z),
        \end{equation}
        with later introduced functions $\gjrs(\xu),\ j \in \Ns$;
        \item satisfies a tightened set of the obstacle avoidance constraints:
        \begin{equation}
            M \x_{\tpt}\tr \in \bar{\F}_{\tt},
        \end{equation}
        where $\bar{\F}_{\tt}$ is defined as
        \begin{equation}\label{eq:r_obs}
            \bar{\F}_{\tt} \coloneqq \left\{\p_{\tt} \in \Rp\ \middle|\ \gjttro(\p_{\tt}) \leq 0,\ j \in \No\right\} \subseteq \operatorname{int}(\F_{\tt}),\ \tau \in [0,T],\ t \geq 0,
        \end{equation}
        with later introduced functions $\gjttro(\p_{\tt}),\ j \in \No$;
    \end{enumerate}
\end{property}

The tightening of the system and obstacle avoidance constraints is included to ensure that the original constraints are satisfied as long as the real robot tracks the reference trajectory with a bounded error.

In general, generating references trajectories satisfying Property~\ref{property:r} is non-trivial. Therefore, a method to generate such reference trajectories is proposed in \cite{benders2025embedded}.

\section{MPC for trajectory tracking}\label{sec:tmpc}
This section aims to introduce the MPC scheme that will be used to track the reference trajectory defined in Property~\ref{property:r}, the so-called tracking MPC (TMPC). To explain TMPC's properties, Section~\ref{sec:tmpc_formulation} directly states its formulation. Then, we will discuss the elements in the formulation and the properties it should satisfy to ensure that the mobile robot safely navigates through the environment. This is the topic of Section~\ref{sec:tmpc_properties}. After discussing the TMPC design, we will provide some concluding remarks on the TMPC formulation in Section~\ref{sec:tmpc_remarks}.

\subsection{TMPC formulation}\label{sec:tmpc_formulation}
Consider the following MPC formulation for tracking a reference trajectory defined in Property~\ref{property:r}:
\begin{subequations}\label{eq:tmpc}
    \begin{alignat}{3}
        \J_t^*(\x_t,\r_t) = \underset{\substack{\x_{\ct},\bu_{\ct}}}{\opmin}\ \ &\mathrlap{\Jf(\x_{\Tt},\x_{\tpT}\tr) + \int_{0}^{T} \Js(\x_{\tt},\bu_{\tt},\r_{\tpt})\ d\tau,}&&&&\hspace{120pt} \label{eq:tmpc_obj}\\
        \operatorname{s.t.}\ &\x_{\0t} = \x_t,&&&& \label{eq:tmpc_x0}\\
        &\dx_{\tt} = f(\x_{\tt},\bu_{\tt}),&&&&\ \tau \in [0,T], \label{eq:tmpc_xdot}\\
        &\gjs(\x_{\tt},\bu_{\tt}) \leq 0,&&\ j \in \Ns,&&\ \tau \in [0,T], \label{eq:tmpc_sys}\\
        &\gjtto(\p_{\tt}) \leq 0,&&\ j \in \No,&&\ \tau \in [0,T], \label{eq:tmpc_obs}\\
        &\x_{\Tt} \in \Xf(\x_{\tpT}\tr), \label{eq:tmpc_term}
    \end{alignat}
\end{subequations}
with stage cost $\Js(\x,\bu,\r)=\norm{\x-\xr}_Q^2+\norm{\bu-\ur}_R^2, Q \succ 0, R \succ 0$ and terminal cost $\Jf(\x,\xr)=\norm{\x-\xr}_P^2, P \succ 0$. In this formulation, $Q$ and $R$ are tuning matrices, and $P$ should be suitably computed to ensure stability as given in Section~\ref{sec:tmpc_proof_tracking}. The optimal solution to \eqref{eq:tmpc} is denoted by $\x_{\ct}^*$ and $\bu_{\ct}^*$, with $\x_{\ct}^*$ being the optimal state trajectory and $\bu_{\ct}^*$ being the optimal control input trajectory.

TMPC problem \eqref{eq:tmpc} is applied in a receding-horizon fashion, meaning that it is solved every $\Ts$ seconds and the optimal control input $\bu_{\ct}^*$ is applied during each interval with length $\Ts$. According to this principle, the control law for the closed-loop system is defined as follows for $t = i\Ts, i \in \N$:
\begin{equation}\label{eq:tmpc_cl}
    \bu_{t+\tau} = \bu_{\tt}^*,\ \tau \in [0,\Ts].
\end{equation}

\subsection{TMPC properties}\label{sec:tmpc_properties}
The advantage of running a TMPC on a mobile robot is that if \eqref{eq:tmpc} can be solved, it is the optimal solution to the problem. The optimal solution is a feasible solution to \eqref{eq:tmpc}, i.e., one that satisfies all constraints \eqref{eq:tmpc_x0}-\eqref{eq:tmpc_term}. Furthermore, it minimizes objective \eqref{eq:tmpc_obj}. If the problem is suitably formulated, the robot will efficiently navigate through the environment by accurately tracking the reference trajectory.

Note that MPC is a trajectory optimization approach. Theoretically, one would like to optimize a suitable trajectory for all future times. However, this is not a tractable problem to solve. Therefore, the MPC horizon is finite. To ensure that using a finite horizon does not lead to unavoidable infeasible problems in the future, so-called terminal ingredients are designed \cite{kouvaritakis2016model}. The terminal ingredients include the terminal control law $\kappaf$, terminal cost $\Jf$, and terminal set $\Xf$. The idea behind the terminal ingredients is that one could safely execute the open-loop MPC strategy appended with $\kappaf$. This might not be the optimal behavior since \eqref{eq:tmpc} is not solved every $\Ts$ seconds, but it would be feasible. If designed suitably, the execution of $\kappaf$ after $\bu_{[0,T]|t}^*$ renders $\Jf$ an upper bound for the `infinite-horizon' cost that is not accounted for in the integral term of \eqref{eq:tmpc_obj}. Furthermore, if $\Xf$ is suitably designed, this set is positive invariant under $\kappaf$. This means that the system is guaranteed to remain within this set. As one can imagine, the terminal ingredients are crucial for the safety guarantees of an MPC scheme. Therefore, they are central to the MPC design.

The discussion on terminal ingredients above holds for classical stabilization MPC types that bring the system to an equilibrium state. Recently, this theory has been extended to tracking dynamic reference trajectories, such as the one defined in Property~\ref{property:r}, rendering them useful for mobile robot deployment. As a result, the terminal ingredients all have \eqref{eq:r} as input argument. In particular, their commonly used expressions are
\begin{equation}\label{eq:tmpc_kappaf}
    \kappafxr=\ur+K(\x-\xr),
\end{equation}
for the terminal control law,
\begin{equation}\label{eq:tmpc_jf}
    \Jf(\x,\xr)=\norm{\x-\xr}_P^2,
\end{equation}
for the terminal cost and
\begin{equation}\label{eq:tmpc_xf}
    \Xf(\xr) = \left\{\x \in \Rx\ \middle|\ \Jf(\x,\xr)\leq\alpha^2\right\},
\end{equation}
for the terminal set. Thus, $\kappafxr$ consists of a feed-forward reference input term and feedback gain matrix $K \in \Rutx$, correcting the error between the reference state $\xr$ and real state $\x$. Furthermore, $\Xf(\xr)$ is a sublevel set of the terminal cost with scaling $\alpha > 0$.

\begin{remark}
    In general, feedback gain matrix $K$ can be state-dependent or time-dependent; see \cite{kohler2020nonlinear} and the discussion therein. For the sake of simplicity, we assume that $K$ is constant in this guide.
\end{remark}

Important to note is that one does not implement $\kappafxr$. Instead, the MPC scheme is implemented in a receding-horizon fashion as given by \eqref{eq:tmpc_cl}. This means that the control action is recomputed at times $t=i\Ts, i \in \N$ and applied during intervals $[i\Ts,(i+1)\Ts), i \in \N$. A problem that could arise in practice is that TMPC problem \eqref{eq:tmpc} becomes infeasible at some point. For example, this happens when the solver experiences numerical problems. This means the solver cannot find a solution satisfying all constraints. In this case, the robot has no known control action to take. The TMPC formulation should satisfy the recursive feasibility property to prevent this issue. This property ensures that, given that a feasible solution to \eqref{eq:tmpc} is found at time $t$, a feasible solution can also be constructed for time $\tTs$ without solving the optimization problem, such that the constraints in \eqref{eq:tmpc} are satisfied at all future times. Recursive feasibility is considered a necessary condition to ensure safe mobile robot deployment.

\begin{remark}
    In dynamic environments, the constraints might vary over time. This could break the recursive feasibility property since the previously optimal solution appended by $\kappafxr$ does not necessarily satisfy the updated constraints. This does not mean the receding-horizon MPC implementation will not find an optimal solution at each time step. This means that we cannot guarantee that a feasible solution will always exist. Therefore, to prove safety in such environments, one must make an assumption about the environment. For example, as long as the robot's trajectory can be predicted well, the moving obstacles will avoid the robot. Alternatively, and possibly more standard, is the assumption that there exist known sets that over-approximate the dynamic obstacles over time \cite{soloperto2019collision}. The latter assumption is usually limited to structured environments with predictable object movements or results in conservative plans. Less conservative motion plans are constructed in more commonly adopted approaches such as stochastic MPC, e.g., \cite{de2021scenario}. These approaches plan safe trajectories in dynamic environments up to a certain risk level based on future object trajectory predictions. However, the level of safety guarantees decreases for these schemes.
\end{remark}

Given the receding-horizon implementation of TMPC, it is also important to show that, even though the predicted state might come closer to the reference trajectory over the prediction horizon, the closed-loop robot state converges to the reference trajectory. We will call this the trajectory tracking property.

Important to note is that the existence of the above-specified terminal ingredients is not guaranteed in general. We will show later how to design them. For now, we assume their existence. This gives the following assumption:
\begin{assumption}[Terminal ingredients]\label{ass:tmpc_term}
    There exist control law \eqref{eq:tmpc_kappaf} with feedback matrix $K \in \Rutx$, terminal cost \eqref{eq:tmpc_jf} with terminal cost matrix $P \in \Rxtx$, and terminal set \eqref{eq:tmpc_xf} with terminal set scaling $\alpha > 0$, such that the following properties hold for all $\x \in \Xf(\xr)$:
    \begin{subequations}\label{eq:tmpc_ass_term}
        \begin{align}
            &\Jf(\x_{\TTt}^*,\x_{\tpTsT}\tr) - \Jf(\x_{\Tt}^*,\x_{\tpT}\tr) \leq -\int_{T}^{T+\Ts} \Js(\x_{\tt}^*,\bu_{\tt}^*,\r_{\tpt})d\tau,\label{eq:tmpc_ass_term_jf}\\
            &\left(\x,\kappafxr\right) \in \Z,\label{eq:tmpc_ass_term_sys}\\
            &M\x \in \F,\label{eq:tmpc_ass_term_obs}
        \end{align}
    \end{subequations}
    with sampling time $\Ts > 0$, optimal solution $(\x_{\ct}^*,\bu_{\ct}^*)$ at $t$ for all $t \geq 0$ and its extended version using terminal control law \eqref{eq:tmpc_kappaf} $\x_{\Tpt|t}^*$ for $\tau \in [0,\Ts]$, as defined in Section~\ref{sec:tmpc_proof_candidate}.
\end{assumption}
This assumption is used to complete the trajectory tracking and recursive feasibility proofs in Sections~\ref{sec:tmpc_proof_tracking} and~\ref{sec:tmpc_proof_rec_feas}. The meaning of the quantities in this assumption will become clearer in the following sections.

\begin{figure}[h!]
    \centering
    \includesvg[width=\textwidth]{tmpc_proofs.svg}
    \caption{Visualization of a 1D reference trajectory $\x_{t+[0,4\Ts]}\tr$, previously optimal solution $\x_{[0,T]|t}^*$, appended part of previously optimal solution $\x_{[T,T+\Ts]|t}^*$, candidate solution $\tx_{[0,T]|\tTs}$, terminal set $\Jf(\x^*,\xr)\leq\alpha^2$ corresponding to $\x_{[0,T]|t}^*$, and terminal set $\Jf(\tx,\xr)\leq\alpha^2$ corresponding to candidate solution. In this case we set $T=4\Ts$. Note that the terminal sets are visualized as ellipses to clarify their shapes, whereas they should be visualized as vertical lines in this case.}
    \label{fig:tmpc}
\end{figure}

To provide some intuition behind the proofs beforehand, Figure~\ref{fig:tmpc} shows the optimal solution at time $t$ and the corresponding candidate solution at time $\tTs$. Thus, the candidate solution covers the prediction horizon of the previously optimal solution shifted by $\Ts$. The idea behind the recursive feasibility proof is to show that the candidate satisfies all constraints \eqref{eq:tmpc_x0}-\eqref{eq:tmpc_term} at time $\tTs$. In interval \circled{2}, these constraints are trivially satisfied since the candidate overlaps with the optimal solution at time $t$. The main thing to show is that once terminal set \eqref{eq:tmpc_xf} is reached at the end of the prediction horizon at $t$, ensured by \eqref{eq:tmpc_term}, we can append control law \eqref{eq:tmpc_kappaf} to the previously optimal solution to obtain a candidate solution that satisfies system and obstacle avoidance constraints \eqref{eq:tmpc_sys} and \eqref{eq:tmpc_obs}, respectively, and terminal set constraint \eqref{eq:tmpc_term}. A trivial method to prove recursive feasibility is to show that control law \eqref{eq:tmpc_kappaf} renders \eqref{eq:tmpc_xf} invariant. This guarantees that the candidate satisfies \eqref{eq:tmpc_term}. Furthermore, by \eqref{eq:tmpc_ass_term_sys} and \eqref{eq:tmpc_ass_term_obs} in Assumption~\ref{ass:tmpc_term}, \eqref{eq:tmpc_sys} and \eqref{eq:tmpc_obs} are also satisfied in terminal set \eqref{eq:tmpc_xf}.

Given that the candidate is a feasible solution to \eqref{eq:tmpc} at $\tTs$, the trajectory tracking proof entails showing that cost \eqref{eq:tmpc_obj} is lower for the candidate than for the previously optimal solution, meaning that the robot will converge to the reference trajectory even though \eqref{eq:tmpc} is not solved again at $\tTs$. Intuitively, this holds if the cost related to interval \circled{4}, shortly cost \circled{4}, is lower than terminal cost \circled{3} such that the candidate cost decreases at least by cost \circled{1} compared to the cost of the previously optimal solution since cost \circled{2} remains the same. This is exactly the condition that is written mathematically in \eqref{eq:tmpc_ass_term_jf} in Assumption~\ref{ass:tmpc_term}.

In summary, ensuring that TMPC formulation \eqref{eq:tmpc} leads to safe and efficient mobile robot deployment requires the design of suitable terminal ingredients. To this end, we define the candidate solution based on the terminal ingredients in Section~\ref{sec:tmpc_proof_candidate} that allows us to prove recursive feasibility in Section~\ref{sec:tmpc_proof_rec_feas} and trajectory tracking in Section~\ref{sec:tmpc_proof_tracking}. As will become clear, Assumption~\ref{ass:tmpc_term} is not the only assumption to satisfy to complete the proofs. Therefore, the goal of Section~\ref{sec:tmpc_properties_design} is to design the involved quantities so that all assumptions are satisfied and the desired TMPC properties hold.

\subsubsection{Candidate solution}\label{sec:tmpc_proof_candidate}
The goal of defining the candidate solution is to show that there exists a feasible, not necessarily optimal, solution to \eqref{eq:tmpc} at time $\tTs$ given that \eqref{eq:tmpc} is successfully solved at time $t$. The trivial way to construct such a candidate is to pick the overlapping part with the optimal solution from time $t$, i.e., the part in prediction interval $\tau\in[\Ts,T]$ at $t$, since this trajectory is known to have satisfied all constraints in the MPC run at time $t$ and overlaps with $\tau\in[0,T-\Ts]$ at $\tTs$. What is left is to define the candidate for $\tau\in[T-\Ts,T]$. This trajectory part satisfies the constraints if we find $\kappafxr$ with the abovementioned properties.

Let's now mathematically define the candidate. For convenience of notation, we define the appended part of the previously optimal solution, for $\tau \in [T,T+\Ts]$, as
\begin{equation}\label{eq:tmpc_u_append}
    \bu_{\tt}^* = \kappaf(\x_{\tt}^*,\r_{\tpt}),
\end{equation}
with $\dx_{\tt}^*$ given by \eqref{eq:xdot_nom} by applying \eqref{eq:tmpc_u_append}.

Given these definitions, we can write the candidate for time $\tTs$ as
\begin{equation}\label{eq:tmpc_candidate}
    \tu_{\ttT} = \bu_{\tTt}^*,\ \tx_{\ttT} = \x_{\tTt}^*,\ \tau \in [0,T].
\end{equation}

The candidate, and thus $\kappafxr$, is not necessarily implemented in practice. It is only used to prove recursive feasibility and trajectory tracking. One would need this control law if the problem runs infeasible, and one wants the robot to continue executing the open-loop MPC prediction during deployment. However, in that situation, the control action no longer considers any changes in the robot's state or environment, which can be dangerous. In that case, it would be better to stop, check the log to see why the problem was infeasible, and fix it.

\begin{remark}
    Since the candidate gives the desired recursive feasibility and trajectory tracking properties as shown in the next sections, it would be sufficient to only solve \eqref{eq:tmpc} at $t=0$ and apply the candidate for $t > 0$. This reasoning conforms to the arguments presented in \cite{scokaert1999suboptimal}: feasibility implies stability. However, the resulting control policy is sub-optimal regarding the defined objective function. Therefore, the user can explicitly trade optimality and computational efficiency.
\end{remark}

Although the candidate is not necessarily implemented in practice, it is a suitable warm start for the solver since it is a feasible solution that might be close to the optimal solution. Providing a proper warm start to the solver is vital to speeding up the solve time, as the solver needs fewer iterations to go from the initial point to the optimal solution.

\subsubsection{Recursive feasibility proof}\label{sec:tmpc_proof_rec_feas}
To prove recursive feasibility, our task is to show that candidate \eqref{eq:tmpc_candidate} satisfies all constraints \eqref{eq:tmpc_x0}-\eqref{eq:tmpc_term} at time $\tTs$. First, note that, without knowing anything about the terminal ingredients, the candidate solution implies that constraint \eqref{eq:tmpc_x0} is satisfied and constraint \eqref{eq:tmpc_xdot} is satisfied for $\tau\in[0,T]$. That leaves us to show that constraints \eqref{eq:tmpc_sys} and \eqref{eq:tmpc_obs} are satisfied for $\tau\in[0,T]$ and that the terminal set constraint \eqref{eq:tmpc_term} is satisfied at $\tau=T$.

To show that constraints \eqref{eq:tmpc_sys} and \eqref{eq:tmpc_obs} are satisfied for $\tau\in[0,T]$, we split this interval up into the half-open interval $\tau\in[0,T-\Ts)$ and the closed interval $\tau\in[T-\Ts,T]$.\\

\noindent\textbf{System constraints satisfaction for $\bm{\tau\in[0,T-\Ts)}$}\\
Let's start with the proofs for $\tau\in[0,T-\Ts)$. For this interval, the following holds for the system constraints $j \in \Ns$:
\begin{equation}\label{eq:tmpc_proof_sys}
    \gjs(\tx_{\ttT},\tu_{\ttT}) \overset{\eqref{eq:tmpc_candidate}}{=} \gjs(\x_{\tTt}^*,\bu_{\tTt}^*) \overset{\eqref{eq:tmpc_sys}}{\leq} 0.
\end{equation}

\noindent\textbf{Obstacle avoidance constraints satisfaction for $\bm{\tau\in[0,T-\Ts)}$}\\
Similarly, the obstacle avoidance constraints $j \in \No$ are satisfied for $\tau\in[0,T-\Ts)$:
\begin{equation}\label{eq:tmpc_proof_obs}
    \gjttTo(\tp_{\ttT}) \overset{\eqref{eq:tmpc_candidate}}{=} \gjttTo(\p_{\tTt}^*) \overset{\eqref{eq:tmpc_obs}}{\leq} 0.
\end{equation}
Note that, in contrast to the system constraints, the obstacle avoidance constraints are time-varying. Therefore, we need the following assumption on the obstacle avoidance constraints in order to show that $\gjttTo(\p_{\tTt}^*) \leq 0$ actually holds:
\begin{assumption}[Obstacle avoidance constraints]\label{ass:obs}
    The obstacle avoidance constraints should satisfy
    \begin{subequations}\label{eq:obs_contained}
        \begin{align}
            \p_{\tTt}^* &\in \F_{\ttT},\ \tau \in [0,T-\Ts],\\
            \p_{\Tt}^* &\in \F_{\Tt}.
        \end{align}
    \end{subequations}
    In words, the previously optimal path $\p_{\tTt}^*$ should be contained in the corresponding constraint regions $\F_{\ttT}$ applied for $\tau \in [0,T-\Ts]$ at $\tTs$. Similarly, the terminal position $\p_{\Tt}^*$ should be contained in the terminal constraint region $\F_{\Tt}$. The latter property ensures that the appended part of the candidate solution, given by \eqref{eq:tmpc_u_append}, satisfies the updated obstacle avoidance constraints at time $\tTs$. This property is used to show terminal constraint satisfaction below. We assume the `corresponding' constraint regions are constructed using the previously optimal solution.
\end{assumption}

\noindent\textbf{Terminal set constraint satisfaction for $\bm{\tau\in[T-\Ts,T]}$}\\
Next, we prove that if \eqref{eq:tmpc_term} is satisfied at time $t$, the candidate \eqref{eq:tmpc_candidate} satisfies \eqref{eq:tmpc_term} in prediction interval $\tau\in[T-\Ts,T]$ for time $\tTs$. For this proof, we leverage the assumed decrease in $\Jf(\x,\xr)$, given by \eqref{eq:tmpc_ass_term_jf} in Assumption~\ref{ass:tmpc_term}. Correspondingly, the following holds for $\tau \in [T-\Ts,T]$:
\begin{equation}\label{eq:tmpc_proof_term}
    \Jf(\tx_{\ttT},\x_{\tpTst}\tr) \overset{\eqref{eq:tmpc_candidate}}{=} \Jf(\x_{\tTt}^*,\x_{\tpTst}\tr) \overset{\eqref{eq:tmpc_ass_term_jf}}{\leq} \Jf(\x_{\Tt},\x_{\tpT}\tr) \overset{\eqref{eq:tmpc_term}\eqref{eq:tmpc_xf}}{\leq} \alpha^2.
\end{equation}

\noindent\textbf{System constraints satisfaction for $\bm{\tau\in[T-\Ts,T]}$}\\
Satisfaction of system constraints \eqref{eq:tmpc_sys} in prediction interval $\tau\in[T-\Ts,T]$ is a direct consequence of the combination of terminal set invariance, as shown in the previous section, and invoking \eqref{eq:tmpc_ass_term_sys} in Assumption~\ref{ass:tmpc_term}.\\

\noindent\textbf{Obstacle avoidance constraints satisfaction for $\bm{\tau\in[T-\Ts,T]}$}\\
Similar to the system constraints satisfaction, satisfaction of obstacle avoidance constraints \eqref{eq:tmpc_obs} in prediction interval $\tau\in[T-\Ts,T]$ is a direct consequence of the combination of terminal set invariance, as shown in the previous section, and invoking \eqref{eq:tmpc_ass_term_obs} in Assumption~\ref{ass:tmpc_term}.
\qed

\subsubsection{Trajectory tracking proof}\label{sec:tmpc_proof_tracking}
Similar to proving the stability of an autonomous system, proving that nominal system \eqref{eq:xdot_nom} tracks reference trajectory \eqref{eq:r} under closed-loop control law \eqref{eq:tmpc_cl} follows arguments from Lyapunov theory. A useful analogy to Lyapunov's theory is the energy captured in a mechanical system. In this analogy, the Lyapunov function describes the energy stored in the system. In such a system, the energy decreases over time because of dissipation effects like friction, meaning that the Lyapunov function value will strictly decrease over time. In the context of MPC, the energy is based on the tracking error, and the dissipation effect that reduces the tracking error is the closed-loop control law \eqref{eq:tmpc_cl}. Thus, a logical choice for the Lyapunov function is to pick the optimal cost $\J_t^*(\x_t,\r_t)$ in \eqref{eq:tmpc} since this cost indicates how close the predicted trajectory is to the reference trajectory. Ideally, we would like the optimal cost to decrease over time, meaning the tracking error converges to zero.

To derive a bound on the optimal cost, we follow similar steps to the standard descent proof for the candidate cost in \cite{rawlings2017model}:
\begin{subequations}\label{eq:tmpc_jopt_db}
    \begin{alignat}{2}
        \J_{\tTs}^*(\x_{\tTs},\r_{\tTs})
        &\overset{\text{(a)}}{\leq} &&\int_{0}^{T} \Js(\tx_{\ttT},\tu_{\ttT},\r_{\tpTst})d\tau + \Jf(\tx_{\TtT},\x_{\tpTsT}\tr)\\
        &\overset{\text{(b)}}{=} &&\int_{0}^{T-\Ts} \Js(\tx_{\ttT},\tu_{\ttT},\r_{\tpTst})d\tau \notag\\
        &&&+\int_{T-\Ts}^{T} \Js(\tx_{\ttT},\tu_{\ttT},\r_{\tpTst})d\tau + \Jf(\tx_{\TtT},\x_{\tpTsT}\tr)\\
        &\overset{\text{(c)}}{=} &&\int_{\Ts}^{T} \Js(\x_{\tt}^*,\bu_{\tt}^*,\r_{\tpt})d\tau+\int_{T}^{T+\Ts} \Js(\x_{\tt}^*,\bu_{\tt}^*,\r_{\tpt})d\tau \notag\\
        &&&+\Jf(\tx_{\TtT},\x_{\tpTsT}\tr)\\
        &\overset{\text{(d)}}{=} &&\int_{0}^{T} \Js(\x_{\tt}^*,\bu_{\tt}^*,\r_{\tpt})d\tau - \int_{0}^{\Ts} \Js(\x_{\tt}^*,\bu_{\tt}^*,\r_{\tpt})d\tau \notag\\
        &&&+\int_{T}^{T+\Ts} \Js(\x_{\tt}^*,\bu_{\tt}^*,\r_{\tpt})d\tau + \Jf(\tx_{\TtT},\x_{\tpTsT}\tr)\\
        &\overset{\text{(e)}}{=} &&\J_t^*(\x_t,\r_t) - \Jf(\x_{\Tt}^*,\x_{\tpT}\tr) - \int_{0}^{\Ts} \Js(\x_{\tt}^*,\bu_{\tt}^*,\r_{\tpt})d\tau \notag\\
        &&&+\int_{T}^{T+\Ts} \Js(\x_{\tt}^*,\bu_{\tt}^*,\r_{\tpt})d\tau + \Jf(\tx_{\TtT},\x_{\tpTsT}\tr)\\
        &\overset{\text{(f)}}{=} &&\J_t^*(\x_t,\r_t) - \int_{0}^{\Ts} \Js(\x_{\tt}^*,\bu_{\tt}^*,\r_{\tpt})d\tau + \Jf(\tx_{\TtT},\x_{\tpTsT}\tr) \notag\\
        &&&-\Jf(\x_{\Tt}^*,\x_{\tpT}\tr) + \int_{T}^{T+\Ts} \Js(\x_{\tt}^*,\bu_{\tt}^*,\r_{\tpt})d\tau\\
        &\overset{\text{(g)}}{=} &&\J_t^*(\x_t,\r_t) - \int_{0}^{\Ts} \Js(\x_{\tt}^*,\bu_{\tt}^*,\r_{\tpt})d\tau + \Jf(\x_{\TTt}^*,\x_{\tpTsT}\tr) \notag\\
        &&&-\Jf(\x_{\Tt}^*,\x_{\tpT}\tr) + \int_{T}^{T+\Ts} \Js(\x_{\tt}^*,\bu_{\tt}^*,\r_{\tpt})d\tau\\
        &\overset{\text{(h)}}{\leq} &&\J_t^*(\x_t,\r_t) - \int_{0}^{\Ts} \Js(\x_{\tt}^*,\bu_{\tt}^*,\r_{\tpt})d\tau,
    \end{alignat}
\end{subequations}
where (a) is obtained by filling in \eqref{eq:tmpc_obj} with the feasible, but not necessarily optimal candidate \eqref{eq:tmpc_candidate}, the so-called candidate cost, (b) by splitting the integral term at $T-\Ts$, (c) by filling in candidate \eqref{eq:tmpc_candidate}, (d) by adding and subtracting $\int_{0}^{\Ts} \Js(\x_{\tt}^*,\bu_{\tt}^*,\r_{\tpt})d\tau$, (e) by replacing $\int_{0}^{T} \Js(\x_{\tt}^*,\bu_{\tt}^*,\r_{\tpt})d\tau$ with $\J_t^*(\x_t,\r_t) - \Jf(\x_{\Tt}^*,\x_{\tpT}\tr)$, (f) by conveniently re-arranging the terms, (g) by \eqref{eq:tmpc_candidate}, and (h) by \eqref{eq:tmpc_ass_term_jf} in Assumption~\ref{ass:tmpc_term}.

In simple words, \eqref{eq:tmpc_ass_term_jf} in Assumption~\ref{ass:tmpc_term} states that the added candidate cost in prediction interval $\tau\in[T-\Ts,T]$ at $\tTs$ should be less than the terminal predicted cost at predicted time $T$ at $t$. If this is the case, the candidate cost $\J_{\tTs}(\x_{\tTs},\r_{\tTs})$ is at least the stage cost for $\tau\in[0,\Ts]$ at $t$ smaller than the previously optimal cost $\J_t^*(\x_t,\r_t)$. Thus, even without optimizing \eqref{eq:tmpc} at time $\tTs$ and instead applying candidate \eqref{eq:tmpc_candidate}, the cost strictly decreases. Given that the optimal cost $\J_{\tTs}^*(\x_{\tTs},\r_{\tTs})$ is upper-bounded by the candidate cost, the optimal cost also strictly decreases.

Thus, \eqref{eq:tmpc_ass_term_jf} in Assumption~\ref{ass:tmpc_term} implies
\begin{equation}\label{eq:tmpc_jopt_db_2}
    \J_{\tTs}^*(\x_{\tTs},\r_{\tTs}) - \J_t^*(\x_t,\r_t) \leq -\int_{0}^{\Ts} \Js(\x_{\tt}^*,\bu_{\tt}^*,\r_{\tpt})d\tau.
\end{equation}
It is not trivial to find a $\classKinf$ function that forms a lower bound to $\int_{0}^{\Ts} \Js(\x_{\tt}^*,\bu_{\tt}^*,\r_{\tpt})d\tau$, which would allow us to show that $\int_{0}^{\Ts} \Js(\x_{\tt}^*,\bu_{\tt}^*,\r_{\tpt})d\tau > 0$. Hence, we cannot prove stability utilizing standard Lyapunov arguments outlined in \cite{rawlings2017model}. As a convergence proof does not require this bound, we prove convergence instead.

The descent property of the optimal cost \eqref{eq:tmpc_jopt_db_2} implies
\begin{equation}\label{eq:tmpc_jopt_db_3}
    \J_{\tTs}^*(\x_{\tTs},\r_{\tTs}) - \J_t^*(\x_t,\r_t) \leq -c^{\J,\mathrm{d}} \int_{t}^{\tTs} \norm{\x_\tau-\x_\tau\tr}_Q^2d\tau,
\end{equation}
with constant $c^{\J,\mathrm{d}} > 0$. Reversing the sign in \eqref{eq:tmpc_jopt_db_3}, iterating this inequality from $0$ to $t$ with $t \rightarrow \infty$, and using the fact that $\J_t^*(\x_t,\r_t)$ is uniformly bounded yields
\begin{equation}\label{eq:tmpc_err_ub}
    \underset{t\rightarrow\infty}{\operatorname{lim}} \int_{0}^{t} \norm{\x_\tau-\x_\tau\tr}_Q^2d\tau \leq \J_t^*(\x_0,\r_0) - \underset{t\rightarrow\infty}{\operatorname{lim}}\J_t^*(\x_t,\r_t) \leq \J_t^*(\x_0,\r_0) < \infty.
\end{equation}

By applying Barbalat's lemma \cite{barbalat1959systemes}, we can conclude that the tracking error $\norm{\x_t-\x_t\tr}$ asymptotically converges to zero. \qed

\subsubsection{Design to satisfy the assumptions}\label{sec:tmpc_properties_design}
This section will detail the design of:
\begin{itemize}
    \item the obstacle avoidance constraints, which need to satisfy Assumption~\ref{ass:obs};
    \item the terminal ingredients, which need to satisfy \eqref{eq:tmpc_ass_term_jf} in Assumption~\ref{ass:tmpc_term};
    \item the reference trajectory design, which needs to satisfy \eqref{eq:tmpc_ass_term_sys} and \eqref{eq:tmpc_ass_term_obs} in Assumption~\ref{ass:tmpc_term}.
\end{itemize}

\noindent\textbf{Obstacle avoidance constraints design}\\
To prove recursive feasibility, we need the obstacle avoidance constraints to satisfy Assumption~\ref{ass:obs}. A trivial way to satisfy this assumption is to generate the obstacle avoidance constraints based on the previously optimal path $\p_{\ct}^*$ and ensure that this path is also contained in the corresponding constraint regions.

\begin{remark}
    Since TMPC \eqref{eq:tmpc} will be implemented using multiple shooting with discretized dynamics \cite{rawlings2017model}, we also define the obstacle avoidance constraints based on piecewise-constant segments between MPC stages. This means that the obstacle avoidance constraints are also piecewise defined according to
    \begin{equation}\label{eq:obs_piecewise}
        \F_{i\Ts+\tau|\tTs} = \F_{i\Ts|\tTs},\ \tau \in (0,\Ts],\ i \in \mathbb{N}_{[0,N-1]},
    \end{equation}
    where $N$ is the number of discrete predicted stages.
\end{remark}

\noindent\textbf{Terminal ingredients design}\\
The terminal ingredients design involves computing a suitable combination of terminal control law \eqref{eq:tmpc_kappaf}, terminal cost \ref{eq:tmpc_jf}, and terminal set \eqref{eq:tmpc_xf} that satisfy terminal cost decrease \eqref{eq:tmpc_ass_term_jf} in Assumption~\ref{ass:tmpc_term}. This design involves finding a suitable combination of matrices $P$ and $K$ and constant $\alpha > 0$.

To design $P$ and $K$, recent work \cite{kohler2020nonlinear} proposes a linear matrix inequality (LMI) to ensure direct satisfaction of Assumption~\ref{ass:tmpc_term}:
\begin{equation}\label{eq:lmi_tracking}
    \left(A(\r)+B(\r)K(\r)\right)^\top P(\r) + P(\r)\left(A(\r)+B(\r)K(\r)\right) + \sum_{j=1}^{\nx+\nuu} \frac{\partial P(\r)}{\partial \x}\dx_j + \left(\Qeps+K(\r)RK(\r)\right) \leq 0,
\end{equation}
with Jacobians
\begin{equation}\label{eq:tmpc_a_b}
    A(\r) \coloneqq \left.\frac{\partial f(\xu)}{\partial \x}\right|_{\r}, \quad B(\r) \coloneqq \left.\frac{\partial f(\xu)}{\partial \bu}\right|_{\r},
\end{equation}
$\Qeps \coloneqq Q+\epsilon \Inx$, and $\epsilon > 0$. See \cite{kohler2020nonlinear} for the corresponding proof.

In particular, this LMI ensures a system property called local incremental stabilizability. See \cite{angeli2002lyapunov,kohler2018nonlinear} for more details. Note that $P$ and $K$ are parameterized based on local values $\r$ in the state-input space of the reference trajectory. This is where the word `local' comes from. It suffices to show that certain properties hold locally in the state-input space to ensure tracking of reference trajectories in the complete state-input space. However, note that the results only hold for sufficiently small values of terminal set scaling $\alpha$ such that Jacobians \eqref{eq:tmpc_a_b} are accurate enough to describe the nonlinear system behavior. The word `incremental' refers to the fact that we consider the stabilizability property of the system trajectory that evolves close to the reference trajectory.

As mentioned before, terminal cost \eqref{eq:tmpc_jf} is an upper bound of the `infinite-horizon' cost that we would ideally like to minimize. Therefore, the goal is to find the minimum-valued matrix $P(\r)$ and corresponding matrix $K(\r)$ satisfying LMI \eqref{eq:lmi_tracking}. Correspondingly, we can formulate the following semi-definite program (SDP):
\begin{subequations}\label{eq:tmpc_sdp}
    \begin{align}
        \underset{X,Y}{\opmin}\ \ &-\operatorname{log}\operatorname{det}\Xmin,\label{eq:tmpc_sdp_obj}\\
        \operatorname{s.t.}\ &X(\r) \succeq 0,\label{eq:tmpc_sdp_lmi_x}\\
        &\begin{bmatrix}
            A(\r)X(\r)+B(\r)Y(\r)+\left(A(\r)X(\r)+B(\r)Y(\r)\right)^\top-\dot{X}(\r)&\left(\Qeps^\frac{1}{2} X(\r)\right)^\top&\left(R^\frac{1}{2}Y(\r)\right)^\top\\
            \Qeps^\frac{1}{2} X(\r)&-\Inx&0^{\nxtu}\\
            R^\frac{1}{2}Y(\r)&0^{\nutx}&-\Inu
        \end{bmatrix} \preceq 0,\label{eq:tmpc_sdp_lmi_tracking}\\
        &\Xmin \preceq X(\r),\ \forall \r \in \bZ,\notag
    \end{align}
\end{subequations}
with $\dot{X}(\r) \coloneqq \frac{\partial X(\r)}{\partial \xr}\dxr$ and LMI \eqref{eq:tmpc_sdp_lmi_tracking} being the direct result of taking the Schur complement of LMI \eqref{eq:lmi_tracking} and applying the following coordinate transform:
\begin{equation}\label{eq:tmpc_p_k}
    P(\r) = X(\r)^{-1},\ K(\r) = Y(\r)P(\r).
\end{equation}

Note that $\bZ$ represents the continuous state-input space of the reference trajectory. Therefore, SDP \eqref{eq:tmpc_sdp} is semi-infinite, and we need to grid or convexify $\bZ$ to solve it. For solving the SDP, the states and inputs in $\r$ that do not appear in the Jacobians \eqref{eq:tmpc_a_b} can be set to zero, we can take the vertices for the ones that appear linearly, and the ones that appear nonlinearly need to be gridded.

\begin{remark}
    Since the computation time of generating LMI \eqref{eq:tmpc_sdp_lmi_tracking} can become significantly large, a solution is to solve \eqref{eq:tmpc_sdp} with a coarse grid of points and evaluate the result at a denser grid to check the validity of the result.
\end{remark}

Given this method to compute $P(\r)$ and $K(\r)$, the only thing left is to design a suitable value for terminal set scaling $\alpha$. Theoretically, one could set $\alpha = 0$, such that \eqref{eq:tmpc_term} becomes an equality constraint. However, this unnecessarily restricts the solver from finding a solution. This might lead to infeasibility, either because a solution satisfying this equality constraint is non-existent or because the problem is numerically infeasible. Therefore, the idea is to find a non-conservative maximum value for $\alpha$. Recall that \eqref{eq:tmpc_xf} is centered around the reference trajectory, and the reference trajectory satisfies tightened system and obstacle avoidance constraints \eqref{eq:r_sys} and \eqref{eq:r_obs}, respectively. Given this information, $\alpha$ can only grow as large as the tightening applied to the original system and obstacle avoidance constraints.

Let's assume the reference trajectory is only subject to tightened system constraints and is constrained by the following polytopic set: $\bZ = \left\{\r \in \Rux\ \middle|\ \Ljrs \r \leq \ljrs,\ j \in \Ns\right\}$. The tightening equals $\ljs - \Ljrs \r, \r \in \bZ, j \in \No$. As a result, $\alpha$ can be computed by solving the following linear program (LP) \cite{kohler2020nonlinear}, which is proven in Appendix~\ref{app:lp_alpha}:
\begin{subequations}\label{eq:lp_alpha}
    \begin{alignat}{2}
        \alpha^* = &\underset{\alpha}{\opmax}\ &&\alpha,\label{eq:lp_alpha_obj}\\
        &\operatorname{s.t.}\ &&\lrnorm{P(\r)^{-\frac{1}{2}} \left[K(\r)^\top \Inx\right] \Ljrs^\top} \alpha \leq (\ljs - \Ljrs \r),\label{eq:lp_alpha_con}\\
        &&&\forall \r \in \bZ,\ j \in \Ns.
    \end{alignat}
\end{subequations}
Naturally, this result can also be extended to include obstacle avoidance constraints.

Note that using this expression, we can either compute the value of $\alpha$ based on the tightening of the system constraints for the reference trajectory or vice versa. Therefore, we need more details on the system and obstacle avoidance constraints before finishing the design of $\alpha$. This will be described in the next section.\\

\noindent\textbf{Reference trajectory design}\\
The reference trajectory needs to satisfy tightened system constraints \eqref{eq:r_sys} and tightened obstacle avoidance constraints \eqref{eq:r_obs}. These constraint sets need to be constructed such that the system and obstacle avoidance constraints are satisfied in terminal set \eqref{eq:tmpc_xf} according to \eqref{eq:tmpc_ass_term_sys} and \eqref{eq:tmpc_ass_term_obs} in Assumption~\ref{ass:tmpc_term}.

Given the fact that system constraints \eqref{eq:con_sys} are continuously differentiable and defined on compact set $\Z$, they are Lipschitz continuous. Therefore, we can express them as
\begin{equation}
    \gjs(\x',\bu') - \gjs(\xu) \leq \cjs \alpha,\ j \in \Ns,
\end{equation}
for some $\x'\neq\x \in \Rx$ and $\bu'\neq\bu \in \Ru$ with Lipschitz constants $\cjs, j \in \Ns$. As a result, if we set the tightened system constraints of the reference trajectory as
\begin{equation}\label{eq:tmpc_r_sys_ineq}
    \gjrs(\xu) \leq \gjs(\xu) + \cjs \alpha,\ j \in \Ns,
\end{equation}
we get a strict equality in \eqref{eq:lp_alpha_con} with
\begin{equation}\label{eq:tmpc_cs}
    \cjs = \lrnorm{P(\r)^{-\frac{1}{2}} \left[K(\r)^\top \Inx\right] \Ljrs^\top},\ j \in \Ns.
\end{equation}

Equivalently, if we set the upper bound of the tightened obstacle avoidance constraints of the reference trajectory as
\begin{equation}\label{eq:tmpc_r_obs_ineq}
    \gro(\p) \leq \go(\p) + \co \alpha,
\end{equation}
we can replace $\left[K(\r)^\top \Inx\right] \Ljrs^\top$ by $M^\top$ to obtain the following expression for $\co$:
\begin{equation}\label{eq:tmpc_co}
    \co = \lrnorm{P(\r)^{-\frac{1}{2}} M^\top}.
\end{equation}

Given this design, we choose a value for $\alpha$ that results in a desired tightening of either of the system constraints or the obstacle avoidance constraints. A natural choice for mobile robots is to decide on the maximum distance between reference trajectory and obstacles. Let's call this quantity $\dmax$. Then, we know that $\co \alpha = \dmax$, which results in the following expression for the terminal set scaling:
\begin{equation}\label{eq:tmpc_alpha}
    \alpha = \frac{\dmax}{\co}.
\end{equation}

By solving \eqref{eq:tmpc_sdp} some system constraints might be tightened more than others, leading to empty feasible sets. This issue can be avoided by including the system constraints tightening constants in \eqref{eq:tmpc_sdp_obj} and setting their penalties equal to the inverse of the space in the corresponding system constraint direction $j \in \Ns$. This way, the solver is encouraged to tighten the system constraints equally. Working out Lipschitz inequality on the system constraints \eqref{eq:tmpc_r_sys_ineq} using $\alpha = \norm{\x-\xr}_{P(\r)}$ and taking the Schur complement according to \cite[p. 29]{nubert2019learning} results in the following LMI:
\begin{equation}\label{eq:lmi_sys}
    \begin{bmatrix}
        {c_{j}^\mathrm{s}}^2&L_{j}^\mathrm{s}\begin{bmatrix}Y\\X\end{bmatrix}\\
        \left(L_{j}^\mathrm{s}\begin{bmatrix}Y\\X\end{bmatrix}\right)^\top&X
    \end{bmatrix} \succeq 0,\ j \in \Ns.
\end{equation}

This LMI is included in the SDP formulation in \cite{benders2025embedded}. In that formulation, the state- and input-dependency of $P$ and $K$ are also removed, so the results hold in the complete state-input space, not just locally. Therefore, the constant $\epsilon$ from \cite{kohler2020nonlinear}, that accounts for the Lipschitz continuity of Jacobians \eqref{eq:tmpc_a_b}, can be removed. The advantage of including the state- and input-dependency is the reduced conservatism, i.e., the terminal cost is smaller and the terminal set is larger. This results in a smaller tracking error of the closed-loop system. However, the increased conservatism was not a major concern in \cite{benders2025embedded} and resulted in a simpler formulation.

\subsection{Concluding remarks}\label{sec:tmpc_remarks}
In conclusion, the presented TMPC scheme is an effective tool for providing safety guarantees for tracking dynamically feasible reference trajectories with a mobile robot described by nominal dynamics \eqref{eq:xdot_nom}. The scheme is based on the idea that suitably designed terminal ingredients result in a strict decrease in terminal cost and positive invariance of the terminal set, thereby allowing for proving trajectory tracking and recursive feasibility of the scheme.

Note that the solver will only find a feasible solution to \eqref{eq:tmpc} if the system starts sufficiently close to the reference trajectory such that terminal set constraint \eqref{eq:tmpc_term} is satisfied.

In practice, a physical robot is subject to dynamical disturbances that could be impossible to model. To ensure safe mobile robot navigation despite these disturbances, we need to know how big the effects of the disturbances on the dynamics are and account for that in the MPC design. This is the topic of Section~\ref{sec:rmpc} in which the disturbed robot dynamics and a corresponding robust MPC (RMPC) design are provided.

\section{Robust MPC for trajectory tracking with disturbances}\label{sec:rmpc}
The goal of this section is to introduce the RMPC scheme that will be used to track reference trajectories satisfying Property~\ref{property:r} when the system is subject to disturbances. The description of the system dynamics is provided in Section~\ref{sec:rmpc_sys_w}. Like Section~\ref{sec:tmpc_formulation}, Section~\ref{sec:rmpc_formulation} states the optimization problem and closed-loop control law. Then, Section~\ref{sec:rmpc_properties} builds intuition for RMPC and explains the required elements in the formulation and related properties. Finally, Section~\ref{sec:rmpc_conclusion} concludes the RMPC formulation.

\subsection{Disturbed mobile robot description}\label{sec:rmpc_sys_w}
The disturbed system dynamics are given by
\begin{equation}\label{eq:xdot_w}
    \dx = \fwarg \coloneqq f(\x, \bu) + E\w,
\end{equation}
with the same properties as the nominal system \eqref{eq:xdot_nom}, disturbance $\w \in \Rw$, and disturbance selection matrix $E \in \mathbb{R}^{\nx \times \nw}$. From now on, to make a distinction between the nominal and disturbed system, let's denote nominal and disturbed states by $\z$ and $\x$, respectively.

By recording data one can compute a bounding box with lower bound $\ubw_i\rec$ and upper bound $\bbw_i\rec$ for each dimension $i \in \Nw$ of the recorded disturbance values. Let's denote the offset for each dimension $i$ by $\w_i\b \coloneqq \frac{\ubw_i\rec+\bbw_i\rec}{2}$. Then, the disturbance set centered around $\bm{0}$ is given by
\begin{equation}\label{eq:w_bb}
    \W \coloneqq \left\{\w \in \Rw\ \middle|\ \ubw_i^\mathrm{rec,0} \leq \w_i \leq \bbw_i^\mathrm{rec,0},\ i \in \Nw\right\},
\end{equation}
with $\ubw_i^\mathrm{rec,0} \coloneqq \ubw_i\rec - \w_i\b$ and $\bbw_i^\mathrm{rec,0} \coloneqq \bbw_i\rec - \w_i\b$, such that the recorded disturbance values are all contained in set $\wb\oplus\W$. Correspondingly, we write the disturbance values in the next sections as
\begin{equation}
    \w \coloneqq \wb + \w^0 \in \wb\oplus\W.
\end{equation}
Model bias $\wb$ can be included in the nominal prediction model to account for the static offset of disturbances as described in the next section.

\subsection{RMPC formulation}\label{sec:rmpc_formulation}
Consider the following RMPC formulation for tracking reference trajectory \eqref{eq:r}:
\begin{subequations}\label{eq:rmpc}
    \begin{alignat}{3}
        \J_t^*(\x_t,\r_t) = \underset{\substack{\z_{\ct},\bv_{\ct}}}{\opmin}\ \ &\mathrlap{\Jf(\z_{\Tt},\x_{\tpT}\tr) + \int_{0}^{T} \Js(\z_{\tt},\bv_{\tt},\r_{\tpt})\ d\tau,}&&&&\hspace{120pt} \label{eq:rmpc_obj}\\
        \operatorname{s.t.}\ &\z_{\0t} = \x_t,&&&& \label{eq:rmpc_z0}\\
        &s_\tau = 0,&&&& \label{eq:rmpc_s0}\\
        &\dz_{\tt} = f(\z_{\tt},\bv_{\tt}) + E\wb,&&&&\ \tau \in [0,T], \label{eq:rmpc_zdot}\\
        &\ds_\tau = -\rho s_\tau + \bw,&&&&\ \tau \in [0,T], \label{eq:rmpc_sdot}\\
        &\gjs(\z_{\tt},\bv_{\tt}) + \cjs s_\tau \leq 0,&&\ j \in \Ns,&&\ \tau \in [0,T], \label{eq:rmpc_sys}\\
        &\gjtto(M\z_{\tt}) + \co s_\tau \leq 0,&&\ j \in \No,&&\ \tau \in [0,T], \label{eq:rmpc_obs}\\
        &(\z_{\Tt},s_\tau) \in \Xf(\x_{\tpT}\tr), \label{eq:rmpc_term}
    \end{alignat}
\end{subequations}
with stage cost $\Js(\z,\bv,\r)=\norm{\z-\xr}_Q^2+\norm{\bv-\ur}_R^2, Q \succ 0, R \succ 0$ and terminal cost $\Jf(\z,\xr)=\norm{\z-\xr}_P^2, P \succ 0$. In this formulation, $Q$ and $R$ are tuning matrices and $P$ should be suitably computed to show practical asymptotic convergence, see Section~\ref{sec:rmpc_proof_tracking}. Note the inclusion of model bias $\wb$ in nominal system dynamics \eqref{eq:rmpc_zdot} such that the model best describes the minimum and maximum disturbance bounds.

Similar to the TMPC formulation in Section~\ref{sec:tmpc}, the RMPC formulation is applied in a receding-horizon fashion. In this case, the control law for the closed-loop system is defined as follows for $t \geq 0$:
\begin{equation}\label{eq:rmpc_cl}
    \bu_{\tpt} = \kappa(\x_{\tpt},\z_{\tt}^*,\bv_{\tt}^*) \coloneqq \bv_{\tt}^* + \kappad(\x_{\tpt},\z_{\tt}^*),\ \tau \in [0,\Ts],
\end{equation}
with control law $\kappaxzv : \kappadef$ including feedback law $\kappadxz : \kappaddef$ that will be defined later according to the design presented in this section.

\subsection{RMPC properties}\label{sec:rmpc_properties}
If the nominal system \eqref{eq:xdot_nom} is subject to a predictable disturbance, we could adjust \eqref{eq:xdot_nom} to incorporate the corresponding disturbance dynamics and apply the TMPC described in the previous section. However, in general, disturbances are not or only partially predictable. The predictable part could be learned with a learning-based method such as Gaussian Processes, see for example \cite{probst2023data,carena2024wind,dubied2025robust}. However, in almost all cases, there will be a part of the dynamics that cannot be predicted. In that case, we still need to be able to guarantee that operating our MPC scheme on the robot is safe.

Then why not execute the TMPC from Section~\ref{sec:tmpc} and hope for the best? Let's consider the scenario where the reference trajectory is computed using the same constraints as the TMPC. At some time $t$, the state might exactly overlap with the reference state, i.e., $\x_t = \x_t\tr$. However, the worst-case disturbance happens during interval $[t,\tTs]$. This causes the state to exceed the constraint boundary. Suddenly, the problem has become infeasible, and there is no structured way to recover, i.e., there does not exist a suitable candidate solution that could be applied in an open loop. From the analysis above, we can conclude that if the reference trajectory is generated far enough away from the constraints, i.e., the tightening $\cjs \alpha, j \in \Ns$ or $\co \alpha$ is sufficiently large, we can prevent the occurrence of this problem. This is the exact reason why the experiments on the real robot using the TMPC scheme presented in \cite{benders2025embedded} were successful despite explicit robust design: the distance $\dmax$ to the obstacle avoidance constraints was tuned sufficiently well to avoid the obstacles and make sure the system constraints were satisfied at all times in the presence of disturbances.

This is not the complete story, though. Consider the scenario where the reference trajectory is generated far enough from the constraint boundaries. And let's again consider the positive scenario where $\x_t = \x_t\tr$. At $t$, the TMPC would make a nominal prediction according to dynamics \eqref{eq:xdot_nom} and execute the control law in a closed loop. However, during interval $[t,t+\Ts]$, the worst-case disturbance impacts the system. So, instead of continuing to track the reference, the state moves away from the reference. The TMPC did not anticipate this. At time $t+\Ts$, the TMPC optimizes a nominal predicted trajectory again from the last received sample of the disturbed state. By executing the optimal input during the interval $[t+\Ts,t+2\Ts]$, the system will move towards the reference trajectory again. Thus, even without explicitly considering disturbances, the TMPC might be able to track \eqref{eq:r}. Of course, the tracking is not perfect, but at least the system stays close to the reference trajectory. This feedback property of TMPC is called inherent robustness \cite{yu2014inherent}. It is a property that most practical MPC implementations rely on, and that might be sufficient to track the reference trajectory.

Inherent robustness alone might not be sufficient to ensure the robot's safety. For example, another disturbance may impact the system during $[t+\Ts,t+2\Ts]$ and, despite the optimal control effort of the MPC, moves the system further away from the reference. In this case, the closed-loop system is called unstable. In other words, the MPC is not `strong' enough to attenuate the impact of the disturbances. This notion of strongness is a property that is commonly called contractivity. In other words, the difference between the actual and the reference trajectory should contract with a minimum rate. How close the actual system comes to the reference depends on the contraction rate, usually denoted by $\rho$, and the worst-case impact of the disturbances, usually denoted by $\bw$. These are the quantities you can also find in RMPC formulation \eqref{eq:rmpc}.

A common approach to improve the disturbance attenuation capability of the MPC and ensure that the closed-loop system stays close to the reference trajectory is to design a feedback law executed at a higher frequency than the sampling frequency of the MPC. This way, the impact of disturbances can be diminished before the MPC can do so at the discrete sampling moments. Similar to the design of the terminal control law \eqref{eq:tmpc_kappaf} that renders terminal set \eqref{eq:tmpc_xf} invariant, one could design the feedback law to render a so-called tube around the reference trajectory invariant to the disturbances. This forms the basis for a popular class of RMPC methods: tube MPC \cite{mayne2011tube,bayer2013discrete,zhao2022tube,singh2023robust}.

Tube MPC is designed to tighten all the constraints by a rigid tube size such that the closed-loop system satisfies the actual constraints if the nominal predicted trajectory satisfies the tightened constraints. Specifically, in tube MPC, the complete predicted trajectory is optimized, including initial state constraint \eqref{eq:rmpc_z0}. This increases computational complexity but gives a trivial expression for the terminal set \cite{kohler2024advanced}.

Central to the design of tube MPC schemes is to find a suitable expression for the tube. A common approach is to design a feedback law $\kappadxz$ that renders a sublevel set of incremental Lyapunov function $\Vdxz$ invariant to the disturbances acting on the system actuated by closed-loop control law \eqref{eq:rmpc_cl}. Such combination of feedback law in $\kappadxz$ and $\Vdxz$ exists if the system satisfies the following assumption \cite{angeli2002lyapunov,nubert2020safe}:
\begin{assumption}[Incremental stabilizability]\label{ass:rmpc_inc_stab}
    There exist a feedback law $\kappadxz : \kappaddef$, an incremental Lyapunov function $\Vdxz : \Vddef$ that is continuously differentiable and satisfies $\Vd(\zz)=0, \forall \z \in \X$, parameters $\cdl, \cdu, \rho > 0$, and Lipschitz constants $\cjs > 0, j \in \Ns$ and $\co > 0$, such that the following properties hold for all $(\xzv)\in\XtZ$:
    \begin{subequations}\label{eq:ass_inc_stab}
        \begin{alignat}{2}
            &\cdl \norm{\x-\z}^2 \leq \Vdxz \leq \cdu \norm{\x-\z}^2,&&\label{eq:ass_inc_stab_vd_bounds}\\
            &\gjs(\x,\kappaxzv) - \gjs(\zv) \leq \cjs \sqrt{\Vdxz},\ &&j \in \Ns,\label{eq:ass_inc_stab_cjs}\\
            &\go(M\x) - \go(M\z) \leq \co \sqrt{\Vdxz},&&\label{eq:ass_inc_stab_co}\\
            &\frac{d}{dt}\Vdxz \leq -2\rho \Vdxz,&&\label{eq:ass_inc_stab_vd_contract}
        \end{alignat}
    \end{subequations}
    with $\dx = f(\x,\kappaxzv) + E\wb$ and $\dz = f(\zv) + E\wb$. Furthermore, the following norm-like inequality holds $\forall \x_1, \x_2, \x_3 \in \Rx$:
    \begin{equation}\label{eq:ass_inc_stab_triangle_ineq}
        \sqrt{\Vd(\x_1,\x_3)} \leq \sqrt{\Vd(\x_1,\x_2)} + \sqrt{\Vd(\x_2,\x_3)}.
    \end{equation}
\end{assumption}
In other words, we want to find a squared incremental Lyapunov function $\Vdxz$ that quantifies the difference between disturbed trajectory $\x$ and nominal, or reference, trajectory $\z$. $\Vdxz$ is lower- and upper-bounded by \eqref{eq:ass_inc_stab_vd_bounds}, that contracts at least with rate $2\rho$ according to \eqref{eq:ass_inc_stab_vd_contract}, and satisfies norm-like inequality \eqref{eq:ass_inc_stab_triangle_ineq}, such that the system and obstacle avoidance constraints are Lipschitz-bounded by a constant multiplied with the $\opsqrt$ of $\Vdxz$. The meaning and purpose of the above equations will become more clear later in this section.

\begin{remark}
    The superscript $\delta$ denotes the incremental property of the Lyapunov function, i.e., the fact that the Lyapunov function describes the difference between two time-varying trajectories $x$ and $z$ that are close to each other.
\end{remark}

\begin{remark}
    When comparing the expressions in Assumption~\ref{ass:rmpc_inc_stab} with \cite[Ass. 1]{kohler2018nonlinear}, note that $\Vd$ contains a different number of arguments. In general, one can construct $\Vd$ that also depends on the nominal input $\bv$. However, here, we do not leverage this property and leave the additional argument out for notational convenience.
\end{remark}

Thus, if we can upper bound $\sqrt{\Vdxz}$, we know that we can tighten the system and obstacle avoidance constraints using \eqref{eq:ass_inc_stab_cjs} and \eqref{eq:ass_inc_stab_co} to ensure that the open-loop execution of control law \eqref{eq:rmpc_cl} guarantees that the real system trajectory $\x$ satisfies the non-tightened constraints.

Note that applying the $\opsqrt$ operator to \eqref{eq:ass_inc_stab_vd_contract} gives the following upper bound on the derivative of $\sqrt{\Vdxz}$:
\begin{equation}\label{eq:vd_sqrt}
    \frac{d}{dt}\sqrt{\Vdxz} \leq -\rho \sqrt{\Vdxz}.
\end{equation}
This upper bound evolves for $t, \tau \geq 0$ as
\begin{equation}\label{eq:vd_sqrt_t}
    \sqrt{\Vd(\x_{\tpt},\z_{\tpt})} \leq e^{-\rho \tau} \sqrt{\Vd(\x_t,\z_t)}.
\end{equation}
In other words, if the trajectories $\x$ and $\z$ deviate from each other, they will contract over time according to \eqref{eq:vd_sqrt_t}. Correspondingly, the error
\begin{equation}\label{eq:rmpc_bdelta}
    \bdelta \coloneqq \x-\z
\end{equation}
exponentially converges to zero. As mentioned above, this is not the case if disturbances act on the system. Therefore, we would like to find an expression for \eqref{eq:vd_sqrt} that involves the impact of disturbances $\w$. To this end, let's first define the following property that holds with the definition of disturbance set \eqref{eq:w_bb}:
\begin{property}\label{property:w}
    There exists a constant $\bw > 0$ such that for any $(\x,\z,\bv) \in \X \times \Z$ and any $\w \in \wb\oplus\W$, it holds that
    \begin{equation}\label{eq:vd_sqrt_w}
        \frac{d}{dt}\sqrt{\Vdxz} \leq -\rho \sqrt{\Vdxz} + \bw,
    \end{equation}
    with $\dx = f(\x,\kappaxzv) + E\w$ and $\dz = f(\zv) + E\wb$.
\end{property}
Consequently, if we denote the upper bound of $\sqrt{\Vdxz}$, also called the tube size, by $s$, i.e.,
\begin{equation}\label{eq:tube}
    \sqrt{\Vdxz} \leq s,
\end{equation}
we can write the time-derivative of $s$ over time as
\begin{equation}\label{eq:sdot}
    \ds = -\rho s + \bw.
\end{equation}
Rewriting \eqref{eq:sdot} as a function of time gives
\begin{equation}\label{eq:rmpc_s_t}
    s_t = \left(1-e^{-\rho t}\right)\frac{\bw}{\rho},
\end{equation}
and taking the limit $t \rightarrow \infty$ results in a maximum tube size of
\begin{equation}\label{eq:rmpc_sbar}
    \bs = \frac{\bw}{\rho}.
\end{equation}
Thus, $s$ does not grow unbounded, and the resulting tube size \eqref{eq:rmpc_sbar} is suitable for tightening the constraints in a tube MPC formulation.

In this work, we do not leverage tube MPC, but a slightly different variant: constraint tightening MPC \cite{marruedo2002input,villanueva2017robust,kohler2018novel,pin2009robust}. This is the formulation used in \eqref{eq:rmpc}. The idea behind constraint tightening MPC is that at the current time $t$, in contrast to tube MPC, the initial predicted state $\z_{\0t}$ equals the measured state $\x_t$ \eqref{eq:rmpc_z0} and the initial tube size is set to zero \eqref{eq:rmpc_s0}. From there, given a proper design of closed-loop control law \eqref{eq:rmpc_cl}, the predicted uncertainty grows according to \eqref{eq:sdot}, which is reflected in \eqref{eq:rmpc_sdot}, with contraction rate $\rho$ and worst-case disturbance impact $\bw$. The growing tube size is used to tighten the system constraints in \eqref{eq:rmpc_sys}, using system constraints tightening constants $\cjs, j \in \Ns$, and obstacle avoidance constraints \eqref{eq:rmpc_obs}, using obstacle avoidance constraints tightening constant $\co$. Since the tube size does not grow unbounded as given in \eqref{eq:rmpc_sbar}, $\bs$ can be used to construct terminal set constraint \eqref{eq:rmpc_term} such that the system stays within the terminal set for all future times given a suitable terminal control law.

\begin{remark}
    As will be shown later, $\bs$ can be used to tighten the constraints of the reference trajectory, similar to the usage of the terminal set scaling $\alpha$ for tightening the constraints of the reference trajectory in the TMPC section.
\end{remark}

Similar to Section~\ref{sec:tmpc_properties}, we need an assumption on the terminal ingredients such that the following terminal cost property is satisfied:
\begin{assumption}[Terminal ingredients]\label{ass:rmpc_term_ing}
    There exist a terminal control law $\kappafxr : \XtbZ \rightarrow \Ru$ and terminal cost $\Jf(\z,\xr) : \XtbX \rightarrow \R$ such that the following property holds for all $\z \in \X$:
    \begin{equation}\label{eq:rmpc_ass_term_jf}
        \Jf(\z_{\TTt}^*,\x_{\tpTsT}\tr) - \Jf(\z_{\Tt}^*,\x_{\tpT}\tr) \leq -\int_{T}^{T+\Ts}\Js(\z_{\tt}^*,\bv_{\tt}^*,\r_{\tpt}) d\tau,
    \end{equation}
    with sampling time $\Ts > 0$, optimal solution $(\z_{\ct}^*,\bv_{\ct}^*)$ at $t$ for all $t \geq 0$ and its extended version $\z_{\Tpt|t}^*$ using terminal control law $\kappafxr$ for $\tau \in [0,\Ts]$, as defined in Section~\ref{sec:rmpc_proof_candidate}.
\end{assumption}
Next to this assumption on terminal control law $\kappafxr$ and terminal cost $\Jf(\z,\xr)$ we want to design terminal set $\Xf(\xr)$ such that tightened system and obstacle avoidance constraints \eqref{eq:rmpc_sys} and \eqref{eq:rmpc_obs} are satisfied in $\Xf(\xr)$, respectively, and such that terminal control law $\kappafxr$ ensures robust positive invariance of $\Xf(\xr)$. The specific formulation of a tight version of $\Xf(\xr)$ is presented in Section~\ref{sec:rmpc_proof_rec_feas} as part of the proof and therefore not included in Assumption~\ref{ass:rmpc_term_ing}.

\begin{figure}[h!]
    \centering
    \includesvg[width=\textwidth]{rmpc_proofs.svg}
    \caption{Visualization of a 1D reference trajectory $\x_{t+[0,4\Ts]}\tr$, previously optimal solution $\z_{[0,T]|t}^*$, appended part of previously optimal solution $\z_{[T,T+\Ts]|t}^*$, candidate solution $\tz_{[0,T]|\tTs}$, tube $\Vd(\x,\z^*) \leq s$ around $\z_{[0,T]|t}^*$, tube $\Vd(\x,\tz) \leq s$ around $\tz_{[0,T]|\tTs}$, and terminal sets $\Xf(\xr)$ around the reference trajectory corresponding to the previously optimal and the candidate solution. Again, we set $T=4\Ts$. Compared to Figure~\ref{fig:tmpc}, $\w$ impacts the system during interval \circled{1}. As a result, $\tz_{0|\tTs} \neq \z_{\Ts|t}^*$. Note that the tubes and terminal sets are visualized as ellipses to clarify their shapes, whereas they should actually be visualized as vertical lines in this case.}
    \label{fig:rmpc}
\end{figure}

The assumptions above allow us to prove recursive feasibility and trajectory tracking. To provide intuition for these proofs beforehand, similar to Section~\ref{sec:tmpc_properties}, Figure~\ref{fig:rmpc} shows the optimal solution at time $t$ and the corresponding candidate solution for time $\tTs$. Note that, in contrast to the TMPC case, we cannot just take the tail part of the previously optimal solution as a candidate and append it for prediction interval \circled{4} since the system is subject to disturbances $\w$ during interval \circled{1}. Instead, we measure the disturbed state at $\tTs$ and leverage the suitably designed feedback term in the control law to steer the candidate back to the previously optimal trajectory. Since the tube size, or equivalently, the constraints tightening in \eqref{eq:rmpc_sys} and \eqref{eq:rmpc_obs}, increases over the horizon, the constraints in interval \circled{2} of the candidate are tightened less than the constraints in the same interval of the previously optimal solution. The visual interpretation is that the blue tube is always contained in the orange tube. As a result, it can be shown that the impact of the disturbance during interval \circled{1} sufficiently contracts by the feedback term in the candidate control law such that the candidate also satisfies the tightened constraints in interval \circled{2}. The same holds for a suitably designed terminal set. Furthermore, if we append the candidate with terminal control law $\kappafxr$ in interval \circled{4}, where the nominal term equals $\ur$ and the feedback term steers the candidate towards $\xr$, it can be shown that a suitably designed terminal set is invariant to this control law. Therefore, the candidate also satisfies \eqref{eq:rmpc_term}. If the system and obstacle constraints are also satisfied in the terminal set, the satisfaction of \eqref{eq:rmpc_sys} and \eqref{eq:rmpc_obs} in interval \circled{4} trivially follows from the invariance property of the terminal set. Thus, a feasible solution at $t$ implies that a feasible solution at $\tTs$ can be constructed. Therefore, the scheme is recursively feasible.

The trajectory tracking proof entails showing that cost \eqref{eq:rmpc_obj} is lower for the candidate than for the previously optimal solution such that convergence to the reference trajectory can be concluded. In the case of TMPC, we could show that cost \circled{4} is lower than terminal cost \circled{3} such that the candidate cost decreases at least by cost \circled{1} since cost \circled{2} is the same. However, cost \circled{2} is not the same. In fact, cost \circled{2} and cost \circled{4} both increase depending on the worst-case disturbance impact during interval \circled{1}. Therefore, strict convergence to the reference trajectory cannot be concluded. Instead, we can prove that the average tracking error gets small.

In summary, showing that RMPC formulation \eqref{eq:rmpc} ensures safe and efficient mobile robot deployment in the presence of disturbances requires the robot system to be incrementally stabilizable according to Assumption~\ref{ass:rmpc_inc_stab}. This means that a feedback controller can be constructed to guarantee that the tracking error of the closed-loop system with respect to the reference trajectory can be quantified and upper-bounded. Furthermore, if the terminal ingredients are suitably designed satisfying Assumption~\ref{ass:rmpc_term_ing} and the candidate is designed as described in Section~\ref{sec:rmpc_proof_candidate}, we can prove recursive feasibility in Section~\ref{sec:rmpc_proof_rec_feas} and trajectory tracking with, on average, an error that gets small in Section~\ref{sec:rmpc_proof_tracking}. The details of the incremental stabilizability and terminal ingredients design is provided in Section~\ref{sec:rmpc_properties_design}.

\subsubsection{Candidate solution}\label{sec:rmpc_proof_candidate}
Similar to Section~\ref{sec:tmpc_proof_candidate}, the goal of defining the candidate solution is to show that there exists a feasible, not necessarily optimal, solution to \eqref{eq:rmpc} at time $\tTs$ given that \eqref{eq:rmpc} is successfully solved at time $t$. Again, this candidate solution is used to prove recursive feasibility and trajectory tracking and is not necessarily implemented in practice.

In the robust case, we pick a candidate solution similar to the nominal case. For convenience, we define for $\tau \in [T,T+\Ts]$:
\begin{equation}\label{eq:rmpc_u_append}
    \bv_{\tt}^* = \kappaf(\z_{\tt}^*,\r_{\tpt}),
\end{equation}
with $\dz_{\tt}^*$ given by \eqref{eq:xdot_nom} by applying \eqref{eq:rmpc_u_append}.

Given these definitions, we can write the candidate for prediction interval $\tau \in [0,T]$ at time $\tTs$ as
\begin{subequations}\label{eq:rmpc_candidate}
    \begin{align}
        \tz_{0|\tTs} &= \x_{\tTs},\label{eq:rmpc_candidate_z0}\\
        \ts_0 &= 0,\label{eq:rmpc_candidate_s0}\\
        \tv_{\ttT} &= \kappa(\tz_{\ttT},\z_{\tTt}^*,\bv_{\tTt}^*),\label{eq:rmpc_candidate_v}
    \end{align}
\end{subequations}
with $\dtz_{\ttT}$ and $\dts_\tau$ according to \eqref{eq:rmpc_zdot} and \eqref{eq:rmpc_sdot}, respectively.

Note that this candidate deviates from the one in Section~\ref{sec:tmpc_proof_candidate}. The main difference is that we cannot take the previously optimal state-input trajectory since the state at $\tTs$ is disturbed compared to prediction $\tau=\Ts$ at $t$. Therefore, we need to measure the state at $\tTs$ and use this as the initial state of the candidate solution. Consequently, following the previously optimal input trajectory $\bv_{\tTt}^*, \tau \in [0,T]$ from a different state results in a different trajectory. Thus, this solution cannot be directly used. Instead, we add the feedback law $\kappad(\x_{\ttT},\z_{\tTt}^*), \tau \in [0,T]$ described in Assumption~\ref{ass:rmpc_inc_stab}.

\subsubsection{Recursive feasibility proof}\label{sec:rmpc_proof_rec_feas}
The recursive feasibility proof follows similar arguments as the proof in Section~\ref{sec:tmpc_proof_rec_feas} with some additional technicalities to account for the disturbances. As mentioned, the first difference is in the satisfaction of \eqref{eq:rmpc_z0}. Whereas in Section~\ref{sec:tmpc_proof_rec_feas}, we could just take the previously optimal solution, in this case, the state at $\tTs$ does not equal the nominal predicted one at predicted time $\Ts$ at time $t$ because of the disturbance effect $\w_\tau \neq \bm{0}, \tau \in [t,\tTs]$. Therefore, we satisfy this constraint by measuring the disturbed state at $\tTs$ as given by \eqref{eq:rmpc_candidate_z0} of the candidate.

The initial state constraint, initial tube constraint, system dynamics constraint, and tube dynamics constraint are all trivially satisfied by \eqref{eq:rmpc_candidate}. That leaves us to show that the (tightened) system constraints \eqref{eq:rmpc_sys}, the (tightened) obstacle avoidance constraints \eqref{eq:rmpc_obs}, and the terminal set constraint \eqref{eq:rmpc_term} are satisfied in the interval $\tau\in[0,T]$.

Similar to the procedure in Section~\ref{sec:tmpc_proof_rec_feas}, we split the interval into two subintervals, $\tau\in[0,T-\Ts)$ and $\tau\in[T-\Ts,T]$, to show satisfaction of \eqref{eq:rmpc_sys} and \eqref{eq:rmpc_obs}.\\

\noindent\textbf{System constraints satisfaction for $\bm{\tau\in[0,T-\Ts)}$}\\
Let's start with the proofs for the interval $\tau\in[0,T-\Ts)$. For this interval, \eqref{eq:rmpc_sys} are satisfied by candidate \eqref{eq:rmpc_candidate} if the following holds for $\tau \in [0,T-\Ts), j \in \Ns$:
\begin{equation}\label{eq:rmpc_gjs_tTs}
    \gjs(\tz_{\ttT},\tv_{\ttT}) + \cjs s_\tau \leq 0.
\end{equation}

We know that for $\tau \in [0,T-\Ts), j \in \Ns$ the previously optimal solution satisfies
\begin{equation}\label{eq:rmpc_gjs_t}
    \gjs(\z_{\tTt}^*,\bv_{\tTt}^*) + \cjs s_\Tstau \leq 0.
\end{equation}
Therefore, we would like to create an upper bound on $\gjs(\tz_{\ttT},\tv_{\ttT})$ in terms of $\gjs(\z_{\tTt}^*,\bv_{\tTt}^*)$. This can be done using the Lipschitz assumption \eqref{eq:ass_inc_stab_cjs} under control law $\kappa(\tz_{\ttT},\z_{\tTt}^*,\bv_{\tTt}^*)$. This gives the following upper bound for $j \in \Ns$:
\begin{equation}\label{eq:rmpc_gjs_ub}
    \gjs(\tz_{\ttT},\tv_{\ttT}) \leq \gjs(\z_{\tTt}^*,\bv_{\tTt}^*) + \cjs \sqrt{\Vd(\tz_{\ttT},\z_{\tTt}^*)}.
\end{equation}
Filling in \eqref{eq:rmpc_gjs_ub} in the left-hand side of \eqref{eq:rmpc_gjs_tTs} gives for $j \in \Ns$:
\begin{equation}
    \gjs(\tz_{\ttT},\tv_{\ttT}) + \cjs s_\tau \leq \gjs(\z_{\tTt}^*,\bv_{\tTt}^*) + \cjs \sqrt{\Vd(\tz_{\ttT},\z_{\tTt}^*)} + \cjs s_\tau.
\end{equation}
Thus, a sufficient condition to satisfy \eqref{eq:rmpc_gjs_tTs} for $j \in \Ns$ is the following:
\begin{equation}\label{eq:rmpc_gjs_t_ub}
    \gjs(\z_{\tTt}^*,\bv_{\tTt}^*) + \cjs \sqrt{\Vd(\tz_{\ttT},\z_{\tTt}^*)} + \cjs s_\tau \leq 0.
\end{equation}

We know that \eqref{eq:rmpc_gjs_t} is satisfied by solving \eqref{eq:rmpc} at $t$. Therefore, to show that \eqref{eq:rmpc_gjs_t_ub} holds, we need to show that the following inequality holds for $j \in \Ns$:
\begin{equation}\label{eq:rmpc_cjs_ub}
    \cjs \sqrt{\Vd(\tz_{\ttT},\z_{\tTt}^*)} + \cjs s_\tau \leq \cjs s_\Tstau,
\end{equation}
Dividing the left-hand and right-hand sides by $\cjs > 0, j \in \Ns$ gives
\begin{equation}\label{eq:rmpc_vd_s_ub}
    \sqrt{\Vd(\tz_{\ttT},\z_{\tTt}^*)} + s_\tau \leq s_\Tstau.
\end{equation}

We know that $\Vdxz$ is upper-bounded by tube size $s$ according to \eqref{eq:tube} and that the evolution of $s$ over time is given by \eqref{eq:rmpc_s_t}. Thus, after $\Ts$, the tube has the following size:
\begin{equation}\label{eq:s_ts}
    s_\Ts = \left(1-e^{\rho \Ts}\right)\frac{\bw}{\rho},
\end{equation}
and we can write the following upper bound on $\sqrt{\Vd(\tz_{0|\tTs},\z_{\Tst}^*)}$ at $\tTs$:
\begin{equation}\label{eq:vd_bound_s}
    \sqrt{\Vd(\tz_{0|\tTs},\z_{\Tst}^*)} \leq s_\Ts.
\end{equation}
By applying control law $\kappa(\tz_{\ttT},\z_{\tTt}^*,\bv_{\tTt}^*)$, to the system, we can show that the candidate response $\tz_{\ttT}$ will contract to the previously optimal nominal prediction $\z_{\tTt}^*$ with factor $\rho$ over time according to \eqref{eq:vd_sqrt_t}. Thus, by iteratively applying the contractivity, we can write the following upper bound for $\sqrt{\Vd(\tz_{\ttT},\z_{\tTt}^*)}$:
\begin{equation}\label{eq:vd_bound_s_tau}
    \sqrt{\Vd(\tz_{\ttT},\z_{\tTt}^*)} \leq e^{-\rho \tau} s_\Ts.
\end{equation}
Filling in \eqref{eq:vd_bound_s_tau} in \eqref{eq:rmpc_vd_s_ub} yields the following sufficient condition:
\begin{equation}
    e^{-\rho \tau} s_\Ts + s_\tau \leq s_\Tstau,
\end{equation}
or, alternatively,
\begin{equation}\label{eq:s_bound}
    s_\tau \leq s_\Tstau - e^{-\rho \tau} s_\Ts.
\end{equation}

Appendix~\ref{app:s_bound} shows that \eqref{eq:s_bound} holds for $\tau \geq 0$, which includes $\tau \in [0,T]$. Therefore, the system constraints are satisfied for $\tau\in[0,T-\Ts]$ and thus for $\tau\in[0,T-\Ts)$. Note that we cannot prove satisfaction for $\tau\in[0,T]$ since this reasoning relies on the fact that \eqref{eq:rmpc_gjs_t} holds, which is only enforced for $\tau\in[0,T]$ at $t$, and thus for $\tau\in[0,T-\Ts]$ at $\tTs$.\\

\noindent\textbf{Obstacle avoidance constraints satisfaction for $\bm{\tau\in[0,T-\Ts)}$}\\
For interval $\tau\in[0,T-\Ts)$, we know that \eqref{eq:rmpc_obs} are satisfied by candidate \eqref{eq:rmpc_candidate} if the following holds for $\tau \in [0,T-\Ts), j \in \No$:
\begin{equation}\label{eq:rmpc_gjo_tTs}
    \gjo(M\tz_{\ttT}) + \co s_\tau \leq 0.
\end{equation}

We know that for $\tau \in [0,T-\Ts), j \in \No$ the previously optimal solution satisfies
\begin{equation}
    \gjo(M\z_{\tTt}^*) + \co s_\Tstau \leq 0.
\end{equation}
Therefore, similar to the system constraints proof for $\tau\in[0,T-\Ts)$, we would like to create an upper bound on $\gjo(M\tz_{\ttT})$ in terms of $\gjo(M\z_{\tTt}^*)$. Lipschitz assumption \eqref{eq:ass_inc_stab_co} gives the following upper bound for $j \in \No$:
\begin{equation}
    \gjo(M\tz_{\ttT}) \leq \gjo(M\z_{\tTt}^*) + \co \sqrt{\Vd(\tz_{\ttT},\z_{\tTt}^*)}.
\end{equation}

By following the same steps as for the system constraints, we can show that the proof boils down to showing that \eqref{eq:s_bound} holds, which is true by the combination of the proof in Appendix~\ref{app:s_bound} and Assumption~\ref{ass:obs} from Section~\ref{sec:tmpc_proof_rec_feas}. Thus, the obstacle avoidance constraints are also satisfied for $\tau\in[0,T-\Ts)$.\\

\noindent\textbf{Terminal set constraint satisfaction for $\bm{\tau\in[T-\Ts,T]}$}\\
Next, we prove that if \eqref{eq:rmpc_term} is satisfied at time $t$, candidate \eqref{eq:rmpc_candidate} satisfies \eqref{eq:rmpc_term} in prediction interval $\tau\in[T-\Ts,T]$ at time $\tTs$. Before proving this, we need to know a suitable expression for the terminal set. In the TMPC case, $\Xf(\xr)$ was chosen as a sublevel set of $\Jf(\z,\xr)$. If we use this set in the robust case, it means that candidate \eqref{eq:rmpc_candidate} should satisfy this constraint for $\tau\in[T-\Ts,T]$ at $\tTs$. As discussed, the minimum terminal set scaling $\alpha$ could be chosen equal to 0 in the case of TMPC. However, even if the previously optimal solution $\z_{[0,T]|t}^*$ would exactly overlap with reference $\x_{t+[0,T]}\tr$, the disturbance impact during interval $[t,\tTs]$ renders terminal set constraint \eqref{eq:rmpc_term} infeasible for candidate \eqref{eq:rmpc_candidate} since we only have a finite horizon and exponential contraction of the disturbance impact as given by \eqref{eq:vd_bound_s_tau}. In other words, candidate $\tz_{[T-\Ts,T]|\tTs}$ deviates from previously optimal solution $\z_{[T,T+\Ts]|t}^*$. Therefore, we need another expression for the minimum terminal set that accounts for the disturbance impact during $[t,\tTs]$.

Besides the mentioned disturbance impact, we also need to account for the fact that it is likely to happen that the previously optimal solution does not overlap with the reference for $\tau\in[T,T+\Ts]$, i.e., $\z_{[T,T+\Ts]|t}^* \neq \x_{t+[T,T+\Ts]}\tr$, on a disturbed system, even if such a system has exponential disturbance rejection properties by closed-loop terminal control law $\kappafxr$.

In summary, to show that candidate \eqref{eq:rmpc_candidate} satisfies terminal set constraint \eqref{eq:rmpc_term} for $\tau\in[T-\Ts,T]$ at time $\tTs$, we need a terminal set that accounts for both the error between previously optimal solution $\z_{[T,T+\Ts]|t}^*$ and reference $\x_{t+[T,T+\Ts]}\tr$ and the error between candidate $\tz_{[T-\Ts,T]|\tTs}$ and previously optimal solution $\z_{[T,T+\Ts]|t}^*$.

The error between $\z_{[T,T+\Ts]|t}^*$ and $\x_{t+[T,T+\Ts]}\tr$ contracts with rate $\rho$ according to \eqref{eq:vd_sqrt_t}. For $\tau \in [0,\Ts]$ it is given by
\begin{equation}
    \sqrt{\Vd(\z_{T+\tau|t}^*,\x_{t+T+\tau}\tr)} \leq e^{-\rho \tau} \sqrt{\Vd(\z_{\Tt}^*,\x_{\Tt}\tr)}.
\end{equation}

The error between $\tz_{[T-\Ts,T]|\tTs}$ and $\z_{[T,T+\Ts]|t}^*$ is caused by the described disturbance impact during $[t,\tTs]$ and also contracts with rate $\rho$ according to the following expression for $\tau \in [0,\Ts]$:
\begin{align}
    \sqrt{\Vd(\tz_{T-\Ts+\tau|\tTs},\z_{T+\tau|t}^*)} \overset{\eqref{eq:vd_bound_s_tau}}{\leq} &e^{-\rho (T-\Ts+\tau)} s_\Ts.
\end{align}

We can construct an upper bound on the total tracking error of the candidate using \eqref{eq:ass_inc_stab_triangle_ineq} in Assumption~\ref{ass:rmpc_inc_stab}. For $\tau \in [0,\Ts]$ this yields
\begin{equation}
    \sqrt{\Vd(\tz_{T-\Ts+\tau|\tTs},\x_{t+T+\tau}\tr)} \leq \sqrt{\Vd(\z_{T+\tau|t}^*,\x_{t+T+\tau}\tr)} + \sqrt{\Vd(\tz_{T-\Ts+\tau|\tTs},\z_{T+\tau|t}^*)}.
\end{equation}
Therefore, we can write the upper bound on the total terminal tracking error for $\tau \in [0,\Ts]$ as
\begin{equation}\label{eq:rmpc_term_err_ub}
    \sqrt{\Vd(\tz_{T-\Ts+\tau|\tTs},\x_{t+T+\tau}\tr)} \leq e^{-\rho \tau} \sqrt{\Vd(\z_{\Tt}^*,\x_{\tpT}\tr)} + e^{-\rho (T-\Ts+\tau)} s_\Ts.
\end{equation}
To prove that the candidate satisfies the terminal set constraint given this inequality, we leverage the expression of the terminal set proposed in \cite{nubert2020safe}:
\begin{equation}\label{eq:rmpc_xf}
    \Xf(\xr) \coloneqq \left\{\z \in \X, s_T \in \R\ \middle|\ \sqrt{\Vd(\z,\xr)} + s_T \leq \alpha\right\},
\end{equation}
with $s_T$ representing the tube size evaluated at $T$ according to \eqref{eq:rmpc_s_t} and $\alpha$ denoting the terminal set scaling, similar to the definition of $\alpha$ in \eqref{eq:tmpc_xf}. Note that the terminal set is expressed in non-squared form, so we use notation $\alpha$ instead of $\alpha^2$.

Extending this terminal set to prediction interval $\tau \in [0,\Ts]$ gives
\begin{equation}\label{eq:rmpc_xf_ext}
    \Xf(\x_{\tpTt}\tr) \coloneqq \left\{\z_{\Tpt|t} \in \X, s_{\Tpt} \in \R\ \middle|\ \sqrt{\Vd(\z_{\Tpt|t},\x_{\tpTt}\tr)} + s_{\Tpt} \leq \alpha\right\}.
\end{equation}

We know that pair $(\z_{\Tt}^*,s_T)$ satisfies terminal set constraint \eqref{eq:rmpc_xf}. Therefore, it holds that
\begin{equation}
    \sqrt{\Vd(\z_{\Tt}^*,\x_{\tpT}\tr)} \leq \alpha - s_T,
\end{equation}
which leads to the following expression for \eqref{eq:rmpc_term_err_ub} for $\tau \in [0,\Ts]$:
\begin{equation}\label{eq:rmpc_term_err_ub_alpha}
    \sqrt{\Vd(\tz_{T-\Ts+\tau|\tTs},\x_{t+T+\tau}\tr)} \leq e^{-\rho \tau} (\alpha - s_T) + e^{-\rho (T-\Ts+\tau)} s_\Ts.
\end{equation}

We can use this expression to derive a minimum value for terminal set scaling $\alpha$ such that $\tz_{T-\Ts+\tau|\tTs}$ satisfies the terminal set constraint for $\tau \in [0,\Ts]$, i.e.,
\begin{equation}\label{eq:rmpc_term_cand_ub_alpha}
    \sqrt{\Vd(\tz_{T-\Ts+\tau|\tTs},\x_{t+T+\tau}\tr)} \leq \alpha - s_{T-\Ts+\tau}.
\end{equation}
This inequality is satisfied if the addition of the two errors described above, i.e., the right-hand side of \eqref{eq:rmpc_term_err_ub_alpha}, is upper-bounded by the maximum error allowed in the terminal set, i.e., the right-hand side of \eqref{eq:rmpc_term_cand_ub_alpha}:
\begin{equation}
    e^{-\rho \tau}(\alpha - s_T) + e^{-\rho (T-\Ts+\tau)}s_\Ts \leq \alpha-s_{T-\Ts+\tau},
\end{equation}
which trivially holds for $\tau=0$ by Appendix~\ref{app:s_bound}. Working out this inequality for $\tau \in (0,\Ts]$ gives the following lower bound on $\alpha$:
\begin{subequations}\label{eq:rmpc_term_inv_der}
    \begin{alignat}{2}
        e^{-\rho \tau}(\alpha - s_T) + e^{-\rho (T-\Ts+\tau)}s_\Ts &\overset{\text{(a)}}{\leq} &&\alpha-s_{T-\Ts+\tau}\label{eq:rmpc_term_inv_der_0}\\
        \left(1-e^{-\rho \tau}\right)\alpha + e^{-\rho \tau}s_T - s_{T-\Ts+\tau} &\overset{\text{(b)}}{\geq} &&e^{-\rho (T-\Ts+\tau)} s_\Ts\\
        \left(1-e^{-\rho \tau}\right)\alpha + e^{-\rho \tau}\left(1-e^{-\rho T}\right)\frac{\bw}{\rho} - \left(1-e^{-\rho (T-\Ts+\tau)}\right)\frac{\bw}{\rho} &\overset{\text{(c)}}{\geq} &&e^{-\rho (T-\Ts+\tau)} \left(1-e^{-\rho \Ts}\right)\frac{\bw}{\rho}\\
        \left(1-e^{-\rho \tau}\right)\alpha + \left(e^{-\rho \tau}-e^{-\rho (\Tpt)}-1+e^{-\rho (T-\Ts+\tau)}\right)\frac{\bw}{\rho} &\overset{\text{(d)}}{\geq} &&\left(e^{-\rho (T-\Ts+\tau)}-e^{-\rho (\Tpt)}\right)\frac{\bw}{\rho}\\
        \left(1-e^{-\rho \tau}\right)\alpha + \left(e^{-\rho \tau}-1\right)\frac{\bw}{\rho} &\overset{\text{(e)}}{\geq} &&0\\
        \left(1-e^{-\rho \tau}\right)\alpha - \left(1-e^{-\rho \tau}\right)\frac{\bw}{\rho} &\overset{\text{(f)}}{\geq} &&0\\
        \left(1-e^{-\rho \tau}\right)\alpha &\overset{\text{(g)}}{\geq} &&\left(1-e^{-\rho \tau}\right)\frac{\bw}{\rho}\\
        \alpha &\overset{\text{(h)}}{\geq} &&\frac{\bw}{\rho},
    \end{alignat}
\end{subequations}
where (a) is obtained by setting the upper bound of \eqref{eq:rmpc_term_err_ub_alpha} smaller equal to \eqref{eq:rmpc_term_cand_ub_alpha}, (b) by re-ordering terms and combining $\alpha$ multiplication factors, (c) by writing out the tube dynamics according to \eqref{eq:rmpc_s_t}, (d) by combining $\frac{\bw}{\rho}$ multiplication factors, (e) by canceling common terms on both sides and (f)-(h) by trivially working out the inequality.

In other words, $\alpha$ should be at least as large as the maximum tube size given in \eqref{eq:rmpc_sbar}. This makes sense since this means that terminal control law $\kappafxr$ can keep the system within that tube around the reference trajectory, guaranteeing that applying this solution will keep the system safe at all times. Note that $\alpha$ can also be chosen larger than $\frac{\bw}{\rho}$, similar to the discussion in Section~\ref{sec:tmpc_properties_design}. This will result in a larger terminal set, which will render \eqref{eq:rmpc} easier to solve, but it will also increase conservatism in planning the reference trajectory.

\begin{remark}\label{remark:rmpc_term_inv}
    An alternative to the proof above is to show terminal set constraint satisfaction for $\tau\in[T-\Ts,T]$ at $\tTs$ by first proving that candidate $\tz_{T-\Ts|\tTs}$ and corresponding tube $s_{T-\Ts}$ satisfy terminal set constraint \eqref{eq:rmpc_xf} and then leveraging the fact that the terminal set is invariant for $\alpha \geq \frac{\bw}{\rho}$, as shown in Appendix~\ref{app:rmpc_xf_inv}.
\end{remark}

\noindent\textbf{System constraints satisfaction for $\bm{\tau\in[T-\Ts,T]}$}\\
As mentioned before, given that tube inequality \eqref{eq:s_bound} is proven to hold for $\tau\in[0,T]$, the only thing we need to ensure satisfaction of candidate system constraints \eqref{eq:rmpc_gjs_tTs} for $\tau\in[T-\Ts,T]$ is to show that system constraints of previously optimal solution \eqref{eq:rmpc_gjs_t} are satisfied for $\tau\in[T-\Ts,T]$ at $\tTs$, or, equivalently, for $\tau\in[T,T+\Ts]$ at $t$, given that terminal set \eqref{eq:rmpc_xf} is invariant.

Appendix~\ref{app:rmpc_xf_inv} proves that terminal set \eqref{eq:rmpc_xf} is invariant for $(\z_{[T,\Tpt]|t}^*, s_{\Tpt}), \tau \geq 0$ under the same condition $\alpha \geq \frac{\bw}{\rho}$ as given in \eqref{eq:rmpc_term_inv_der}. Given the proof of \eqref{eq:s_bound} from before, candidate system constraints \eqref{eq:rmpc_gjs_tTs} are satisfied for $\tau\in[T-\Ts,T]$ under the following assumption:
\begin{assumption}[System constraints tightening for reference trajectory]\label{ass:rmpc_gjrs}
    The reference trajectory satisfies the tightened systems constraints, i.e., \eqref{eq:r_sys} holds, with $\gjrs(\xu)$ given by
    \begin{equation}\label{eq:ass_rmpc_gjrs}
        \gjrs(\xu) = \gjs(\xu) + \cjs \alpha,\ j \in \Ns,
    \end{equation}
    where $\cjs$ is defined in \eqref{eq:ass_inc_stab_cjs} for $j\in\Ns$ and $\alpha \geq \frac{\bw}{\rho}$.
\end{assumption}
This assumption is similar to the design of \eqref{eq:tmpc_r_sys_ineq} in Section~\ref{sec:tmpc_properties_design}, but with the required lower bound $\alpha \geq \frac{\bw}{\rho}$ to account for the effect of disturbances.\\

\noindent\textbf{Obstacle avoidance constraints satisfaction for $\bm{\tau\in[T-\Ts,T]}$}\\
Proving that candidate obstacle avoidance constraints \eqref{eq:rmpc_gjo_tTs} are satisfied for $\tau\in[T-\Ts,T]$ follows the same arguments as the proof for the system constraints for $\tau\in[T-\Ts,T]$ under the combination of Assumption~\ref{ass:obs} and a similar assumption as above:
\begin{assumption}[Obstacle avoidance constraints tightening for reference trajectory]\label{ass:rmpc_gjro}
    The reference trajectory satisfies the tightened obstacle avoidance constraints, i.e., \eqref{eq:r_obs} holds, with $\gjro(\p)$ given by
    \begin{equation}\label{eq:ass_rmpc_gjro}
        \gjro(\p) = \gjo(\p) + \co \alpha,\ j \in \No,
    \end{equation}
    where $\co$ is defined in \eqref{eq:ass_inc_stab_co} and $\alpha \geq \frac{\bw}{\rho}$.
\end{assumption}
This assumption is similar to the design of \eqref{eq:tmpc_r_obs_ineq} in Section~\ref{sec:tmpc_properties_design} but with the required lower bound $\alpha \geq \frac{\bw}{\rho}$ to account for the effect of disturbances.
\qed

\subsubsection{Trajectory tracking proof}\label{sec:rmpc_proof_tracking}
The trajectory tracking proof in this section follows similar arguments as the one presented in Section~\ref{sec:tmpc_proof_tracking}. For convenience, let's define
\begin{align}
    \J_{\tt}\tops &\coloneqq \Js(\z_{\tt}^*,\bv_{\tt}^*,\r_{\tpt}),\\
    \J_{\Tt}\topf &\coloneqq \Jf(\z_{\Tt}^*,\x_{\tpT}\tr),\\
    \J_t^* &\coloneqq \J_t^*(\x_t,\r_t),\\
    \J_{\ttT}\tops &\coloneqq \Js(\tz_{\ttT},\tv_{\ttT},\r_{\tpTst}),\\
    \J_{\TtT}\topf &\coloneqq \Jf(\tz_{\TtT},\x_{\tpTsT}\tr),\\
    \J_{\tTs}^* &\coloneqq \J_{\tTs}^*(\x_{\tTs},\r_{\tTs}),
\end{align}
and start by following the same proof as in \eqref{eq:tmpc_jopt_db} in Section~\ref{sec:tmpc_proof_tracking}:
\begin{subequations}\label{eq:rmpc_jopt_db}
    \begin{alignat}{2}
        \J_{\tTs}^*
        &\leq &&\int_{0}^{T}\J_{\ttT}\tops d\tau + \J_{\TtT}\topf,\\
        &= &&\int_{0}^{T-\Ts}\J_{\ttT}\tops d\tau + \int_{T-\Ts}^{T}\J_{\ttT}\tops d\tau + \J_{\TtT}\topf.
    \end{alignat}
\end{subequations}
So far, the proof for TMPC and RMPC is the same. However, note that the stage costs are not equal at the same points in time, i.e., $\Js(\tz_{\ttT},\tv_{\ttT},\r_{\tpTst}) \neq \Js(\z_{\tTt}^*,\bv_{\tTt}^*,\r_{\tpTst}), \tau \in [0,T]$, since the candidate $\tz_{\ttT}$ deviates from the previously optimal solution $\z_{\tTt}^*$. Therefore, (c) in \eqref{eq:tmpc_jopt_db} does not hold. Specifically, for $\tau \in [0,T]$, candidate stage cost $\Js(\tz_{\ttT},\tv_{\ttT},\r_{\tpTst})$ is defined as
\begin{equation}
    \Js(\tz_{\ttT},\tv_{\ttT},\r_{\tpTst}) = \norm{\tz_{\ttT}-\x_{\tpTst}\tr}_Q^2 + \norm{\tv_{\ttT}-\bu_{\tpTst}\tr}_R^2,
\end{equation}
and previously optimal stage cost $\Js(\z_{\tTt}^*,\bv_{\tTt}^*,\r_{\tpTst})$ as
\begin{equation}
    \Js(\z_{\tTt}^*,\bv_{\tTt}^*,\r_{\tpTst}) = \norm{\z_{\tTt}^*-\x_{\tpTst}\tr}_Q^2 + \norm{\bv_{\tTt}^*-\bu_{\tpTst}\tr}_R^2.
\end{equation}
Writing out the difference between the two gives
\begin{align}\label{eq:rmpc_js_ub}
    &\Js(\tz_{\ttT},\tv_{\ttT},\r_{\tpTst}) - \Js(\z_{\tTt}^*,\bv_{\tTt}^*,\r_{\tpTst})\notag\\
    &= \norm{\tz_{\ttT}-\x_{\tpTst}\tr}_Q^2 + \norm{\tv_{\ttT}-\bu_{\tpTst}\tr}_R^2 - \norm{\z_{\tTt}^*-\x_{\tpTst}\tr}_Q^2 + \norm{\bv_{\tTt}^*-\bu_{\tpTst}\tr}_R^2\notag\\
    &= \norm{\tz_{\ttT}-\x_{\tpTst}\tr}_Q^2 - \norm{\z_{\tTt}^*-\x_{\tpTst}\tr}_Q^2 + \norm{\tv_{\ttT}-\bu_{\tpTst}\tr}_R^2 - \norm{\bv_{\tTt}^*-\bu_{\tpTst}\tr}_R^2.
\end{align}
Note that this expression is continuously differentiable in compact set $\Z \times \bZ$. Therefore, it is Lipschitz continuous in this set, and the following inequality holds for $\tau \in [0,T]$:
\begin{align}\label{eq:rmpc_js_ub_2}
    &\norm{\tz_{\ttT}-\x_{\tpTst}\tr}_Q^2 - \norm{\z_{\tTt}^*-\x_{\tpTst}\tr}_Q^2 + \norm{\tv_{\ttT}-\bu_{\tpTst}\tr}_R^2 - \norm{\bv_{\tTt}^*-\bu_{\tpTst}\tr}_R^2\notag\\
    &\leq \Ljsx \norm{\tz_{\ttT}-\z_{\tTt}^*} + \Ljsu \norm{\tv_{\ttT}-\bv_{\tTt}^*}.
\end{align}
This result shows that we can find an upper bound on the difference in stage costs.

Ideally, we would like to find the smallest possible value for this upper bound. We can rewrite \eqref{eq:ass_inc_stab_vd_bounds} in Assumption~\ref{ass:rmpc_inc_stab} as
\begin{equation}\label{eq:rmpc_vd_ub}
    \norm{\x-\z} \leq \frac{1}{\sqrt{\cdl}}\sqrt{\Vdxz}.
\end{equation}
Inserting this upper bound in \eqref{eq:rmpc_js_ub_2} gives for $\tau \in [0,T]$:
\begin{equation}\label{eq:rmpc_jsx_ub}
    \Ljsx \norm{\tz_{\ttT}-\z_{\tTt}^*} \leq \frac{\Ljsx}{\sqrt{\cdl}}\sqrt{\Vd(\tz_{\ttT},\z_{\tTt}^*)}.
\end{equation}
In other words, the difference between candidate and previously optimal states is upper-bounded by some other Lipschitz constant $\frac{\Ljsx}{\sqrt{\cdl}}$ multiplied with the incremental Lyapunov function evaluated at the corresponding point.

To refine the expression for $\Ljsu \norm{\tv_{\ttT}-\bv_{\tTt}^*}$, note that \eqref{eq:ass_inc_stab_cjs} implies that control law $\kappaxzv$ is Lipschitz continuous given that the inputs are constrained by a box, as defined in \eqref{eq:con_sys}. Thus, the following holds for $\tau \in [0,T]$:
\begin{alignat}{2}\label{eq:rmpc_jsu_ub}
    \norm{\tv_{\ttT}-\bv_{\tTt}^*} &\overset{\text{(a)}}{=} &&\norm{\kappa(\tz_{\ttT},\z_{\tTt}^*,\bv_{\tTt}^*)-\bv_{\tTt}^*}\notag\\
    &\overset{\text{(b)}}{=} &&\norm{\bv_{\tTt}^* + \kappad(\tz_{\ttT},\z_{\tTt}^*) - \bv_{\tTt}^*}\notag\\
    &\overset{\text{(c)}}{\leq} &&\bkappa \sqrt{\Vd(\tz_{\ttT},\z_{\tTt}^*)},
\end{alignat}
where (a) is obtained by filling in \eqref{eq:rmpc_candidate_v}, (b) by filling in control law definition \eqref{eq:rmpc_cl}, and (c) by the resulting Lipschitz bound on the feedback law in $\kappad(\tz_{\ttT},\z_{\tTt}^*)$ with constant $\bkappa$.

Combining \eqref{eq:rmpc_js_ub} and \eqref{eq:rmpc_js_ub_2} and filling in upper bounds \eqref{eq:rmpc_jsx_ub} and \eqref{eq:rmpc_jsu_ub} gives the following stage cost upper bound for $\tau \in [0,T]$:
\begin{align}\label{eq:rmpc_js_ub_3}
    &\Js(\tz_{\ttT},\tv_{\ttT},\r_{\tpTst}) - \Js(\z_{\tTt}^*,\bv_{\tTt}^*,\r_{\tpTst})\notag\\
    &\leq \left(\frac{\Ljsx}{\sqrt{\cdl}}+\Ljsu \bkappa\right)\sqrt{\Vd(\tz_{\ttT},\z_{\tTt}^*)}\notag\\
    &\eqqcolon \Ljsxu \sqrt{\Vd(\tz_{\ttT},\z_{\tTt}^*)}.
\end{align}

Using the upper bound for $\sqrt{\Vd(\tz_{\ttT},\z_{\tTt}^*)}$ in \eqref{eq:vd_bound_s_tau}, we can define the following upper bound on the stage costs for $\tau \in [0,T]$:
\begin{align}\label{eq:rmpc_js_ub_4}
    &\Js(\tz_{\ttT},\tv_{\ttT},\r_{\tpTst}) - \Js(\z_{\tTt}^*,\bv_{\tTt}^*,\r_{\tpTst})\notag\\
    &\leq \Ljsxu e^{-\rho \tau} \left(1-e^{\rho \Ts}\right)\frac{\bw}{\rho} \eqqcolon \alpha_\tau\tops(\rho,\bw,\Ts),
\end{align}
or, alternatively,
\begin{equation}\label{eq:rmpc_js_ub_5}
    \Js(\tz_{\ttT},\tv_{\ttT},\r_{\tpTst}) \leq \Js(\z_{\tTt}^*,\bv_{\tTt}^*,\r_{\tpTst}) + \alpha_\tau\tops(\rho,\bw,\Ts).
\end{equation}
In other words, this means that the candidate stage cost is maximum $\alpha_\tau\tops(\rho,\bw,\Ts)$ away from the previously optimal stage cost at prediction time $\tau \in [0,T]$.

Next to the stage costs, the terminal costs of the candidate and extended previously optimal solution \eqref{eq:rmpc_candidate} are not equal, i.e., $\Jf(\tz_{\TtT},\x_{\tpTsT}\tr) \neq \Jf(\z_{\TTt}^*,\x_{\tpTsT}\tr)$. Since the terminal cost is also continuously differentiable in compact set $\XtbX$, it is Lipschitz continuous. Therefore, we can write a similar inequality as for the difference in stage costs \eqref{eq:rmpc_js_ub_2}:
\begin{align}\label{eq:rmpc_jf_ub}
    &\Jf(\tz_{\TtT},\x_{\tpTsT}\tr) - \Jf(\z_{\TTt}^*,\x_{\tpTsT}\tr)\notag\\
    &= \norm{\tz_{\TtT}-\x_{\tpTsT}\tr}_P^2 - \norm{\z_{\TTt}^*-\x_{\tpTsT}\tr}_P^2\notag\\
    &\leq \Ljfx \norm{\tz_{\TtT}-\z_{\TTt}^*}.
\end{align}
Following the same steps above, we can write the upper bound on the difference in terminal costs as
\begin{align}\label{eq:rmpc_jf_ub_2}
    &\Jf(\tz_{\TtT},\x_{\tpTsT}\tr) - \Jf(\z_{\TTt}^*,\x_{\tpTsT}\tr)\notag\\
    &\leq \Ljf e^{-\rho T} \left(1-e^{\rho \Ts}\right)\frac{\bw}{\rho} \eqqcolon \alpha\topf(\rho,\bw,\Ts).
\end{align}

We can now further work out the trajectory tracking proof in \eqref{eq:rmpc_jopt_db} by replacing the equality for (c) in \eqref{eq:tmpc_jopt_db} with the inequality in \eqref{eq:rmpc_js_ub_5}:
\begin{subequations}
    \begin{alignat}{2}
        \J_{\tTs}^*
        &\overset{\text{(a)}}{\leq} &&\int_{0}^{T-\Ts}\J_{\ttT}\tops d\tau + \int_{T-\Ts}^{T}\J_{\ttT}\tops d\tau + \J_{\TtT}\topf\\
        &\overset{\text{(b)}}{\leq} &&\int_{\Ts}^{T}\J_{\tt}\tops d\tau + \int_{\Ts}^{T}\alpha_\tau\tops(\rho,\bw,\Ts) d\tau \notag\\
        &&&+\int_{T}^{T+\Ts}\J_{\tt}\tops d\tau + \int_{T}^{T+\Ts}\alpha_\tau\tops(\rho,\bw,\Ts) d\tau + \J_{\TtT}\topf\\
        &\overset{\text{(c)}}{=} &&\int_{0}^{T}\J_{\tt}\tops d\tau - \int_{0}^{\Ts}\J_{\tt}\tops d\tau + \int_{\Ts}^{T}\alpha_\tau\tops(\rho,\bw,\Ts) d\tau \notag\\
        &&&+\int_{T}^{T+\Ts}\J_{\tt}\tops d\tau + \int_{T}^{T+\Ts}\alpha_\tau\tops(\rho,\bw,\Ts) d\tau + \J_{\TtT}\topf\\
        &\overset{\text{(d)}}{=} &&\J_t^* - \J_{\Tt}\topf - \int_{0}^{\Ts}\J_{\tt}\tops d\tau + \int_{\Ts}^{T}\alpha_\tau\tops(\rho,\bw,\Ts) d\tau \notag\\
        &&&+\int_{T}^{T+\Ts}\J_{\tt}\tops d\tau + \int_{T}^{T+\Ts}\alpha_\tau\tops(\rho,\bw,\Ts) d\tau + \J_{\TtT}\topf\\
        &\overset{\text{(e)}}{=} &&\J_t^* - \int_{0}^{\Ts}\J_{\tt}\tops d\tau + \int_{\Ts}^{T}\alpha_\tau\tops(\rho,\bw,\Ts) d\tau \notag\\
        &&&+\J_{\TtT}\topf - \J_{\Tt}\topf + \int_{T}^{T+\Ts}\J_{\tt}\tops d\tau + \int_{T}^{T+\Ts}\alpha_\tau\tops(\rho,\bw,\Ts) d\tau\\
        &\overset{\text{(f)}}{\leq} &&\J_t^* - \int_{0}^{\Ts}\J_{\tt}\tops d\tau + \int_{\Ts}^{T}\alpha_\tau\tops(\rho,\bw,\Ts) d\tau \notag\\
        &&&+\J_{\TTt}\topf + \alpha\topf(\rho,\bw,\Ts) - \J_{\Tt}\topf + \int_{T}^{T+\Ts}\J_{\tt}\tops d\tau + \int_{T}^{T+\Ts}\alpha_\tau\tops(\rho,\bw,\Ts) d\tau\\
        &\overset{\text{(g)}}{\leq} &&\J_t^* - \int_{0}^{\Ts}\J_{\tt}\tops d\tau + \int_{\Ts}^{T}\alpha_\tau\tops(\rho,\bw,\Ts) d\tau + \int_{T}^{T+\Ts}\alpha_\tau\tops(\rho,\bw,\Ts) d\tau + \alpha\topf(\rho,\bw,\Ts),
    \end{alignat}
\end{subequations}
where (a) is obtained by \eqref{eq:rmpc_jopt_db}, (b) by filling in \eqref{eq:rmpc_js_ub_5} in $\int_{0}^{T-\Ts}\J_{\ttT}\tops d\tau$ and $\int_{T-\Ts}^{T}\J_{\ttT}\tops d\tau$, (c), (d), and (e) by following the same steps as (d), (e), and (f) in \eqref{eq:tmpc_jopt_db}, (f) by \eqref{eq:rmpc_jf_ub_2}, and (g) by \eqref{eq:rmpc_ass_term_jf} in Assumption~\ref{ass:rmpc_term_ing}. Leveraging \eqref{eq:rmpc_ass_term_jf} in Assumption~\ref{ass:rmpc_term_ing} gives
\begin{equation}\label{eq:rmpc_jopt_db_2}
    \J_{\tTs}^* - \J_t^* \leq -\int_{0}^{\Ts} \J_{\tt}\tops d\tau + \int_{\Ts}^{T}\alpha_\tau\tops(\rho,\bw,\Ts) d\tau + \int_{T}^{T+\Ts}\alpha_\tau\tops(\rho,\bw,\Ts) d\tau + \alpha\topf(\rho,\bw,\Ts).
\end{equation}
Thus, we get the same descent bound for $\J_t^*$ as in Section~\ref{sec:tmpc_proof_tracking}, up to
\begin{equation}
    \balphasf(\rho,\bw,\Ts) \coloneqq \int_{\Ts}^{T}\alpha_\tau\tops(\rho,\bw,\Ts) d\tau + \int_{T}^{T+\Ts}\alpha_\tau\tops(\rho,\bw,\Ts) d\tau + \alpha\topf(\rho,\bw,\Ts).
\end{equation}

Similar to Section~\ref{sec:tmpc_proof_tracking}, we prove convergence by first noting that the descent property of the optimal cost \eqref{eq:rmpc_jopt_db_2} implies
\begin{equation}
    \J_{\tTs}^* - \J_t^* \leq -c^{\J,\mathrm{d}} \int_{t}^{\tTs} \norm{\z_{\tau-t|t}^*-\x_\tau\tr}_Q^2d\tau + \balphasf(\rho,\bw,\Ts),
\end{equation}
with constant $c^{\J,\mathrm{d}} > 0$. Re-arranging the terms gives
\begin{equation}\label{eq:rmpc_track_ub_int}
    c^{\J,\mathrm{d}} \int_{t}^{\tTs} \norm{\z_{\tau-t|t}^*-\x_\tau\tr}_Q^2d\tau \leq \J_t^* - \J_{\tTs}^* + \balphasf(\rho,\bw,\Ts).
\end{equation}
Instead of $\norm{\z_{\tau-t|t}^*-\x_\tau\tr}_Q^2$ we are interested in the actual tracking error $\norm{\x_\tau-\x_\tau\tr}_Q^2$, which is upper-bounded by
\begin{subequations}\label{eq:rmpc_track_ub_norm}
    \begin{align}
        \norm{\x_\tau-\x_\tau\tr}_Q^2 &\overset{\text{(a)}}{\leq} c^\mathrm{t} \norm{\x_\tau-\z_{\tau-t|t}^*} + \norm{\z_{\tau-t|t}^*-\x_\tau\tr}_Q^2\\
        &\overset{\text{(b)}}{\leq} c^\mathrm{t,1} s_\tau + \norm{\z_{\tau-t|t}^*-\x_\tau\tr}_Q^2
    \end{align}
\end{subequations}
where (a) is obtained by a Lipschitz bound with constant $c^\mathrm{t}$ and (b) by another Lipschitz bound with constant $c^\mathrm{t,1}$ on the norm using a combination of \eqref{eq:ass_inc_stab_vd_bounds}, \eqref{eq:tube}, and \eqref{eq:rmpc_s_t}. Therefore, we can write a similar upper bound as in \eqref{eq:rmpc_track_ub_int} but now for $\norm{\x_\tau-\x_\tau\tr}_Q^2$:
\begin{subequations}\label{eq:rmpc_track_ub_int_2}
    \begin{align}
        c^{\J,\mathrm{d}} \int_{t}^{\tTs} \norm{\x_\tau-\x_\tau\tr}_Q^2 d\tau &\leq c^{\J,\mathrm{d}} \int_{t}^{\tTs} \norm{\z_{\tau-t|t}^*-\x_\tau\tr}_Q^2 + c^\mathrm{t,1} s_\tau d\tau\\
        &\overset{\text{(a)}}{=} c^{\J,\mathrm{d}} \int_{t}^{\tTs} \norm{\z_{\tau-t|t}^*-\x_\tau\tr}_Q^2 d\tau + c^{\J,\mathrm{d}} \int_{t}^{\tTs} c^\mathrm{t,1} s_\tau d\tau\\
        &\overset{\text{(b)}}{\leq} \J_t^* - \J_{\tTs}^* + \balphasfone(\rho,\bw,\Ts),
    \end{align}
\end{subequations}
with $\balphasfone(\rho,\bw,\Ts) \coloneqq \balphasf(\rho,\bw,\Ts) + c^{\J,\mathrm{d}} c^\mathrm{t,1} \left(\Ts-\frac{1}{\rho}\left(1-e^{-\rho\Ts}\right)\right)\frac{\bw}{\rho}$, and where (a) is obtained by writing out the integral expression and (b) by \eqref{eq:rmpc_track_ub_int}. Iterating this inequality from $0$ to $t=2\Ts$ yields
\begin{subequations}
    \begin{align}
        c^{\J,\mathrm{d}} \int_{0}^{2\Ts} \norm{\x_\tau-\x_\tau\tr}_Q^2d\tau \leq &\J_0^* - \J_{\Ts}^* + \balphasfone(\rho,\bw,\Ts) + \J_{\Ts}^* - \J_{2\Ts}^* + \balphasfone(\rho,\bw,\Ts)\\
        \leq &\J_0^* - \J_{2\Ts}^* + 2\balphasfone(\rho,\bw,\Ts)\\
        = &\J_0^* - \J_{2\Ts}^* + \frac{t}{\Ts}\balphasfone(\rho,\bw,\Ts),
    \end{align}
\end{subequations}
In contrast to \eqref{eq:tmpc_err_ub} in Section~\ref{sec:tmpc_proof_tracking}, iterating this inequality for $t \rightarrow \infty$ is not necessarily bounded for the worst-case disturbance impact. Therefore, we can conclude that the average of this inequality gets small by dividing it by $t$ and leveraging the fact that $\J_t^*$ is uniformly bounded:
\begin{equation}
    \underset{t\rightarrow\infty}{\oplim}\frac{\int_{0}^{t} \norm{\x_\tau-\x_\tau\tr}_Q^2d\tau}{t} \leq \frac{\J_0^* - \underset{t\rightarrow\infty}{\operatorname{lim}}\J_t^*}{c^{\J,\mathrm{d}} t} + \frac{1}{c^{\J,\mathrm{d}}\Ts}\balphasfone(\rho,\bw,\Ts).
\end{equation}
Thus, by applying Barbalat's lemma \cite{barbalat1959systemes} on $\frac{\int_{0}^{t} \norm{\x_\tau-\x_\tau\tr}_Q^2d\tau}{t}$ we conclude that, on average, the tracking error $\norm{\x_t-\x_t\tr}$ gets small over time, proportional to $\balphasfone(\rho,\bw,\Ts)$. Note that this conclusion makes sense in the presence of unpredictable and bounded disturbances. \qed

\subsubsection{Design to satisfy the assumptions}\label{sec:rmpc_properties_design}
This section will detail the design of:
\begin{itemize}
    \item the obstacle avoidance constraints, which need to satisfy Assumption~\ref{ass:obs};
    \item incremental Lyapunov function $\Vdxz$ and corresponding Lipschitz constants $\cjs, j \in \Ns$ and $\co$, which need to satisfy Assumption~\ref{ass:rmpc_inc_stab};
    \item the terminal ingredients, which need to satisfy Assumption~\ref{ass:rmpc_term_ing};
    \item the reference trajectory, which needs to satisfy Assumptions~\ref{ass:rmpc_gjrs} and~\ref{ass:rmpc_gjro}.
\end{itemize}

\noindent\textbf{Obstacle avoidance constraints design}\\
To prove recursive feasibility, we need the obstacle avoidance constraint to satisfy Assumption~\ref{ass:obs}. We can achieve this by following the same design as written in Section~\ref{sec:tmpc_properties_design}.\\

\noindent\textbf{Incremental stabilizability design}\\
The incremental stabilizability design follows a similar procedure to the terminal ingredients design in Section~\ref{sec:tmpc_properties_design}. The goal is to compute a suitable incremental Lyapunov function $\Vdxz$ that is bounded by tube size $s$ given in \eqref{eq:rmpc_sdot} and corresponding control law $\kappaxzv$ that satisfy properties \eqref{eq:ass_inc_stab_vd_bounds}-\eqref{eq:ass_inc_stab_triangle_ineq}.

\begin{remark}
    The most important difference with the design in Section~\ref{sec:tmpc_properties_design} is the fact that $\Vdxz$ should satisfy contraction property \eqref{eq:ass_inc_stab_vd_contract} that is used to prove trajectory tracking and recursive feasibility in the presence of disturbances. Furthermore, control law $\kappaxzv$ is used in the proofs and implemented in closed-loop execution according to \eqref{eq:rmpc_cl}.
\end{remark}

We can write the dynamics of $\sVdxz$ as follows:
\begin{equation}\label{eq:ddt_vd_sqrt}
    \frac{d}{dt}\sqrt{\Vdxz} = \frac{d}{dt}\left(\Vdxz\right)^\frac{1}{2} = \frac{1}{2}\Vdxz^{-\frac{1}{2}}\frac{d}{dt}\Vdxz.
\end{equation}
Since it is easier to find an expression for $\frac{d}{dt}\Vdxz$ rather than $\frac{d}{dt}\sqrt{\Vdxz}$, we rewrite \eqref{eq:ddt_vd_sqrt} as
\begin{equation}
    \frac{d}{dt}\Vdxz = 2\sqrt{\Vdxz}\frac{d}{dt}\sqrt{\Vdxz}.
\end{equation}
Filling in desired tube dynamics \eqref{eq:rmpc_sdot} gives the following upper bound on $\frac{d}{dt}\Vdxz$:
\begin{equation}\label{eq:ddt_vdarg_des}
    \frac{d}{dt}\Vdxz \leq 2\sqrt{\Vdxz}\left(-\rho \sVdxz + \bw\right) = -2\rho \Vdxz + 2\sqrt{\Vdxz}\bw.
\end{equation}
Thus, the goal is to find a suitable incremental Lyapunov function of which the upper bound can be written in the form \eqref{eq:ddt_vdarg_des}.

Using control contraction metrics (CCMs) \cite{manchester2017control}, we can define the incremental Lyapunov function $\Vdxz$ as
\begin{equation}\label{eq:rmpc_vd_ccm_actual}
    \Vdxz \coloneqq \underset{\cxs\in C(\xz)}{\opmin} \int_{0}^{1} \cxss^\top \Pdcs \cxss ds,
\end{equation}
with metric $\Pdcs$ and where $\cxs$ is a curve in set of curves
\begin{equation}
    C(\xz) \coloneqq \left\{\cxs : [0,1] \rightarrow \Rx\ \middle|\ \cx(0) = \z, \cx(1) = \x\right\}.
\end{equation}
$\gxs \in C(\xz)$ denotes the geodesic: the curve that minimizes $\Vdxz$.

We know that the evolutions of the nominal and disturbed state are described by \eqref{eq:xdot_nom} and \eqref{eq:xdot_w}, respectively. Thus, we know the evolution of the endpoints $\gx(0)$ and $\gx(1)$ of geodesic $\gxs$ over time. A logical candidate for the evolution of $\gxs, s \in [0,1]$, over time, i.e., one that matches $\dz$ and $\dx$ for $s=0$ and $s=1$, respectively, is thus given by
\begin{equation}\label{eq:rmpc_dgxs}
    \dgxs \coloneqq f(\gxs,\gus) + E(\wb+s\w^0),
\end{equation}
with control law $\gus$ that is obtained by integrating the feedback law over the geodesic:
\begin{equation}\label{eq:rmpc_gus}
    \gus \coloneqq \bv + \int_{0}^{s} \Kd(\gxts) \gxtss d\ts.
\end{equation}
Similarly, a candidate for the evolution of the path derivative of the geodesic over time is given by
\begin{equation}\label{eq:rmpc_dgxss}
    \frac{d}{dt}\gxss \coloneqq \Aclargs\gxss + E\w^0,
\end{equation}
with closed-loop dynamics
\begin{equation}
    \Aclargs \coloneqq \Ags + \Bgs\Kdgs,
\end{equation}
and Jacobians
\begin{equation}\label{eq:rmpc_a_b}
    \Ags \coloneqq \frac{\partial \fwarg}{\partial \x}\left.\right|_{(\gxs,\gus)}, \quad \Bgs \coloneqq \frac{\partial \fwarg}{\partial \bu}\left.\right|_{(\gxs,\gus)},
\end{equation}
since, locally, the nonlinear dynamics are given by its Jacobians.

Using definitions \eqref{eq:rmpc_dgxs} and \eqref{eq:rmpc_dgxss} for $\dgxs$ and $\frac{d}{dt}\gxss$, respectively, and temporarily using the following shorthand notations:
\begin{subequations}\label{eq:rmpc_shorthand}
    \begin{align}
        \gx &\coloneqq \gxs,\label{eq:rmpc_gx_shorthand}\\
        \gu &\coloneqq \gus,\label{eq:rmpc_gu_shorthand}\\
        \Acl &\coloneqq \Aclarg,\\
        A &\coloneqq \Ag,\\
        B &\coloneqq \Bg,\\
        \gxssshort &\coloneqq \gxss,\\
        \dgxssshort &\coloneqq \frac{d}{dt}\gxss,\\
        \Pd &\coloneqq \Pdg,
    \end{align}
\end{subequations}
we can write the time evolution of the upper bound of $\frac{d}{dt}\Vdxz$ in the form \eqref{eq:ddt_vdarg_des} using candidates $\gxssshort$ and $\dgxssshort$:
\begin{subequations}\label{eq:eq:ddt_vdarg}
    \begin{align}
        \frac{d}{dt}\Vdxz &\leq \int_{0}^{1} \dgxssshorttop \Pd \gxssshort + \gxssshorttop \dPd \gxssshort + \gxssshorttop \Pd \dgxssshort ds\\
        &= \int_{0}^{1} \gxssshorttop \left(\Acltop \Pd + \dPd + \Pd \Acl\right) \gxssshort + \left(E\w^0\right)^\top \Pd \gxssshort + \gxssshorttop \Pd E\w^0 ds\label{eq:ddt_vdarg_1}\\
        &= \int_{0}^{1} \gxssshorttop \left(\Acltop \Pd + \dPd + \Pd g\Acl\right) \gxssshort + 2\gxssshorttop \Pd E\w^0 ds\\
        &= \int_{0}^{1} \gxssshorttop \left(\Acltop \Pd + \dPd + \Pd \Acl\right) \gxssshort ds + 2 \int_{0}^{1} \gxssshorttop \Pd E\w^0 ds\\
        &= \int_{0}^{1} \gxssshorttop \left(\Acltop \Pd + \dPd + \Pd \Acl\right) \gxssshort ds + 2 \int_{0}^{1} \gxssshorttop \Pd^\frac{1}{2} \Pd^\frac{1}{2} E\w^0 ds\\
        &\overset{\text{(a)}}{\leq} -2\rho \int_{0}^{1} \gxssshorttop \Pd \gxssshort ds + 2 \int_{0}^{1} \lrnorm{\gxssshorttop \Pd^\frac{1}{2}} \lrnorm{\Pd^\frac{1}{2} E\w^0} ds\\
        &\overset{\text{(b)}}{=} -2\rho \Vdxz + 2 \int_{0}^{1} \lrnorm{\gxssshorttop}_{\Pd} \lrnorm{E\w^0}_{\Pd} ds\\
        &\overset{\text{(c)}}{\leq} -2\rho \Vdxz + 2 \sqrt{\Vdxz} \bw,
    \end{align}
\end{subequations}
for $\w^0\in\W$ where (a) is obtained by the combination of the following contraction LMI \cite{manchester2017control}:
\begin{equation}\label{eq:lmi_contract_nonconvex}
    \begin{aligned}
        &\dPdg + \Pdg\left(\Ag+\Bg\Kdg\right) +\\
        &\left(\Ag+\Bg\Kdg\right)^\top \Pdg + 2\rho\Pdg \preceq 0,
    \end{aligned}
\end{equation}
with $\dPdg \coloneqq \frac{\partial \Pdg}{\partial \gx}\dgx$ and contraction rate $\rho > 0$, which is based on the contraction analysis for nonlinear systems as presented in \cite{lohmiller1998contraction}, and the Cauchy-Schwarz inequality, (b) by the definition of $\Vdxz$ in \eqref{eq:rmpc_vd_ccm_actual}, and (c) by the following definition of $\bw$:
\begin{equation}\label{eq:rmpc_bw_state_dependent}
    \bw \coloneqq \underset{\substack{\w^0\in\W\\\gx \in C(\xz)}}{max} \lrnorm{E\w^0}_\bPdg,
\end{equation}
with
\begin{equation}
    \bPdg \coloneqq \lambdamax\left(\Pdg\right) \Inx,
\end{equation}
where $\lambdamax(\Pdg)$ denotes the maximum eigenvalue of matrix $\Pdg$.

\begin{remark}
    Since we use shorthand notations \eqref{eq:rmpc_gx_shorthand} and \eqref{eq:rmpc_gu_shorthand}, we can evaluate LMI \eqref{eq:lmi_contract_nonconvex} for an infinitely large number of points $s \in [0,1]$ on the geodesic. Thus, \eqref{eq:lmi_contract_nonconvex} can also be considered a set of LMIs. Since the constraints tightening is such that $(\gx,\gu)\in\Z, s\in[0,1]$ \cite[Prop.~5]{sasfi2023robust}, it suffices to grid $\Z$ in a similar way as described in Section~\ref{sec:tmpc_properties_design}.
\end{remark}

\begin{remark}
    Compared to \cite{manchester2017control}, \eqref{eq:lmi_contract_nonconvex} with shorthand notation \eqref{eq:rmpc_shorthand}, does not contain the time-dependency since we do not consider a time-varying system.
\end{remark}

Note that \eqref{eq:rmpc_bw_state_dependent} confirms that Property~\ref{property:w} holds for compact disturbance set \eqref{eq:w_bb}.

Note also that \eqref{eq:lmi_contract_nonconvex} contains the multiplication of $\Pdg$ and $\Kdg$. Therefore, it is a bilinear LMI in the context of an SDP in which both $\Pdg$ and $\Kdg$ are decision variables. Under change of coordinates $\Xdg=\Pdg^{-1}$ and $\Ydg=\Kdg\Pdg^{-1}$ \eqref{eq:lmi_contract_nonconvex} translates to the following convex contraction LMI \cite{manchester2017control}:
\begin{equation}\label{eq:lmi_contract}
    \begin{aligned}
        &-\dXdg + \Ag\Xdg+\Bg\Ydg +\\
        &\left(\Ag\Xdg+\Bg\Ydg\right)^\top + 2\rho\Xdg \preceq 0.
    \end{aligned}
\end{equation}
Thus, any combination of $\Pdg$ and corresponding control law $\kappaxzv$ that is obtained by integrating over the geodesic, i.e.,
\begin{equation}\label{eq:rmpc_kappa_ccm}
    \kappaxzv = \gu(1),
\end{equation}
with $\gus$ given in \eqref{eq:rmpc_gus}, that satisfies LMI \eqref{eq:lmi_contract} ensures that the closed-loop trajectory $\x$ will exponentially contract towards $\z$ with rate $\rho$. For a more elaborate exposition of this result, please refer to \cite{sasfi2023robust} and references therein \cite{lohmiller1998contraction,manchester2017control,zhao2022tube,singh2023robust,schiller2023lyapunov}.

In general, $\gxs$ is a nonlinear curve, so evaluating \eqref{eq:rmpc_vd_ccm_actual} as part of terminal set \eqref{eq:rmpc_xf} in the solver is computationally expensive. To reduce solve time, we choose a state-independent incremental Lyapunov function, i.e., $\Pd$ instead of $\Pdg$, such that the geodesic becomes a straight line connecting $\z$ and $\x$. In this case, the geodesic is mathematically described by
\begin{equation}\label{eq:rmpc_gxs_straight}
    \gxs = \z + s (\x-\z)
\end{equation}
and
\begin{equation}\label{eq:rmpc_gxss_straight}
    \gxss = \x-\z.
\end{equation}
As a result, contraction LMI \eqref{eq:lmi_contract} changes to
\begin{equation}\label{eq:lmi_contract_noxdarg}
    \Ag\Xdg+\Bg\Ydg + \left(\Ag\Xdg+\Bg\Ydg\right)^\top + 2\rho\Xd \preceq 0,
\end{equation}
in which the term $-\dXdg$ disappears compared to \eqref{eq:lmi_contract} as the time derivative is given by $\dXdg = \frac{\partial \dXdg}{\partial \gx}\dgx$. Furthermore, incremental Lyapunov function \eqref{eq:rmpc_vd_ccm_actual} simplifies to the following quadratic expression:
\begin{equation}\label{eq:rmpc_vd}
    \Vdxz = (\x-\z)^T \Pd (\x-\z),
\end{equation}
or, equivalently,
\begin{equation}\label{eq:rmpc_vd_sqrt}
    \sqrt{\Vdxz} = \norm{\x-\z}_{\Pd},
\end{equation}
and the control law is given by
\begin{equation}\label{eq:rmpc_kappa}
    \kappaxzv = \gu(1) = \bv + \int_{0}^{1} \Kdgs(\x-\z) ds.
\end{equation}
Note that this implies that \eqref{eq:ass_inc_stab_vd_bounds} is trivially satisfied with $\cdl = \lambdamin(\Pd)$ and $\cdu = \lambdamax(\Pd)$, where $\lambdamin(\Pd)$ and $\lambdamax(\Pd)$ denote the minimum and maximum eigenvalues of $\Pd$, respectively, and \eqref{eq:ass_inc_stab_triangle_ineq} is also satisfied. Correspondingly, we can constant $\bw$ in \eqref{eq:vd_sqrt_w} as
\begin{equation}\label{eq:rmpc_bw}
    \bw \coloneqq \underset{\w^0\in\opvert(\W)}{\opmax}\norm{E\w^0}_{\Pd},
\end{equation}
since $\W$ is a convex polytopic set. The choice for a state-independent $\Pd$ increases conservatism since the resulting curve length is larger than nonlinear geodesic $\gxs$ in parts of the state space where Jacobians \eqref{eq:rmpc_a_b} are highly nonlinear. Consequently, integrating the control law over this curve results in a sub-optimal control strategy. This control strategy will result in a contraction rate of at least $\rho$, but it will be slightly less effective than the control strategy based on a nonlinear geodesic.

To reduce conservatism, we leverage a state-dependent feedback gain $\Kdg$. $\Kdg$ is computed outside the solver using \eqref{eq:rmpc_kappa_ccm}, so computation time is less important for enabling real-time implementation.

\begin{remark}
    Note that CCM methods provide a way to ensure uniform exponential stability, i.e., exponential stability guarantees that hold in the complete state space, even though metric $\Pdarg$ is state-dependent. This contrasts with the design of tracking LMI \eqref{eq:lmi_tracking}, which includes reference-dependent matrix $P(\r)$, but is only valid in local regions $\epsilon$ around the reference trajectory. In practice, the value of $\epsilon$ for which the results hold is hard to obtain. Therefore, to obtain strict safety guarantees in the robust case, it is practically recommended to use CCMs according to the design above or remove the state dependency in $\Pd$. Using CCMs, uniform exponential stabilizability follows from computing integral expressions in the metric space. Using state-independent $\Pd$, uniform exponential stabilizability follows as a direct consequence of the mean value theorem (MVT) since choosing a constant CCM $\Pd$ results in a straight line minimizing geodesic $\gxs$ between $\z$ and $\x$ as described above.
\end{remark}

Contraction LMI \eqref{eq:lmi_contract} is the main property that enables the robust recursive feasibility and trajectory tracking proofs to hold. The resulting metric $\Pd$ is used to construct incremental Lyapunov function \eqref{eq:rmpc_vd_ccm_actual} that satisfies \eqref{eq:ass_inc_stab_vd_contract}. Consequently, tube \eqref{eq:tube} grows to a robust positive invariant (RPI) set:
\begin{equation}\label{eq:rmpc_rpi_set}
    \sqrt{\Vd(\x_t,\z_t)}\leq\bs,\ t\geq 0
\end{equation}
with $\bs$ defined in \eqref{eq:rmpc_sbar}. To reduce conservatism we optimize the shape of \eqref{eq:rmpc_rpi_set} by satisfying the following RPI LMI:
\begin{equation}\label{eq:lmi_rpi}
    \begin{bmatrix}
        \Aarg\Xd+\Barg\Ydarg+\left(\Aarg\Xd+\Barg\Ydarg\right)^\top+\lambda \Xd&E\w^0\\
        \left(E\w^0\right)^\top&-\lambda\delta^2
    \end{bmatrix} \preceq 0,
\end{equation}
with $\bmzeta\in\Z$, $\w^0\in \opvert(\W)$ and multiplier $\lambda \geq 0$. This LMI enforces that sublevel set $\Vdxz \leq \delta^2$ is RPI for disturbed dynamics \eqref{eq:xdot_w} for all $\w^0\in\W$ and with $\delta\coloneqq\bs$. Note that $\rho$ is a hyperparameter and $\bw$ can be computed after optimizing for $\Pd$ using \eqref{eq:rmpc_bw}. Therefore, without loss of generality, we can set $\delta = 1$, which helps to bridge between expressions in quadratic form $\Vdxz$ and linear form $\sqrt{\Vdxz}$. Furthermore, note that, since the disturbances enter linearly in system dynamics \eqref{eq:xdot_w} and the set $\W$ is convex, it suffices to evaluate a set of LMIs of the form \eqref{eq:lmi_rpi} at the vertices of $\W$. Appendix~\ref{app:lmi_rpi} provides the proof for \eqref{eq:lmi_rpi}.

\begin{remark}
    An alternative to designing the RPI set using polytopic set $\W$ is to leverage universal $\Linf$-gain bounds on disturbances $\w^0$ as presented in \cite{zhao2022tube}.
\end{remark}

In contrast to the TMPC design, in which terminal set scaling $\alpha$ is computed based on the desired distance to obstacles $\dmax$, the terminal set and corresponding RPI set of $\Vdxz$ have a minimum size caused by disturbances $\w^0\in\W$. To prevent the sets from becoming large, thus resulting in conservative behavior, $\Pd$ and $\Kdarg$ are computed using the following SDP, in which the tightening constants $\cjs, j \in \Ns$ and $\co$, normalized with respect to the corresponding constraint interval, are minimized:
\begin{subequations}\label{eq:rmpc_sdp}
    \begin{align}
        \underset{\substack{\Xd,\Ydarg\\{\cjs}^2,{\co}^2}}{\opmin}\ \ &\cco {\co}^2 + \sum_{j=1}^{\ns} \cjcs {\cjs}^2,\label{eq:rmpc_sdp_obj}\\
        \operatorname{s.t.}\ &\Xd \succeq 0,\label{eq:rmpc_sdp_lmi_x}\\
        &\Aarg\Xd+\Barg\Ydarg + \left(\Aarg\Xd+\Barg\Ydarg\right)^\top + 2\rho\Xd \preceq 0,\label{eq:rmpc_sdp_lmi_contract}\\
        &\begin{bmatrix}
            \Aarg\Xd+\Barg\Ydarg+\left(\Aarg\Xd+\Barg\Ydarg\right)^\top+\lambda \Xd&E\w^0\\
            \left(E\w^0\right)^\top&-\lambda
        \end{bmatrix} \preceq 0,\label{eq:rmpc_sdp_lmi_rpi}\\
        &\begin{bmatrix}
            {\cjs}^2&\Ljs\begin{bmatrix}\Ydarg\\\Xd\end{bmatrix}\\
            \left(\Ljs\begin{bmatrix}\Ydarg\\\Xd\end{bmatrix}\right)^\top&\Xd
        \end{bmatrix} \succeq 0,\ j \in \Ns,\label{eq:rmpc_sdp_lmi_sys}\\
        &\begin{bmatrix}
            {\co}^2 I^{\np}&M\Xd\\
            \left(M\Xd\right)^\top&\Xd
        \end{bmatrix} \succeq 0,\label{eq:rmpc_sdp_lmi_obs}\\
        &\bmzeta \in \Z, \quad \w^0\in\opvert\left(\W\right),\notag
    \end{align}
\end{subequations}
with contraction rate $\rho > 0$ and multiplier $\lambda \geq 0$ that can be computed using bi-section over this SDP and $\delta = 1$.

Note that LMI \eqref{eq:rmpc_sdp_lmi_sys}, to ensure Lipschitz continuity of the system constraints, is the same as \eqref{eq:lmi_sys}. We can derive similar a LMI \eqref{eq:rmpc_sdp_lmi_obs} to ensure Lipschitz continuity of the obstacle avoidance constraints. The proof is similar to the proof for \eqref{eq:lmi_sys} as written in \cite{nubert2019learning} and included in Appendix~\ref{app:lmi_obs} for completeness.

Similar to SDP \eqref{eq:tmpc_sdp}, SDP \eqref{eq:rmpc_sdp} is semi-infinite, so we need to evaluate the expressions at grid points and vertices, both in $\Z$ and $\W$ to account for the effect of disturbances.

\begin{remark}\label{remark:rmpc_sdp_lmi_rpi_split}
    Note that LMI \eqref{eq:rmpc_sdp_lmi_rpi} has to be evaluated at all grid points in both $\Z$ and $\W$, meaning that the number of LMIs equals the multiplication of the number of grid points in both sets. This might result in a drastic increase in (offline) computation time. To alleviate this effect, we can introduce decision variable $\bW$ and use the following combination of LMIs that is sufficient to ensure the satisfaction of \eqref{eq:rmpc_sdp_lmi_rpi}:
    \begin{equation}
        \begin{bmatrix}
            0^{\nx}&E\w^0\\
            \left(E\w^0\right)^\top&0
        \end{bmatrix} \preceq \bW, \quad \w^0\in \opvert\left(\W\right),
    \end{equation}
    \begin{equation}
        \bW+\begin{bmatrix}
            \Aarg\Xd+\Barg\Ydarg+\left(\Aarg\Xd+\Barg\Ydarg\right)^\top+\lambda \Xd&\bm{0}\\
            \bm{0}&-\lambda
        \end{bmatrix} \preceq 0, \bmzeta \in \Z.
    \end{equation}
\end{remark}

In conclusion, by successfully solving \eqref{eq:rmpc_sdp}, we can find matrices $\Pd$ and $\Kdarg$ such that incremental Lyapunov function $\Vdxz$ is given by \eqref{eq:rmpc_vd} and satisfies \eqref{eq:ass_inc_stab_vd_bounds} and \eqref{eq:ass_inc_stab_triangle_ineq}. Furthermore, the design of contraction LMI \eqref{eq:lmi_contract}, RPI LMI \eqref{eq:lmi_rpi}, and Lipschitz LMI \eqref{eq:lmi_sys} and \eqref{eq:rmpc_sdp_lmi_obs} ensures satisfaction of \eqref{eq:ass_inc_stab_cjs}, \eqref{eq:ass_inc_stab_co}, and \eqref{eq:ass_inc_stab_vd_contract}. Therefore, we can conclude that this design satisfies Assumption~\ref{ass:rmpc_inc_stab}.\\

\noindent\textbf{Terminal ingredients design}\\
The terminal ingredients design involves computing a suitable terminal cost $\Jf(\z,\xr)$ such that Assumption~\ref{ass:rmpc_term_ing} is satisfied. Given the design in the previous section, we know that $\sqrt{\Vdxz}$ in \eqref{eq:rmpc_vd_sqrt} decreases with at least factor $\rho$, or, equivalently, $\Vdxz$ decreases with at least factor $2\rho$, given control law $\kappaxzv$ in \eqref{eq:rmpc_kappa}. Therefore, if we design the terminal control law as
\begin{equation}
    \kappafxr = \ur + \kappad(\x,\xr) = \ur + \int_{0}^{1} \Kdgs ds\ (\x-\xr),
\end{equation}
we just need to come up with a matrix $P$ such that the following descent bound based on \eqref{eq:rmpc_ass_term_jf} is satisfied:
\begin{equation}
    \frac{d}{dt}\left((\x-\z)^\top P (\x-\z)\right) \leq -\Js(\z,\bv,\r),
\end{equation}
with $\dx = f(\x,\kappaxzv) + E\wb$ and $\dz = f(\zv) + E\wb$. The following LMI ensures that this condition holds, see also \cite{kohler2020nonlinear}:
\begin{equation}\label{eq:lmi_jf}
    \left(\Aarg+\Barg\Kdarg\right)^\top P + P\left(\Aarg+\Barg\Kdarg\right) \preceq -Q -\Kdarg^\top R \Kdarg, \quad \bmzeta \in \Z.
\end{equation}
Thus, in order to find $P$ with the lowest eigenvalues satisfying \eqref{eq:rmpc_ass_term_jf}, we solve the following SDP:
\begin{subequations}\label{eq:rmpc_sdp_jf}
    \begin{align}
        \underset{P}{\opmin}\ \ &\operatorname{trace} P,\label{eq:rmpc_sdp_jf_obj}\\
        \operatorname{s.t.}\ &\left(\Aarg+\Barg\Kdarg\right)^\top P + P\left(\Aarg+\Barg\Kdarg\right) \preceq -Q -\Kdarg^\top R \Kdarg,\label{eq:rmpc_sdp_jf_lmi_jf}\\
        &\bmzeta \in \Z\notag.
    \end{align}
\end{subequations}

\begin{remark}
    A simpler but more conservative solution would be to define constant scaling factor $\mu$ such that $P = \mu \Pd$. It can be shown that a lower bound for $\mu$ can be formulated as a function of $Q, R, \Kdarg, \rho$, and $\cdu$ such that the contraction of $\Vdxz$ by $2\rho$ implies that the minimum terminal cost decrease in \eqref{eq:rmpc_ass_term_jf} is satisfied.
\end{remark}

Thus, we can conclude that Assumption~\ref{ass:rmpc_term_ing} is satisfied.\\

\noindent\textbf{Reference trajectory design}\\
Assumptions~\ref{ass:rmpc_gjrs} and~\ref{ass:rmpc_gjro} are trivially satisfied if the system and obstacle avoidance constraints used for generating the reference trajectory are constructed according to \eqref{eq:ass_rmpc_gjrs} and \eqref{eq:ass_rmpc_gjro} using tightening constants $\cjs, j \in \Ns$ and $\co$ from SDP \eqref{eq:rmpc_sdp}, respectively, and $\alpha$ satisfying the lower bound computed in \eqref{eq:rmpc_term_inv_der}. Correspondingly, terminal set \eqref{eq:rmpc_xf} is suitably designed around the reference trajectory to ensure closed-loop system and obstacle avoidance constraints satisfaction at all times.\\

\subsection{Concluding remarks}\label{sec:rmpc_conclusion}
In conclusion, the presented RMPC scheme is an effective tool to provide safety guarantees for tracking dynamically feasible reference trajectories with a mobile robot described by disturbed dynamics \eqref{eq:xdot_w}. The scheme is based on the idea that the impact of disturbances can be bounded by a sublevel set of incremental Lyapunov function $\Vdxz$ given a suitably designed control law. This bound, which is also called the tube size, can be used to tighten system and obstacle avoidance constraints such that the closed-loop system always satisfies the actual system and obstacle avoidance constraints. Recursive feasibility is proven by constructing a minimum invariant terminal set based on $\Vdxz$. Furthermore, trajectory tracking is proven by showing that the terminal cost matrix $P$ can be computed using the weighting matrix $\Pd$ used in $\Vdxz$.

Note that tightening the system and obstacle avoidance constraints of the reference trajectory by $\cjs \alpha, j \in \Ns$ and $\co \alpha$, respectively, is a sufficient condition to guarantee closed-loop constraints satisfaction. However, a feasible solution is not required to start at least this far away from the constraints at the beginning of the prediction horizon since the constraints are tightened by an increasing tube size that starts from zero. Therefore, implementing a similar suitably designed constraint tightening scheme in the planner generating the reference trajectory will allow for less conservative motions, i.e., motions closer to obstacles, thereby reaching goals quicker.

In practice, a physical robot is subject not only to dynamic disturbances but also to noisy sensor measurements. This means that we cannot directly measure the states of the system. To provide safety guarantees in the context of sensor noise, one needs to identify the dynamical disturbances and sensor noise, compute suitable ingredients similar to the ones presented in this section, and implement a real-time RMPC scheme for which similar properties can be proven to the ones presented in this section.

\section{Robust output-feedback MPC for trajectory tracking with disturbances}\label{sec:rompc}
The goal of this section is to introduce a robust output-feedback MPC (ROMPC) scheme that will be used to track reference trajectories satisfying Property~\ref{property:r} when the system is described as given in Section~\ref{sec:rompc_sys_w_eta}. In this system, the state can only be measured indirectly via noisy output measurements. Like Section~\ref{sec:tmpc_formulation}, Section~\ref{sec:rompc_formulation} states the optimization problem and control law for the closed-loop system. Then, Section~\ref{sec:rompc_properties} builds an intuition for ROMPC and explains the required elements in the formulation and related properties. Finally, Section~\ref{sec:rompc_conclusion} gives concluding remarks on the ROMPC formulation.

\subsection{Disturbed mobile robot with measurement noise description}\label{sec:rompc_sys_w_eta}
The disturbed system dynamics are given by \eqref{eq:xdot_w}. The corresponding output equation is given by
\begin{equation}\label{eq:y_eta}
    \y = C\x + F\bmeta,
\end{equation}
with measured output $\y \in \Ry$, observation matrix $C \in \RC$, measurement or sensor noise $\bmeta \in \Reta$, and noise selection matrix $F \in \RF$.

The measurement noise is assumed to be contained in a known bounding box $\bmeta \in \H$. By recording data we can compute bounding box $\H$ with lower bound $\ubeta_i\rec$ and upper bound $\bbeta_i\rec$ for each dimension $i \in \Neta$ of the recorded measurement noise values:
\begin{equation}\label{eq:eta_bb}
   \H \coloneqq \left\{\bmeta \in \Reta\ \middle|\ \ubeta_i\rec \leq \bmeta_i \leq \bbeta_i\rec,\ i \in \Neta\right\}.
\end{equation}
This measurement noise bounding box can be used in the ROMPC formulation described in the next section.

\subsection{ROMPC formulation}\label{sec:rompc_formulation}
Consider the following ROMPC formulation for tracking reference trajectory \eqref{eq:r}:
\begin{subequations}\label{eq:rompc}
    \begin{alignat}{3}
        \J_t^*(\x_t,\r_t) = \underset{\substack{\z_{\ct},\bv_{\ct}}}{\opmin}\ \ &\mathrlap{\Jf(\z_{\Tt},\x_{\tpT}\tr) + \int_{0}^{T} \Js(\z_{\tt},\bv_{\tt},\r_{\tpt})\ d\tau,}&&&&\hspace{100pt} \label{eq:rompc_obj}\\
        \operatorname{s.t.}\ &\z_{\0t} = \hx_t,&&&& \label{eq:rompc_z0}\\
        &s_0 = 0,&&&& \label{eq:rompc_s0}\\
        &\dz_{\tt} = f(\z_{\tt},\bv_{\tt}) + E\wb,&&&&\ \tau \in [0,T], \label{eq:rompc_zdot}\\
        &\ds_\tau = -\rho s_\tau + \bwo,&&&&\ \tau \in [0,T], \label{eq:rompc_sdot}\\
        &\gjs(\z_{\tt},\bv_{\tt}) + \cjs s_\tau + \cjso \epsilon \leq 0,&&\ j \in \Ns,&&\ \tau \in [0,T], \label{eq:rompc_sys}\\
        &\gjtto(M\z_{\tt}) + \co (s_\tau + \epsilon) \leq 0,&&\ j \in \No,&&\ \tau \in [0,T], \label{eq:rompc_obs}\\
        &(\z_{\Tt},s_\tau,\epsilon) \in \Xf(\x_{\tpT}\tr), \label{eq:rompc_term}
    \end{alignat}
\end{subequations}
with estimated state $\hx_t$, stage cost $\Js(\z,\bv,\r)=\norm{\z-\xr}_Q^2+\norm{\bv-\ur}_R^2, Q \succ 0, R \succ 0$ and terminal cost $\Jf(\z,\xr)=\norm{\z-\xr}_P^2, P \succ 0$. In this formulation, $Q$ and $R$ are tuning matrices, and $P$ should be suitably computed to show practical asymptotic convergence; see Section~\ref{sec:rompc_proof_tracking}.

Similar to the TMPC formulation in Section~\ref{sec:tmpc}, the ROMPC formulation is applied receding-horizon. In this case, the control law for the closed-loop system is defined as follows for $t \geq 0$:
\begin{equation}\label{eq:rompc_cl}
    \bu_{\tpt} = \kappa(\hx_{\tpt},\z_{\tt}^*,\bv_{\tt}^*),\ \tau \in [0,\Ts],
\end{equation}
with control law $\kappaxzv : \kappadef$ as defined in \eqref{eq:rmpc_cl} that will be detailed later according to the design presented in this section.

\subsection{ROMPC properties}\label{sec:rompc_properties}
The main difference between ROMPC and RMPC is that the actual states $\x$ are hidden, and we can only measure noisy outputs $\y$. A logical choice in the ROMPC formulation is to predict a nominal trajectory starting at some estimated $\hx$ based on $\y$ instead of $\x$. This trajectory will satisfy tightened constraints \eqref{eq:rompc_sys}, \eqref{eq:rompc_obs} and \eqref{eq:rompc_term}. However, tightening these constraints with tube dynamics \eqref{eq:rmpc_sdot} is not sufficient to guarantee satisfaction of the non-tightened constraints for the closed-loop trajectory since $\x$ might be closer to the constraints than $\y$. Therefore, we want to increase the tightening to account for this additional observer error.

To determine the additional required tightening, the error $\y-C\x$ needs to be bounded. This bound is given by the compact set $\H$ described in \eqref{eq:eta_bb}. Furthermore, as further detailed in Section~\ref{sec:rompc_properties_design}, we need to have an explicit expression for the derivative $\frac{d}{dt}(\y-C\x)$ to determine the size of the tightening. Since this information is unavailable, we design an observer to estimate state $\hx$ given output $\y$ with the following dynamics:
\begin{equation}\label{eq:rompc_observer}
    \dhx \coloneqq f(\hx,\bu) + E\wb + L(\y-C\hx),
\end{equation}
with $\bu$ given in \eqref{eq:rompc_cl} and observer gain matrix $L\in\RL$, which serves as a low-pass filter on measurements $\y$.

Similar to the properties described in Section~\ref{sec:rmpc_properties} and given observer \eqref{eq:rompc_observer}, we want to design a feedback law $\kappadhxz$ that renders tube size $s$ of incremental Lyapunov function $\Vdhxz$ invariant. In this case, $\kappadhxz$ and $\Vdhxz$ are based on a geodesic that ends at $\hx$ rather than $\x$. Such a combination of $\kappadhxz$ and $\Vdhxz$ exists if the system satisfies the following assumption:
\begin{assumption}[Incremental stabilizability]\label{ass:rompc_inc_stab}
    There exist a feedback law $\kappadhxz : \kappaddef$, an incremental Lyapunov function $\Vdhxz : \Vddef$ that is continuously differentiable and satisfies $\Vd(\zz)=0, \forall \z \in \X$, parameters $\cdl, \cdu, \rho > 0$, and Lipschitz constants $\cjs > 0, j \in \Ns$ and $\co > 0$, such that the following properties hold for all $(\hxzv)\in\XtZ$, $\w\in\wb\oplus\W$, and $\bmeta\in\H$:
    \begin{subequations}\label{eq:rompc_ass_inc_stab}
        \begin{alignat}{2}
            &\cdl \norm{\hx-\z}^2 \leq \Vdhxz \leq \cdu \norm{\hx-\z}^2,&&\label{eq:rompc_ass_inc_stab_vd_bounds}\\
            &\gjs(\hx,\kappahxzv) - \gjs(\zv) \leq \cjs \sqrt{\Vdhxz},\ &&j \in \Ns,\label{eq:rompc_ass_inc_stab_cjs}\\
            &\go(M\hx) - \go(M\z) \leq \co \sqrt{\Vdhxz},&&\label{eq:rompc_ass_inc_stab_co}\\
            &\frac{d}{dt}\Vdhxz \leq -2\rho \Vdhxz,&&\label{eq:rompc_ass_inc_stab_vd_contract}
        \end{alignat}
    \end{subequations}
    with $\dhx = f(\hx,\kappahxzv) + E\wb + L(\y-C\hx)$, $\dz = f(\zv) + E\wb$, $\y$ defined in \eqref{eq:y_eta}, and $\dx = f(\x,\kappahxzv) + E\wb$. Furthermore, the following norm-like inequality holds $\forall \x_1, \x_2, \x_3 \in \Rx$:
    \begin{equation}\label{eq:rompc_ass_inc_stab_triangle_ineq}
        \sqrt{\Vd(\x_1,\x_3)} \leq \sqrt{\Vd(\x_1,\x_2)} + \sqrt{\Vd(\x_2,\x_3)}.
    \end{equation}
\end{assumption}
This assumption reflects the same properties as Assumption~\ref{ass:rmpc_inc_stab} in Section~\ref{sec:rmpc_properties} with the only difference that it is defined with $\hx$ instead of $\x$ using observer \eqref{eq:rompc_observer}. Following similar arguments as in Section~\ref{sec:rmpc_properties}, we know that the time-derivative of $\Vdhxz$ can be upper-bounded:
\begin{property}\label{property:w_eta}
    There exists a constant $\bwo > 0$ such that for any $(\y,\hx,\z,\bv) \in \Ry \times \Rx \times \Z$, any $\w \in \wb\oplus\W$, and any $\bmeta \in\H$, it holds that
    \begin{equation}\label{eq:vd_sqrt_w_eta}
        \frac{d}{dt}\sqrt{\Vdhxz} \leq -\rho \sqrt{\Vdhxz} + \bwo,
    \end{equation}
    with $\dhx = f(\hx,\kappahxzv) + E\wb + L(\y-C\hx)$, $\dz = f(\zv) + E\wb$, $\y$ defined in \eqref{eq:y_eta}, and $\dx = f(\x,\kappahxzv) + E\w$.
\end{property}
Note that a different constant $\bwo \neq \bw$ is used in this case. Whereas $\bw$ accounts for the impact of disturbances, $\bwo$ accounts for the impact of both disturbances and measurement noise.

By Property~\ref{property:w_eta}, we know that $\Vdhxz$ is upper-bounded by tube size $s$:
\begin{equation}\label{eq:rompc_tube}
    \sqrt{\Vdhxz} \leq s,
\end{equation}
with the tube dynamics given by
\begin{equation}
    \ds = -\rho s + \bwo,
\end{equation}
which is the same as \eqref{eq:rompc_sdot}. Rewriting as a function of time gives
\begin{equation}\label{eq:rompc_s_t}
    s_t = \left(1-e^{-\rho t}\right)\frac{\bwo}{\rho},
\end{equation}
and taking the limit $t \rightarrow \infty$ results in a maximum tube size of
\begin{equation}\label{eq:rompc_sbar}
    \bs = \frac{\bwo}{\rho}.
\end{equation}
Thus, $s$ does not grow unbounded, and the resulting tube size \eqref{eq:rompc_sbar} is suitable for tightening the constraints.

Note that contraction property \eqref{eq:rompc_ass_inc_stab_vd_contract} and corresponding Property~\ref{property:w_eta} and tube dynamics \eqref{eq:rompc_sdot} are used to upper-bound the controller error
\begin{equation}\label{eq:rompc_bdelta}
    \bdelta \coloneqq \hx-\z.
\end{equation}
Thus, tightening using $s$ results in closed-loop safety guarantees for $\hx$, not for $\x$.

To account for the observer error
\begin{equation}\label{eq:rompc_beps}
    \beps \coloneqq \x-\hx,
\end{equation}
we can make the following assumption:
\begin{assumption}[Bounded observer error]\label{ass:rompc_observer_error}
    There observer gain matrix $L$ is such that the observer error $\x-\hx$ is bounded and described by an $\epsilon$-sublevel set of $\sVdxhx$:
    \begin{equation}\label{eq:vd_ub_eps}
        \sqrt{\Vd(\x_t,\hx_t)} \leq \epsilon,\ \forall t\geq 0.
    \end{equation}
\end{assumption}
This assumption implies that we can additionally tighten the system and obstacle avoidance constraints proportional to $\epsilon$ in order to guarantee closed-loop safety of both $\hx$ and $\x$. Note that the tightening proportional to $\epsilon$ only applies to the state constraints, not the input constraints, since control law \eqref{eq:rompc_cl} is independent of $\x$. Therefore, we introduce additional tightening constants $\cjso,\ j\in\Ns$:
\begin{equation}\label{eq:rompc_cjso}
    \cjso = \begin{cases}
        0, &j\in\N_{[1,2\nuu]}\\
        \cjs, &j\in\N_{[2\nuu+1,\ns]}
    \end{cases}.
\end{equation}

Similar to Section~\ref{sec:rmpc_properties}, we need to assume the existence of the following terminal ingredients:
\begin{assumption}[Terminal ingredients]\label{ass:rompc_term_ing}
    There exist a terminal control law $\kappafhxr : \XtbZ \rightarrow \Ru$ and terminal cost $\Jf(\z,\xr) : \XtbX \rightarrow \R$ such that the following property holds for all $\z \in \X$:
    \begin{equation}\label{eq:rompc_ass_term_jf}
        \Jf(\z_{\TTt}^*,\x_{\tpTsT}\tr) - \Jf(\z_{\Tt}^*,\x_{\tpT}\tr) \leq -\int_{T}^{T+\Ts}\Js(\z_{\tt}^*,\bv_{\tt}^*,\r_{\tpt}) d\tau,
    \end{equation}
    with sampling time $\Ts > 0$, optimal solution $(\z_{\ct}^*,\bv_{\ct}^*)$ at $t$ for all $t \geq 0$ and its extended version $\z_{\Tpt|t}^*$ using terminal control law $\kappafhxr$ for $\tau \in [0,\Ts]$, as defined in Section~\ref{sec:rompc_proof_candidate}.
\end{assumption}
Similar to Section~\ref{sec:rmpc_properties_design}, next to this assumption on terminal control law $\kappafhxr$ and terminal cost $\Jf(\z,\xr)$, we want to design terminal set $\Xf(\xr)$ such that tightened system and obstacle avoidance constraints \eqref{eq:rompc_sys} and \eqref{eq:rompc_obs} are satisfied in $\Xf(\xr)$, respectively, and such that terminal control law $\kappafhxr$ ensures robust positive invariance of $\Xf(\xr)$. The specific formulation of a tight version of $\Xf(\xr)$ is presented in Section~\ref{sec:rompc_proof_rec_feas} as part of the proof and therefore not included in Assumption~\ref{ass:rompc_term_ing}.

\begin{figure}[h!]
    \centering
    \includesvg[width=\textwidth]{rompc_proofs.svg}
    \caption{Visualization of a 1D reference trajectory $\x_{t+[0,4\Ts]}\tr$, actual state trajectory $\x_{[t,\tTs]}$, previously optimal solution $\z_{[0,T]|t}^*$ starting from estimated state $\hx_t$, appended part of previously optimal solution $\z_{[T,T+\Ts]|t}^*$, candidate solution $\tz_{[0,T]|\tTs}$ starting from estimated state $\hx_{\tTs}$, tubes $\Vd(\hx,\z^*)\leq s$ and $\Vdxhx\leq\epsilon$ around $\z_{[0,T]|t}^*$, tubes $\Vd(\hx,\tz)\leq s$ and $\Vdxhx\leq\epsilon$ around $\tz_{[0,T]|\tTs}$, and terminal sets $\Xf(\xr)$ around the reference trajectory corresponding to the previously optimal and the candidate solution. Again, we set $T=4\Ts$. Note that the Minkowski sum of tubes $s$ and $\epsilon$ is plotted around the nominal predicted trajectory as the combination of these tubes is used to tighten the constraints. Compared to Figure~\ref{fig:rmpc}, the initial tube size is centered around $\hx$, has a non-zero size, and contains state $\x$. Again, $\w$ impacts the system during interval \circled{1}. Thus, the differences $\tz_{\Tst}-\hx_{\tTs}$ and $\x_{\tTs}-\hx_{\tTs}$ are both the result of the combination of $\w$, $\bmeta$, and observer dynamics \eqref{eq:rompc_observer}. Note that the tubes and terminal sets are visualized as ellipses to clarify their shapes, whereas they should be visualized as vertical lines in this case.}
    \label{fig:rompc}
\end{figure}

Again, the assumptions above allow us to prove recursive feasibility and trajectory tracking. Figure~\ref{fig:rompc} provides intuition for the proofs related to ROMPC formulation \eqref{eq:rompc}. Compared to the RMPC case, instead of $\x$, we can only obtain estimated state $\hx$ via noisy output measurements $\y$. However, by suitably designing observer \eqref{eq:rompc_observer}, we know that $\x$ is contained in a tube around $\hx$. Therefore, the system and obstacle avoidance constraints are tightened by an additional term propertional to this tube size. Furthermore, the tube size used to describe the difference between $\hx$ and $\z$ starts from zero at the beginning of the prediction horizon and is strictly larger than the tube size designed in Section~\ref{sec:rmpc_properties} for both the previously optimal and the candidate solution. Given a suitably designed control law and incremental Lyapunov function, the candidate contracts with at least factor $\rho$ towards the previously optimal solution. Hence, with the adjusted tube sizes above, we can apply a similar reasoning as in Section~\ref{sec:rmpc_proof_rec_feas} to prove recursive feasibility.

Since the feedback law is based on $\hx$ rather than $\x$, the candidate might contract slower to the previously optimal solution compared to the case where $\x$ is used to compute the feedback law. This worst-case effect results in a larger tracking error between $\x$ and the reference trajectory. Since we know that $\x-\hx$ is bounded, we can still conclude that, on average, the tracking error gets small. However, the bound is less strict than the one obtained in Section~\ref{sec:rmpc_proof_tracking}.

\subsubsection{Candidate solution}\label{sec:rompc_proof_candidate}
Similar to Section~\ref{sec:rmpc_proof_candidate}, the goal of defining the candidate solution is to show that there exists a feasible, not necessarily optimal, solution to \eqref{eq:rompc} at time $\tTs$ given that \eqref{eq:rompc} is successfully solved at time $t$. Again, this candidate solution is used to prove trajectory tracking and recursive feasibility, but it is not necessarily implemented in practice.

In the robust output-feedback case, we pick a candidate solution similar to the nominal case. For convenience, we define for $\tau \in [T,T+\Ts]$:
\begin{equation}\label{eq:rompc_u_append}
    \bv_{\tt}^* = \kappaf(\z_{\tt}^*,\r_{\tpt}),
\end{equation}
with $\dz_{\tt}^*$ given by \eqref{eq:xdot_nom} by applying \eqref{eq:rompc_u_append}.

Given these definitions, we can write the candidate for prediction interval $\tau \in [0,T]$ at time $\tTs$ as
\begin{subequations}\label{eq:rompc_candidate}
    \begin{align}
        \tz_{0|\tTs} &= \hx_{\tTs},\label{eq:rompc_candidate_z0}\\
        \ts_0 &= 0,\label{eq:rompc_candidate_s0}\\
        \tv_{\ttT} &= \kappa(\tz_{\ttT},\z_{\tTt}^*,\bv_{\tTt}^*),\label{eq:rompc_candidate_v}
    \end{align}
\end{subequations}
with $\dtz_{\ttT}$ and $\dts_\tau$ according to \eqref{eq:rompc_zdot} and \eqref{eq:rompc_sdot}, respectively.

Note that this candidate deviates from the one in Section~\ref{sec:rmpc_proof_candidate}. The main difference is the initial estimated state. Consequently, feedback control law \eqref{eq:rompc_candidate_v} gives different values compared to \eqref{eq:rmpc_candidate_v} in Section~\ref{sec:rmpc_proof_candidate}.

\subsubsection{Recursive feasibility proof}\label{sec:rompc_proof_rec_feas}
The recursive feasibility proof follows similar arguments as in Section~\ref{sec:rmpc_proof_rec_feas} with additional technicalities to account for measurement noise $\bmeta$. Instead of optimizing from the measured state $\x$, the optimization starts from the estimated state $\hx$, and we ensure that the controller error $\bdelta$ is contained inside $\delta$-sublevel set of $\Vdhxz$ and the observer error $\beps$ inside $\epsilon$-sublevel set of $\Vdxhx$.

To prove recursive feasibility, first note that the initial state, initial tube, system dynamics, and tube dynamics constraints are all trivially satisfied by \eqref{eq:rompc_candidate}. Again, we have to show that the system constraints \eqref{eq:rompc_sys}, the obstacle avoidance constraints \eqref{eq:rompc_obs}, and the terminal set constraint \eqref{eq:rompc_term} are satisfied for $\tau\in[0,T]$.

Similar to the procedure in Section~\ref{sec:rmpc_proof_rec_feas}, we split the interval into two subintervals to show satisfaction of \eqref{eq:rompc_sys} and \eqref{eq:rompc_obs}: $\tau\in[0,T-\Ts)$ and $\tau\in[T-\Ts,T]$. By replacing $\bw$ by $\bwo$ and additionally tightening the state constraints with a factor proportional to $\epsilon$, we obtain trivial satisfaction of the system and obstacle avoidance constraints for $\tau\in[0,T-\Ts)$ following the same procedure as in Section~\ref{sec:rmpc_proof_rec_feas}. Note that in that case the additional tightening terms proportional to $\epsilon$ cancel out in \eqref{eq:rmpc_cjs_ub} and the same proof in Appendix~\ref{app:s_bound} holds with a different expression for $s_{\Ts}$:
\begin{equation}\label{eq:rompc_sTs}
    s_{\Ts} = \left(1-e^{-\rho \Ts}\right)\frac{\bwo}{\rho}.
\end{equation}

\noindent\textbf{Terminal set constraint satisfaction for $\bm{\tau\in[T-\Ts,T]}$}\\
Since the tightening is increased proportional to $\epsilon$ compared to Section~\ref{sec:rmpc}, we need to account for this additional tube size in the terminal set. Therefore, the terminal set is now given by
\begin{equation}\label{eq:rompc_xf}
    \Xf(\xr) \coloneqq \left\{\z \in \X, s_T \in \R, \epsilon\in\R\ \middle|\ \sqrt{\Vd(\z,\xr)} + s_T + \epsilon \leq \alpha\right\}.
\end{equation}
To show satisfaction of terminal set constraint \eqref{eq:rompc_term} for $\tau\in[T-\Ts,T]$, we follow similar steps to the ones outlined in Remark~\ref{remark:rmpc_term_inv} in Section~\ref{sec:rmpc_proof_rec_feas}. We start by showing that terminal set constraint \eqref{eq:rompc_xf} is satisfied at $\tau=T-\Ts$ at $t+\Ts$. This gives an inequality similar to \eqref{eq:rmpc_term_inv_der_0} evaluated at $\tau = 0$:
\begin{equation}
    \alpha - s_T - \epsilon + e^{-\rho(T-\Ts)}s_\Ts \leq \alpha - s_{T-\Ts} - \epsilon,
\end{equation}
or, equivalently,
\begin{equation}
    s_{T-\Ts} \leq s_T - e^{-\rho(T-\Ts)}s_\Ts.
\end{equation}
This is the same expression as \eqref{eq:s_bound}, which holds trivially by the proof in Appendix~\ref{app:s_bound} with $s_\Ts$ given by \eqref{eq:rompc_sTs}.

Appendix~\ref{app:rompc_xf_inv} proves that terminal set \eqref{eq:rompc_xf} is invariant provided that $\alpha\geq\frac{\bwo}{\rho}+\epsilon$. Therefore, we can conclude that the terminal set constraint is also satisfied with this lower bound on $\alpha$ for $\tau\in[T-\Ts,T]$ at $\tTs$.\\

\noindent\textbf{System constraints satisfaction for $\bm{\tau\in[T-\Ts,T]}$}\\
Following the same reasoning as detailed in Section~\ref{sec:rmpc_proof_rec_feas} and leveraging the proof in Appendix~\ref{app:s_bound} and the fact that \eqref{eq:rmpc_xf} is invariant, the candidate system constraints are satisfied for $\tau\in[T-\Ts,T]$ under the following assumption:
\begin{assumption}[System constraints tightening for reference trajectory]\label{ass:rompc_gjrs}
    The reference trajectory satisfies the tightened systems constraints, i.e., \eqref{eq:r_sys} holds, with $\gjrs(\hx,\bu)$ given by
    \begin{equation}\label{eq:ass_rompc_gjrs}
        \gjrs(\hx,\bu) = \gjs(\hx,\bu) + \cjs \frac{\bwo}{\rho} + \cjso \epsilon,\ j \in \Ns,
    \end{equation}
    where $\cjs$ and $\cjso$ are defined in \eqref{eq:rompc_ass_inc_stab_cjs} and \eqref{eq:rompc_cjso}, respectively.
\end{assumption}
This is the same assumption as Assumption~\ref{ass:rmpc_gjrs} in Section~\ref{sec:rmpc_proof_rec_feas} with the only difference that we have access to $\hx$ instead of $\x$ and, correspondingly, the lower bound on $\alpha$ is larger.\\

\noindent\textbf{Obstacle avoidance constraints satisfaction for $\bm{\tau\in[T-\Ts,T]}$}\\
Following a similar reasoning as for the system constraints satisfaction for $\tau\in[T-\Ts,T]$, the candidate obstacle avoidance constraints are satisfied for $\tau\in[T-\Ts,T]$ under the following assumption:
\begin{assumption}[Obstacle avoidance constraints tightening for reference trajectory]\label{ass:rompc_gjro}
    The reference trajectory satisfies the tightened obstacle avoidance constraints, i.e., \eqref{eq:r_obs} holds, with $\gjro(\hp)$ given by
    \begin{equation}\label{eq:ass_rompc_gjro}
        \gjro(\hp) = \gjo(\hp) + \co \alpha,\ j \in \No,
    \end{equation}
    where $\co$ is defined in \eqref{eq:rompc_ass_inc_stab_co} and $\alpha \geq \frac{\bwo}{\rho}+\epsilon$.
\end{assumption}
Again, this assumption is the same as Assumption~\ref{ass:rmpc_gjro} in Section~\ref{sec:rmpc_proof_rec_feas} with the only difference that we have availability of $\hx$ instead of $\x$ and, correspondingly, the lower bound on $\alpha$ is larger.
\qed

\subsubsection{Trajectory tracking proof}\label{sec:rompc_proof_tracking}
The trajectory tracking proof follows the procedure detailed in Section~\ref{sec:rmpc_proof_tracking}. The only difference is that we show tracking for a trajectory optimized from $\hx$ instead of $\x$. Correspondingly, we can follow the same proof as outlined in Section~\ref{sec:rmpc_proof_tracking} with the exceptions that we now leverage \eqref{eq:rompc_ass_term_jf} in Assumption~\ref{ass:rompc_term_ing} instead of \eqref{eq:rmpc_ass_term_jf} in Assumption~\ref{ass:rmpc_term_ing}. The upper bound on the difference in stage costs is now given by
\begin{align}
    &\Js(\tz_{\ttT},\tv_{\ttT},\r_{\tpTst}) - \Js(\z_{\tTt}^*,\bv_{\tTt}^*,\r_{\tpTst})\notag\\
    &\leq \Ljsxu e^{-\rho \tau} \left(1-e^{\rho \Ts}\right)\frac{\bwo}{\rho} \eqqcolon \alpha_\tau\tops(\rho,\bwo,\Ts),
\end{align}
instead of \eqref{eq:rmpc_js_ub_4}, and that the upper bound on the difference in terminal costs is given by
\begin{align}
    &\Jf(\tz_{\TtT},\x_{\tpTsT}\tr) - \Jf(\z_{\TTt}^*,\x_{\tpTsT}\tr)\notag\\
    &\leq \Ljf e^{-\rho T} \left(1-e^{\rho \Ts}\right)\frac{\bwo}{\rho} \eqqcolon \alpha\topf(\rho,\bwo,\Ts),
\end{align}
instead of \eqref{eq:rmpc_jf_ub_2}. Following similar steps as in Section~\ref{sec:rmpc_proof_tracking} with different expressions for $\alpha_\tau\tops(\rho,\bwo,\Ts)$, $\alpha\topf(\rho,\bwo,\Ts)$, and $\balphasf(\rho,\bwo,\Ts)$ changes \eqref{eq:rmpc_track_ub_norm} into the following upper bound:
\begin{subequations}
    \begin{align}
        \norm{\x_\tau-\x_\tau\tr}_Q^2 &\overset{\text{(a)}}{\leq} c^\mathrm{t,1} \norm{\x_\tau-\hx_\tau} + c^\mathrm{t,2} \norm{\hx_\tau-\z_{\tau-t|t}^*} + \norm{\z_{\tau-t|t}^*-\x_\tau\tr}_Q^2\\
        &\overset{\text{(b)}}{\leq} c^\mathrm{t,3} \epsilon + c^\mathrm{t,4} s_\tau + \norm{\z_{\tau-t|t}^*-\x_\tau\tr}_Q^2,
    \end{align}
\end{subequations}
where (a) is obtained by Lipschitz bounds with constants $c^\mathrm{t,1}$ and $c^\mathrm{t,2}$ and (b) by other Lipschitz bounds with constants $c^\mathrm{t,3}$ and $c^\mathrm{t,4}$ on the norms using a combination of \eqref{eq:ass_inc_stab_vd_bounds}, \eqref{eq:vd_ub_eps}, \eqref{eq:rompc_tube}, and \eqref{eq:rompc_s_t}. Therefore, we can write a similar upper bound as in \eqref{eq:rmpc_track_ub_int_2}:
\begin{subequations}
    \begin{align}
        c^{\J,\mathrm{d}} \int_{t}^{\tTs} \norm{\x_\tau-\x_\tau\tr}_Q^2 d\tau \leq &c^{\J,\mathrm{d}} \int_{t}^{\tTs} \norm{\z_{\tau-t|t}^*-\x_\tau\tr}_Q^2 + c^\mathrm{t,4} s_\tau + c^\mathrm{t,3} \epsilon d\tau\\
        \overset{\text{(a)}}{=} &c^{\J,\mathrm{d}} \int_{t}^{\tTs} \norm{\z_{\tau-t|t}^*-\x_\tau\tr}_Q^2 d\tau + c^{\J,\mathrm{d}} \int_{t}^{\tTs} c^\mathrm{t,4} s_\tau d\tau \notag\\
        &+ c^{\J,\mathrm{d}} \int_{t}^{\tTs} c^\mathrm{t,3} \epsilon d\tau\\
        \overset{\text{(b)}}{\leq} &\J_t^* - \J_{\tTs}^* + \balphasfone(\rho,\bwo,\Ts),
    \end{align}
\end{subequations}
with $\balphasfone(\rho,\bwo,\Ts) \coloneqq \balphasf(\rho,\bwo,\Ts) + c^{\J,\mathrm{d}} c^\mathrm{t,4} \left(\Ts-\frac{1}{\rho}\left(1-e^{-\rho\Ts}\right)\right)\frac{\bwo}{\rho} + c^{\J,\mathrm{d}} \Ts c^\mathrm{t,3} \epsilon$, and where (a) is obtained by writing out the integral expressions and (b) by \eqref{eq:rmpc_track_ub_int}. Following the same steps as in Section~\ref{sec:rmpc_proof_tracking}, we obtain upper bound
\begin{align}
    \underset{t\rightarrow\infty}{\oplim}\frac{\int_{0}^{t} \norm{\x_\tau-\x_\tau\tr}_Q^2d\tau}{t} \leq &\frac{\J_0^* - \underset{t\rightarrow\infty}{\oplim}\J_t^*}{c^{\J,\mathrm{d}} t} + \frac{1}{c^{\J,\mathrm{d}}\Ts}\balphasfone(\rho,\bwo,\Ts),
\end{align}
By applying Barbalat's lemma \cite{barbalat1959systemes} on $\frac{\int_{0}^{2\Ts} \norm{\x_\tau-\x_\tau\tr}_Q^2d\tau}{t}$ we conclude that, on average, the tracking error $\norm{\x_t-\x_t\tr}$ gets small over time, in this case proportional to $\balphasfone(\rho,\bwo,\Ts)$.
\qed

\subsubsection{Design to satisfy the assumptions}\label{sec:rompc_properties_design}
This section will detail the design of:
\begin{itemize}
    \item the obstacle avoidance constraints, which need to satisfy Assumption~\ref{ass:obs};
    \item incremental Lyapunov function $\Vdhxz$ and corresponding Lipschitz constants $\cjs, \cjso, j \in \Ns$ and $\co$, which need to satisfy Assumptions~\ref{ass:rompc_inc_stab} and~\ref{ass:rompc_observer_error};
    \item the terminal ingredients, which need to satisfy Assumption~\ref{ass:rompc_term_ing};
    \item the reference trajectory, which needs to satisfy Assumptions~\ref{ass:rompc_gjrs} and~\ref{ass:rompc_gjro}.
\end{itemize}

\noindent\textbf{Obstacle avoidance constraints design}\\
To prove recursive feasibility, we need the obstacle avoidance constraint to satisfy Assumption~\ref{ass:obs}. We can achieve this by following the same design as written in Section~\ref{sec:tmpc_properties_design}.\\

\noindent\textbf{Incremental stabilizability design}\\
Following similar arguments as in Section~\ref{sec:rmpc_properties_design}, the goal of the incremental stabilizability design is to find constant metric $\Pd$ and state-dependent feedback $\Kdgs$ such that we can construct an upper bound on the incremental Lyapunov function. Since the tube dynamics in this case are defined for $\hx$ instead of $\x$, we would like to find the upper bound on $\Vdhxz$ instead of $\Vdxz$ in the following form:
\begin{equation}\label{eq:rompc_dVd_ub}
    \frac{d}{dt}\Vdhxz \leq 2\sVdhxz\left(-\rho \sVdhxz + \bwo\right) = -2\rho \Vdhxz + 2\sVdhxz\bwo,
\end{equation}
with incremental Lyapunov function $\Vdhxz$ defined as
\begin{equation}\label{eq:rompc_vd}
    \Vdhxz \coloneqq (\hx-\z)^\top \Pd (\hx-\z),
\end{equation}
geodesic $\gxs$ defined in \eqref{eq:rmpc_gxs_straight}, and control law $\kappahxzv$ defined as
\begin{equation}\label{eq:rompc_kappa}
    \kappahxzv = \gu(1) = \bv + \int_{0}^{1} \Kdgs(\hx-\z) ds,
\end{equation}
such that \eqref{eq:rompc_sdot} is a valid equation for the tube dynamics. To this end, let's first define the closed-loop observer dynamics:
\begin{align}\label{eq:rompc_dhx}
    \dhx &= f(\hx,\kappa(\hx,\z,\bv)) + E\wb + L(\y-C\hx)\notag\\
    &= f(\hx,\kappa(\hx,\z,\bv)) + E\wb + L(\y-C(\x+\x-\hx))\notag\\
    &= f(\hx,\kappa(\hx,\z,\bv)) + E\wb + L(F\bmeta+C(\x-\hx)),
\end{align}
with $\bmeta$ defined in \eqref{eq:y_eta}.

Using the following shorthand notations and leveraging \eqref{eq:rompc_dhx}:
\begin{align}
    \dbdelta \coloneqq &\dhx-\dz\\
    = &f(\hx,\kappa(\hx,\z,\bv))+E\wb+L(F\bmeta+C(\x-\hx))-\left(f(\z,\bv)-E\wb\right)\notag\\
    = &f(\hx,\kappa(\hx,\z,\bv))-f(\z,\bv)+L(F\bmeta+C(\x-\hx)),\\
    \Acl \coloneqq &\Ag+\Bg\Kdg,
\end{align}
we can upper-bound the time-derivative of $\Vdhxz$ as
\begin{subequations}
    \begin{align}
        \frac{d}{dt}\Vdhxz = &\dbdelta^\top \Pd \bdelta + \bdelta^\top \Pd \dbdelta\\
        = &2\bdelta^\top \Pd \dbdelta\\
        = &2\bdelta^\top \Pd \left(f(\hx,\kappa(\hx,\z,\bv))-f(\z,\bv)+L(F\bmeta+C(\x-\hx))\right)\\
        = &2\bdelta^\top \Pd \left(f(\hx,\kappa(\hx,\z,\bv))-f(\z,\bv)\right) + 2\bdelta^\top \Pd L(F\bmeta+C(\x-\hx))\\
        \overset{\text{(a)}}{\leq} &2\bdelta^\top \Pd \Acl \bdelta + 2\bdelta^\top \Pd L(F\bmeta+C(\x-\hx))\\
        \overset{\text{(b)}}{\leq} &-2\rho \Vdhxz + 2\bdelta^\top \Pd L(F\bmeta+C(\x-\hx))\\
        = &-2\rho \Vdhxz + 2\bdelta^\top \Pd^\frac{1}{2} \Pd^\frac{1}{2} L(F\bmeta+C(\x-\hx))\\
        \leq &-2\rho \Vdhxz + 2\lrnorm{\bdelta}_{\Pd} \lrnorm{L(F\bmeta+C(\x-\hx))}_{\Pd}\\
        = &-2\rho \Vdhxz + 2\sqrt{\Vdhxz}\lrnorm{L(F\bmeta+C(\x-\hx))}_{\Pd}
    \end{align}
\end{subequations}
where (a) is obtained by knowing that there exists an $s \in [0,1]$ such that this expression holds by the MVT and (b) by contraction LMI \eqref{eq:rmpc_sdp_lmi_contract}. Applying the $\opsqrt$ operator to the last expression yields
\begin{equation}
    \frac{d}{dt}\sVdhxz \leq -\rho \sVdhxz + \lrnorm{L(F\bmeta+C(\x-\hx))}_{\Pd}.
\end{equation}
Comparing this equation to \eqref{eq:rompc_sdot} indicates that we need to upper-bound $\lrnorm{L(F\bmeta+C(\x-\hx))}_{\Pd}$ for all $(\hx,\x,\bmeta) \in \XtX \times\H$ to obtain $\bwo$ should upper-bound $\lrnorm{L(F\bmeta+C(\x-\hx))}_{\Pd}$. We design $L$ with bounded eigenvalues and we know that $\bmeta$ is bounded since $\bmeta \in\H$ with compact $\H$. This leaves us to show that $\x-\hx$ is bounded for all $(\x,\hx) \in \XtX$.

A sufficient condition to show boundedness of $\x-\hx$ is to enforce that sublevel set of $\Vdxhx$ \eqref{eq:vd_ub_eps} is RPI. Similar to RPI LMI \eqref{eq:rmpc_sdp_lmi_rpi} this gives the following LMI:
\begin{equation}\label{eq:lmi_rpi_eps}
    \begin{bmatrix}
        \Xd\left(\Aarg-LC\right)^\top + \left(\Aarg-LC\right)\Xd + \lambdaeps\Xd&E\w^0-LF\bmeta\\
        \left(E\w^0-LF\bmeta\right)^\top&-\lambdaeps\epsilon^2
    \end{bmatrix} \preceq 0,
\end{equation}
with multiplier $\lambdaeps \geq 0$ and for $\w^0\in\opvert(\W)$ and $\bmeta \in \opvert(\H)$. The proof of this so-called $\epsilon$-RPI LMI is provided in Appendix~\ref{app:lmi_rpi_eps}.

\begin{remark}
    Since LMI \eqref{eq:lmi_rpi_eps} enforces \eqref{eq:vd_ub_eps} to be RPI, the presented results below are only valid if \eqref{eq:vd_ub_eps} is satisfied at $t=0$, i.e., $\sqrt{\Vd(\x_0,\hx_0)} \leq \epsilon$.
\end{remark}

Given metric $\Pd$ satisfying this LMI, we can write the upper bound on $\lrnorm{L(F\bmeta+C(\x-\hx))}_{\Pd}$ as
\begin{subequations}
    \begin{align}
        \lrnorm{L(F\bmeta+C(\x-\hx))}_{\Pd} &= \lrnorm{\Pd^\frac{1}{2}L\left(F\bmeta+C(\x-\hx)\right)}\\
        &= \lrnorm{\Pd^\frac{1}{2}LF\bmeta+\Pd^\frac{1}{2}LC(\x-\hx)}\\
        &\leq \lrnorm{\Pd^\frac{1}{2}LF\bmeta} + \lrnorm{\Pd^\frac{1}{2}LC(\x-\hx)}\\
        &= \lrnorm{\Pd^\frac{1}{2}LF\bmeta} + \lrnorm{\Pd^\frac{1}{2}LC\Pd^{-\frac{1}{2}}\Pd^\frac{1}{2}(\x-\hx)}\\
        &\leq \lrnorm{\Pd^\frac{1}{2}LF\bmeta} + \lrnorm{\Pd^\frac{1}{2}LC\Pd^{-\frac{1}{2}}} \lrnorm{\Pd^\frac{1}{2}(\x-\hx)}\\
        &= \lrnorm{LF\bmeta}_{\Pd} + \lrnorm{\Pd^\frac{1}{2}LC\Pd^{-\frac{1}{2}}} \lrnorm{\x-\hx}_{\Pd}\\
        &\leq \lrnorm{LF\bmeta}_{\Pd} + \lrnorm{\Pd^\frac{1}{2}LC\Pd^{-\frac{1}{2}}} \epsilon,
    \end{align}
\end{subequations}
with $\bmeta \in\H$. Since $\H$ is a convex polytopic set, evaluating this expression at the vertices of $\H$ suffices. Thus, $\bwo$ is given by
\begin{equation}\label{eq:rompc_bwo}
    \bwo \coloneqq \underset{\bmeta \in \opvert(\H)}{\opmax} \lrnorm{LF\bmeta}_{\Pd} + \lrnorm{\Pd^\frac{1}{2}LC\Pd^{-\frac{1}{2}}} \epsilon.
\end{equation}

LMI \eqref{eq:lmi_rpi_eps} can also be used to optimize the shape of the tube using an RPI LMI for the controller error similar to \eqref{eq:lmi_rpi}. Let's call this LMI the $\delta$-RPI LMI. It is given by
\begin{equation}\label{eq:rompc_lmi_rpi_delta}
    \begin{bmatrix}
        \Aarg\Xd+\Barg\Ydarg+\left(\Aarg\Xd+\Barg\Ydarg\right)^\top+\lambdadelta\Xd&LC\Xd&LF\bmeta\\
        \left(LC\Xd\right)^\top&-\lambdadeltaeps\Xd&\bm{0}\\
        \left(LF\bmeta\right)^\top&\bm{0}&\lambdadeltaeps\epsilon^2-\lambdadelta\delta^2
    \end{bmatrix} \preceq 0,
\end{equation}
with $\bmzeta\in\Z$, $\bmeta\in\opvert(\H)$ and multipliers $\lambdadelta, \lambdadeltaeps \geq 0$. Similar to the design in Section~\ref{sec:rmpc_properties_design}, we can set $\delta=1$ without loss of generality. Note that $\epsilon$ in LMI \eqref{eq:lmi_rpi_eps} does not necessarily equal 1. Instead, it should hold that $\lambdadelta\delta^2\geq\lambdadeltaeps\epsilon^2$ for \eqref{eq:rompc_lmi_rpi_delta} to be satisfied. Since this LMI has a linear dependency on $\bmeta$ and the set $\H$ is convex, it suffices to check \eqref{eq:rompc_lmi_rpi_delta} at the vertices of $\H$. Appendix~\ref{app:rompc_lmi_rpi_delta} provides the proof of \eqref{eq:rompc_lmi_rpi_delta}.

Compared to the RMPC case, we want to solve the same SDP but with replacing RPI LMI \eqref{eq:lmi_rpi} by the combination of $\epsilon$-RPI LMI \eqref{eq:lmi_rpi_eps} and $\delta$-RPI LMI \eqref{eq:rompc_lmi_rpi_delta} and weighting decision variables $\cjs, j\in\N_{[2\nuu+1,\ns]}$ and $\co$ differently in the cost function according to the additional state constraints tightening proportional to $\epsilon$:
\begin{subequations}\label{eq:rompc_sdp}
    \begin{align}
        \underset{\substack{\Xd,\Ydarg\\{\cjs}^2,{\co}^2}}{\opmin}\ \ &(1+\epsilon^2) \cco {\co}^2 + \sum_{j=1}^{2\nuu} \cjcs {\cjs}^2 + (1+\epsilon^2) \sum_{j=2\nuu+1}^{\ns} \cjcs {\cjs}^2,\label{eq:rompc_sdp_obj}\\
        \operatorname{s.t.}\ &\Xd \succeq 0,\label{eq:rompc_sdp_lmi_x}\\
        &\Aarg\Xd+\Barg\Ydarg + \left(\Aarg\Xd+\Barg\Ydarg\right)^\top + 2\rho\Xd \preceq 0,\label{eq:rompc_sdp_lmi_contract}\\
        &\begin{bmatrix}
            \Aarg\Xd+\Barg\Ydarg+\left(\Aarg\Xd+\Barg\Ydarg\right)^\top+\lambdadelta\Xd&LC\Xd&LF\bmeta\\
            \left(LC\Xd\right)^\top&-\lambdadeltaeps\Xd&\bm{0}\\
            \left(LF\bmeta\right)^\top&\bm{0}&\lambdadeltaeps\epsilon^2-\lambdadelta\delta^2
        \end{bmatrix} \preceq 0,\label{eq:rompc_sdp_lmi_rpi_delta}\\
        &\begin{bmatrix}
            \Xd\left(\Aarg-LC\right)^\top + \left(\Aarg-LC\right)\Xd + \lambdaeps\Xd&E\w^0-LF\bmeta\\
            \left(E\w^0-LF\bmeta\right)^\top&-\lambdaeps\epsilon^2
        \end{bmatrix} \preceq 0,\label{eq:rompc_sdp_lmi_rpi_eps}\\
        &\begin{bmatrix}
            {\cjs}^2&\Ljs\begin{bmatrix}\Ydarg\\\Xd\end{bmatrix}\\
            \left(\Ljs\begin{bmatrix}\Ydarg\\\Xd\end{bmatrix}\right)^\top&\Xd
        \end{bmatrix} \succeq 0,\ j \in \Ns,\label{eq:rompc_sdp_lmi_sys}\\
        &\begin{bmatrix}
            {\co}^2 I^{\np}&M\Xd\\
            \left(M\Xd\right)^\top&\Xd
        \end{bmatrix} \succeq 0,\label{eq:rompc_sdp_lmi_obs}\\
        &\bmzeta \in \Z, \quad \w^0\in\opvert\left(\W\right), \quad \bmeta \in \opvert(\H),\notag
    \end{align}
\end{subequations}
contraction rate $\rho > 0$ and multipliers $\lambdadelta,\lambdadeltaeps,\lambdaeps\geq 0$ that can be computed using bi-section over this SDP, user-chosen observer gain matrix $L\in\RL$, and sublevel sets $\delta,\epsilon>0$. Similar to the design in Section~\ref{sec:rmpc_properties_design}, we set $\delta=1$ and we need to make sure $\lambdadelta\delta^2\geq\lambdadeltaeps\epsilon^2$ holds. This SDP is semi-infinite, so we need to evaluate the expressions at grid points and vertices, in this case in $\Z$, $\W$, and $\H$, to account for both the effect of disturbances and measurement noise.

\begin{remark}\label{remark:rompc_sdp_lmi_rpi_split}
    To reduce computation time, we split LMI \eqref{eq:rompc_sdp_lmi_rpi_delta} following the same procedure as outlined in Remark~\ref{remark:rmpc_sdp_lmi_rpi_split}:
    \begin{equation}
        \begin{bmatrix}
            0^{\nx}&0^{\nx}&LF\bmeta\\
            0^{\nx}&0^{\nx}&\bm{0}\\
            \left(LF\bmeta\right)^\top&\bm{0}&0
        \end{bmatrix} \preceq \bH, \quad \bmeta \in \opvert(\H),
    \end{equation}
    \begin{equation}
        \bH+\begin{bmatrix}
            \Aarg\Xd+\Barg\Ydarg+\left(\Aarg\Xd+\Barg\Ydarg\right)^\top+\lambdadelta\Xd&LC\Xd&\bm{0}\\
            \left(LC\Xd\right)^\top&-\lambdadeltaeps\Xd&\bm{0}\\
            \bm{0}&\bm{0}&\lambdadeltaeps\epsilon^2-\lambdadelta\delta^2
        \end{bmatrix} \preceq 0, \quad \bmzeta \in \Z.
    \end{equation}
    Similarly, we split LMI \eqref{eq:rompc_sdp_lmi_rpi_eps} as follows:
    \begin{align}
        \begin{bmatrix}
            0^{\nx}&E\w^0\\
            \left(E\w^0\right)^\top&0
        \end{bmatrix} \preceq \bW, \quad &\w^0\in\opvert(\W),\\
        \begin{bmatrix}
            0^{\nx}&-LF\bmeta\\
            \left(-LF\bmeta\right)^\top&0
        \end{bmatrix} \preceq \bHone, \quad &\bmeta \in \opvert(\H),
    \end{align}
    \begin{equation}
        \bW+\bHone+\begin{bmatrix}
            \Xd\left(\Aarg-LC\right)^\top + \left(\Aarg-LC\right)\Xd + \lambdaeps\Xd&\bm{0}\\
            \bm{0}&-\lambdaeps\epsilon^2
        \end{bmatrix} \preceq 0, \quad \bmzeta \in \Z.
    \end{equation}
\end{remark}

\begin{remark}
    An alternative objective to \eqref{eq:rompc_sdp_obj} is
    \begin{equation}
        \underset{\substack{\Xd,\Ydarg\\{\cjs}^2,{\co}^2,,\epsilon^2}}{\opmin}\ \ \cco {\co}^2 + \sum_{j=1}^{\ns} \cjcs {\cjs}^2 + c^\epsilon \epsilon^2,
    \end{equation}
    in which $\epsilon^2$ is an additional decision variable and $c^\epsilon$ a high penalty term. Minimizing this objective results in a lower bound on $\epsilon$ that renders \eqref{eq:rompc_sdp} feasible. Therefore, it might give less conservative constraint tightening compared to the case where \eqref{eq:rompc_sdp} is optimized with a user-chosen value for $\epsilon$.
\end{remark}

\begin{remark}
    Note that \eqref{eq:rompc_sdp} is bilinear in $\Xd$ and $L$. Hence, a natural solution is to iterate between solving $\Xd$ and $L$. For a given matrix $\Pd=\Xd^{-1}$, we use the following SDP to optimize $L$:
    \begin{subequations}\label{eq:L_sdp}
        \begin{align}
            \underset{\substack{\delta^2,\epsilon^2,l^2,\\\bH,\bHone}}{\opmin}\ \ &\lambdadeltaeps\epsilon^2 + \lambdadelta\delta^2 + c^l l^2,\label{eq:L_sdp_obj}\\
            \operatorname{s.t.}\ &\mathrm{\text{\eqref{eq:rompc_sdp_lmi_rpi_delta}}},\label{eq:L_sdp_lmi_rpi_delta}\\
            &\mathrm{\text{\eqref{eq:rompc_sdp_lmi_rpi_eps}}},\label{eq:L_sdp_lmi_rpi_eps}\\
            &\begin{bmatrix}
                \Pd&\Pd LC\\
                \left(\Pd LC\right)^\top&l^2 \Pd
            \end{bmatrix} \succeq 0,\label{eq:L_sdp_lmi_norm_bound}\\
            &\bmzeta \in \Z, \quad \w^0\in\opvert\left(\W\right), \quad \bmeta \in \opvert(\H),\notag
        \end{align}
    \end{subequations}
    with additional decision variable $l$ such that
    \begin{equation}\label{eq:L_sdp_l_ub}
        \lrnorm{\Pd^\frac{1}{2}LC\Pd^{-\frac{1}{2}}} \leq l,
    \end{equation}
    enforced by LMI \eqref{eq:L_sdp_lmi_norm_bound} as proven in Appendix~\ref{app:L_lmi_norm_bound}, and corresponding penalty term $c^l$. By minimizing $l$, we minimize the upper bound on $\lrnorm{\Pd^\frac{1}{2}LC\Pd^{-\frac{1}{2}}}$, thereby also reducing conservatism in the computation of $\bwo$ given in \eqref{eq:rompc_bwo}.
\end{remark}

In conclusion, by successfully solving \eqref{eq:rompc_sdp}, we can find matrices $\Pd$ and $\Kd(\x)$ such that incremental Lyapunov function $\Vdhxz$ is given by \eqref{eq:rompc_vd} and satisfies \eqref{eq:rompc_ass_inc_stab_vd_bounds} and \eqref{eq:rompc_ass_inc_stab_triangle_ineq}. Furthermore, properties \eqref{eq:rompc_ass_inc_stab_cjs}, \eqref{eq:rompc_ass_inc_stab_co}, and \eqref{eq:rompc_ass_inc_stab_vd_contract} are also satisfied by the design of contraction LMI \eqref{eq:rompc_sdp_lmi_contract}, RPI LMIs \eqref{eq:rompc_sdp_lmi_rpi_delta} and \eqref{eq:rompc_sdp_lmi_rpi_eps}, and Lipschitz LMIs \eqref{eq:rompc_sdp_lmi_sys}, and \eqref{eq:rompc_sdp_lmi_obs}. Therefore, we can conclude that this design satisfies Assumption~\ref{ass:rompc_inc_stab}. Furthermore, Assumption~\ref{ass:rompc_observer_error} is directly satisfied by the design of $\epsilon$-RPI LMI \eqref{eq:lmi_rpi_eps}.\\

\noindent\textbf{Terminal ingredients design}\\
Similar to the terminal ingredients design presented in Section~\ref{sec:rmpc_properties_design}, we need to design a suitable terminal cost $\Jf(\z,\xr)$ such that Assumption~\ref{ass:rompc_term_ing} is satisfied. In this case, we define the terminal control law as
\begin{equation}
    \kappafhxr = \ur + \kappad(\hx,\xr) = \ur + \int_{0}^{1} \Kdgs ds\ (\hx-\xr).
\end{equation}
Furthermore, matrix $P$ needs to be defined such that descent bound \eqref{eq:rompc_ass_term_jf} is satisfied:
\begin{equation}
    \frac{d}{dt}\left((\hx-\z)^\top P (\hx-\z)\right) \leq -\Js(\z,\bv,\r),
\end{equation}
in which $\dhx \coloneqq f(\hx,\kappahxzv) + E\wb + L(\y-C\hx)$ and $\dz = f(\zv) + E\wb$. Again, we find $P$ with the lowest eigenvalues by solving SDP \eqref{eq:rmpc_sdp_jf}.

Thus, we can conclude that Assumption~\ref{ass:rompc_term_ing} is satisfied.\\

\noindent\textbf{Reference trajectory design}\\
Similar to the reference trajectory design in Section~\ref{sec:rmpc_properties_design}, Assumptions~\ref{ass:rompc_gjrs} and~\ref{ass:rompc_gjro} are trivially satisfied if the system and obstacle avoidance constraints used for generating the reference trajectory are constructed according to \eqref{eq:ass_rompc_gjrs} and \eqref{eq:ass_rompc_gjro} using tightening constants $\cjs, j \in \Ns$ and $\co$ from SDP \eqref{eq:rompc_sdp}, respectively, and $\alpha$ satisfying the lower bound computed in \eqref{eq:rompc_term_inv_der}. Correspondingly, terminal set \eqref{eq:rompc_xf} is suitably designed around the reference trajectory to ensure closed-loop system and obstacle avoidance constraints satisfaction at all times.\\

\subsection{Concluding remarks}\label{sec:rompc_conclusion}
In conclusion, the presented ROMPC scheme is an effective tool to provide safety guarantees for tracking dynamically feasible reference trajectories with a mobile robot described by disturbed dynamics \eqref{eq:xdot_w} and noisy output measurements \eqref{eq:y_eta}. The scheme is based on the idea that both the impact of disturbances and measurement noise can be upper-bounded by the combination of an $s$-sublevel set of incremental Lyapunov function $\Vdhxz$ and an $\epsilon$-sublevel set of incremental Lyapunov function $\Vdxhx$. This upper bound starts from $\epsilon$ to account for the observer error and grows using the dynamics of $s$, which results in a larger tube size than the one constructed in Section~\ref{sec:rmpc}. If we use this increased tube size to tighten the system and obstacle avoidance constraints imposed on the trajectory optimization starting from estimated state $\hx$, we can guarantee that the closed-loop system $(\xu)$ always satisfies the actual system and obstacle avoidance constraints. In this work, both Lyapunov functions use the same metric. In general, one could consider using different metrics to quantify the impact of disturbances and measurement noise \cite{kohler2019simple}. Recursive feasibility is proven by constructing a minimum positive invariant terminal set based on $\Vdhxz$ and $\Vdxhx$. This expression is similar to the one found in Section~\ref{sec:rmpc_proof_rec_feas} with the inclusion of the impact of measurement noise. Furthermore, similar to Section~\ref{sec:rmpc_proof_tracking}, trajectory tracking is proven by showing that terminal cost matrix $P$ can be computed using the weighting matrix $\Pd$ used in $\Vdhxz$ and $\Vdxhx$. The upper bound on the tracking error is less strict than the one derived in Section~\ref{sec:rmpc_proof_tracking}, which is caused by the impact of measurement noise $\bmeta$.

\clearpage
{\appendix

\section{Proof of terminal set \eqref{eq:lp_alpha} invariance}\label{app:lp_alpha}
In words, \eqref{eq:lp_alpha} tries to find the largest value for $\alpha$ such that system constraints \eqref{eq:con_sys} are satisfied for all states satisfying terminal set constraint \eqref{eq:tmpc_xf}. Let's define the vector pointing from reference state $\xr$ to state $\x$ by
\begin{equation}\label{eq:tmpc_bdelta}
    \bdelta \coloneqq \x - \xr.
\end{equation}
Terminal set constraint \eqref{eq:tmpc_xf} can now be rewritten as
\begin{equation}\label{eq:xf_bdelta}
    \norm{\bdelta}_{P(\r)}^2 = \bdelta^\top P(\r) \bdelta \leq \alpha^2.
\end{equation}
Furthermore, let's define the error in the control input by
\begin{equation}
    \bu - \ur \overset{\eqref{eq:tmpc_kappaf}}{=} K(\r)(\x-\xr) = K(\r)\bdelta.
\end{equation}
Thus, in the terminal set, it should hold that
\begin{equation}\label{eq:term_set_sys_bdelta}
    \Ljrs \begin{bmatrix}K(\r)\\\Inx\end{bmatrix}\bdelta \leq \ljs - \Ljrs \r,\ j \in \Ns.
\end{equation}

Since the terminal set is described by \eqref{eq:xf_bdelta}, we would like the following implication to hold to satisfy \eqref{eq:tmpc_ass_term_sys} in Assumption~\ref{ass:tmpc_term}:
\begin{eqnarray}
    \bdelta^\top P(\r) \bdelta \leq \alpha^2\ \Rightarrow\ \Ljrs \begin{bmatrix}K(\r)\\\Inx\end{bmatrix}\bdelta \leq \ljs - \Ljrs \r,\ \forall \r \in \bZ,\ j \in \Ns.
\end{eqnarray}
By defining $\tbdelta \coloneqq P(\r)^{\frac{1}{2}} \bdelta$ we obtain
\begin{eqnarray}
    \tbdelta^\top \tbdelta \leq \alpha^2\ \Rightarrow\ \Ljrs \begin{bmatrix}K(\r)\\\Inx\end{bmatrix}P(\r)^{-\frac{1}{2}}\tbdelta \leq \ljs - \Ljrs \r,\ \forall \r \in \bZ,\ j \in \Ns.
\end{eqnarray}
Squaring and taking the norm of the left side of the implication gives
\begin{equation}
    \tbdelta^\top \tbdelta \leq \alpha^2\ \Rightarrow\ \lrnorm{\Ljrs \begin{bmatrix}K(\r)\\\Inx\end{bmatrix}P(\r)^{-\frac{1}{2}}\tbdelta}^2 \leq \lrnorm{\ljs - \Ljrs \r}^2,\ \forall \r \in \bZ,\ j \in \Ns.
\end{equation}
We can now write
\begin{alignat}{2}
    \lrnorm{\Ljrs \begin{bmatrix}K(\r)\\\Inx\end{bmatrix}P(\r)^{-\frac{1}{2}}\tbdelta}^2
    &\overset{\text{(a)}}{\leq} &&\lrnorm{\Ljrs \begin{bmatrix}K(\r)\\\Inx\end{bmatrix}P(\r)^{-\frac{1}{2}}}^2 \lrnorm{\tbdelta}^2\\
    &\overset{\text{(b)}}{=} &&\lrnorm{\Ljrs \begin{bmatrix}K(\r)\\\Inx\end{bmatrix}P(\r)^{-\frac{1}{2}}}^2 \tbdelta^\top \tbdelta\\
    &\overset{\eqref{eq:xf_bdelta}}{\leq} &&\lrnorm{\Ljrs \begin{bmatrix}K(\r)\\\Inx\end{bmatrix}P(\r)^{-\frac{1}{2}}}^2 \alpha^2\\
    &\overset{\text{(c)}}{=} &&\lrnorm{\left(\Ljrs \begin{bmatrix}K(\r)\\\Inx\end{bmatrix}P(\r)^{-\frac{1}{2}}\right)^\top}^2 \alpha^2\\
    &\overset{\text{(d)}}{=} &&\lrnorm{P(\r)^{-\frac{1}{2}} \left[K(\r)^\top \Inx^\top\right] \Ljrs^\top}^2 \alpha^2,
\end{alignat}
where (a) follows from the Cauchy-Schwarz inequality, (b) follows from the definition of the inner product, (c) follows from the fact that the norm of a matrix is equal to the norm of its transpose, and (d) follows from the definition of the transpose and the fact that $P(\r)$ is symmetric.
Thus, the following implication holds:
\begin{equation}
    \lrnorm{P(\r)^{-\frac{1}{2}} \left[K(\r)^\top \Inx^\top\right] \Ljrs^\top}^2 \alpha^2 \leq \left(\ljs - \Ljrs \r\right)^2 \Rightarrow \lrnorm{\Ljrs \begin{bmatrix}K(\r)\\\Inx\end{bmatrix}P(\r)^{-\frac{1}{2}}\tbdelta}^2 \leq \left(\ljs - \Ljrs \r\right)^2.
\end{equation}
Taking the square root of both sides gives
\begin{equation}
    \lrnorm{P(\r)^{-\frac{1}{2}} \left[K(\r)^\top \Inx^\top\right] \Ljrs^\top} \alpha \leq \left(\ljs - \Ljrs \r\right) \Rightarrow \lrnorm{\Ljrs \begin{bmatrix}K(\r)\\\Inx\end{bmatrix}P(\r)^{-\frac{1}{2}}\tbdelta} \leq \left(\ljs - \Ljrs \r\right).
\end{equation}
Therefore, the maximum value for $\alpha$ such that the system constraints are satisfied for all states satisfying terminal set constraint \eqref{eq:tmpc_xf} is given by solving LP \eqref{eq:lp_alpha}.\qed

\section{Proof of tube size upper bound \eqref{eq:s_bound}}\label{app:s_bound}
Intuitively, \eqref{eq:s_bound} says that the upper bound on the candidate tube size starts $s_\Ts$ lower than $s_\Tstau$ and contracts exponentially towards $s_\Tstau$ with factor $e^{-\rho \tau}$ from there. Thus, the upper bound on the candidate tube size contracts with factor $\rho$ to the previously optimal tube size. This makes sense since the tube dynamics \eqref{eq:rmpc_sdot} saturate exponentially to a maximum value of $\frac{\bw}{\rho}$.

Let's prove this by induction. At $\tau = 0$, \eqref{eq:s_bound} is trivially satisfied. Given that \eqref{eq:s_bound} holds at $\tau = 0$, we can prove that it also holds for $\tau>0$:
\begin{subequations}
    \begin{align}
        \frac{\partial}{\partial \tau} \left(s_\tau - s_\Tstau + e^{-\rho \tau} s_\Ts\right) \overset{\eqref{eq:rmpc_sdot}}{=} &-\rho s_\tau + \bw + \rho s_\Tstau - \bw - \rho e^{-\rho \tau} s_\Ts\\
        \overset{\phantom{\eqref{eq:rmpc_sdot}}}{=} &-\rho \left(s_\tau-s_\Tstau+e^{-\rho \tau} s_\Ts\right)\\
        \overset{\eqref{eq:s_bound}}{\geq} &0.
    \end{align}
\end{subequations}
Since the value of $s_\tau - s_\Tstau + e^{-\rho \tau} s_\Ts$ starts negative, its derivative is positive for $s_\tau - s_\Tstau + e^{-\rho \tau} s_\Ts < 0$ and zero for $s_\tau - s_\Tstau + e^{-\rho \tau} s_\Ts = 0$, $s_\tau - s_\Tstau + e^{-\rho \tau} s_\Ts$ exponentially contracts to zero. Therefore, we can conclude that \eqref{eq:s_bound} holds for $\tau \geq 0$.\qed

\section{Proof of terminal set \eqref{eq:rmpc_xf} invariance}\label{app:rmpc_xf_inv}
To prove that terminal set \eqref{eq:rmpc_xf} is invariant for $\tau \geq 0$, we need to show that if the terminal set constraint is satisfied at prediction time $T$ at time $t$, i.e.,
\begin{equation}\label{eq:rmpc_term_t}
    \sqrt{\Vd(\z_{\Tt}^*,\x_{\tpT}\tr)} + s_T \leq \alpha,
\end{equation}
the terminal set constraint should also be satisfied for $\Tpt, \tau \geq 0$ at time $t$, i.e.,
\begin{equation}\label{eq:rmpc_term_tTs}
    \sqrt{\Vd(\z_{\Tpt|t}^*,\x_{\tpTt}\tr)} + s_{\Tpt} \leq \alpha.
\end{equation}

Since there are no disturbances in this part of the prediction horizon, we know that terminal control law $\kappaf(\z_{\Tpt|t}^*,\x_{\tpTt}\tr)$ ensures contraction of $\sqrt{\Vd(\z_{\Tpt|t}^*,\x_{\tpTt}\tr)}$ by at least $\rho$ as given by \eqref{eq:vd_sqrt_t}. Given terminal set satisfaction \eqref{eq:rmpc_term_t}, this gives the following upper bound:
\begin{equation}
    \sqrt{\Vd(\z_{\Tpt|t}^*,\x_{\tpTt}\tr)} \leq e^{-\rho \tau} \left(\alpha - s_T\right).
\end{equation}
If the terminal set is invariant, the right-hand side of this equation is upper-bounded by $\alpha - s_{\Tpt}$ as derived from \eqref{eq:rmpc_term_tTs}:
\begin{equation}
    e^{-\rho \tau} \left(\alpha - s_T\right) \leq \alpha - s_{\Tpt},
\end{equation}
which is trivially satisfied for $\tau = 0$. Working out this inequality for $\tau>0$, similar to \eqref{eq:rmpc_term_inv_der}, gives
\begin{subequations}\label{eq:rmpc_term_inv_der_2}
    \begin{align}
        e^{-\rho \tau} \left(\alpha - s_T\right) &\leq \alpha - s_{\Tpt}\\
        \left(1-e^{-\rho \tau}\right) \alpha &\geq -e^{-\rho \tau} s_T + s_{\Tpt}\\
        \left(1-e^{-\rho \tau}\right) \alpha &\geq -e^{-\rho \tau} \left(1-e^{-\rho T}\right)\frac{\bw}{\rho} + \left(1-e^{-\rho (\Tpt)}\right)\frac{\bw}{\rho}\\
        \left(1-e^{-\rho \tau}\right) \alpha &\geq \left(-e^{-\rho \tau} + e^{-\rho (\Tpt)} + 1 - e^{-\rho (\Tpt)}\right)\frac{\bw}{\rho}\\
        \left(1-e^{-\rho \tau}\right) \alpha &\geq \left(1-e^{-\rho \tau}\right)\frac{\bw}{\rho}\\
        \alpha &\geq \frac{\bw}{\rho}.
    \end{align}
\end{subequations}
Thus, terminal set \eqref{eq:rmpc_xf} is invariant provided that $\alpha \geq \frac{\bw}{\rho}$.
\qed

\section{Proof of RPI LMI \eqref{eq:lmi_rpi}}\label{app:lmi_rpi}
The goal of this appendix is to design an LMI ensuring that sublevel set $\Vdxz \leq \delta^2$ is RPI for disturbed dynamics \eqref{eq:xdot_w} and $\w=\wb+\w^0\in\wb\oplus\W$, with $\W$ given by \eqref{eq:w_bb}. Given state-independent $\Pd$, $\Vdxz$ is given by \eqref{eq:rmpc_vd}. Consequently, we want the following property to hold:
\begin{equation}
    \dVdxz \leq 0, \quad \forall\bdelta^\top \Pd \bdelta \geq \delta^2, \quad \forall\w^0 \in \W,
\end{equation}
with $\bdelta$ defined in \eqref{eq:rmpc_bdelta}.

Using the state-independency of $\Pd$, we can remove the integral and simplify the upper bound on $\dVdxz$ in \eqref{eq:ddt_vdarg_1} as follows:
\begin{equation}
    \bdelta^\top \left(\Acltop\Pd + \Pd\Acl\right) \bdelta + \bdelta^\top \Pd E \w^0 + \left(E\w^0\right)^\top \Pd \bdelta \leq 0,
\end{equation}
or, alternatively,
\begin{equation}
    \begin{bmatrix}
        \bdelta\\
        1
    \end{bmatrix}^\top
    \begin{bmatrix}
        \Acltop\Pd + \Pd\Acl&\Pd E \w^0\\
        \left(E\w^0\right)^\top \Pd&0
    \end{bmatrix}
    \begin{bmatrix}
        \bdelta\\
        1
    \end{bmatrix} \leq 0.
\end{equation}
We want this inequality to hold for all $\bdelta$ such that $\bdelta^\top \Pd \bdelta \geq \delta^2$ and all $\w^0 \in \W$. Since $\W$ is a polytopic set and $\w^0$ appear linearly in the above inequality, it suffices to check the inequality for the vertices of $\W$, i.e., the inequality should hold for all $\w^0 \in \opvert(\W)$.

To enforce these properties, we can use the S-procedure, which states that $\lambda F_1 - F_2 \succeq 0, \lambda \geq 0$ is a sufficient condition for implication $\x^\top F_1 \x \leq 0 \implies \x^\top F_2 \x \leq 0$ to hold \cite{boyd2004convex}. In this case, we leverage the S-procedure to enforce that the following implication holds:
\begin{equation}
    \bdelta^\top \Pd \bdelta \geq \delta^2 \implies
    \begin{bmatrix}
        \bdelta\\
        1
    \end{bmatrix}^\top
    \begin{bmatrix}
        \Acltop\Pd + \Pd\Acl&\Pd E \w^0\\
        \left(E\w^0\right)^\top \Pd&0
    \end{bmatrix}
    \begin{bmatrix}
        \bdelta\\
        1
    \end{bmatrix} \leq 0.
\end{equation}
Re-ordering the terms on the left side of this implication gives
\begin{equation}
    \delta^2 - \bdelta^\top \Pd \bdelta \leq 0,
\end{equation}
which can be written in matrix form as
\begin{equation}
    \begin{bmatrix}
        \bdelta\\
        1
    \end{bmatrix}^\top
    \begin{bmatrix}
        -\Pd&\bm{0}\\
        \bm{0}&\delta^2
    \end{bmatrix}
    \begin{bmatrix}
        \bdelta\\
        1
    \end{bmatrix} \leq 0.
\end{equation}
Now, we can apply the S-procedure with
\begin{subequations}
    \begin{align}
        \x &\coloneqq \begin{bmatrix}\bdelta\\1\end{bmatrix},\\
        F_1 &\coloneqq \begin{bmatrix}
            -\Pd&\bm{0}\\
            \bm{0}&\delta^2
        \end{bmatrix},\\
        F_2 &\coloneqq \begin{bmatrix}
            \Acltop\Pd + \Pd\Acl&\Pd E \w^0\\
            \left(E\w^0\right)^\top \Pd&0
        \end{bmatrix},
    \end{align}
\end{subequations}
to obtain
\begin{equation}
    \lambda \begin{bmatrix}
        -\Pd&\bm{0}\\
        \bm{0}&\delta^2
    \end{bmatrix}-
    \begin{bmatrix}
        \Acltop\Pd + \Pd\Acl&\Pd E \w^0\\
        \left(E\w^0\right)^\top \Pd&0
    \end{bmatrix} \succeq 0.
\end{equation}
Working out this LMI gives
\begin{subequations}
    \begin{align}
        -\begin{bmatrix}
            \Acltop\Pd + \Pd\Acl + \lambda\Pd&\Pd E \w^0\\
            \left(E\w^0\right)^\top \Pd&-\lambda\delta^2
        \end{bmatrix} &\succeq 0,\\
        \begin{bmatrix}
            \Acltop\Pd + \Pd\Acl + \lambda\Pd&\Pd E \w^0\\
            \left(E\w^0\right)^\top \Pd&-\lambda\delta^2
        \end{bmatrix} &\preceq 0,
    \end{align}
\end{subequations}
with $\lambda \geq 0$. Pre- and post-multiplying with $\begin{bmatrix}\Pd^{-1}&\bm{0}\\\bm{0}&1\end{bmatrix}$ yields
\begin{subequations}
    \begin{align}
        \begin{bmatrix}
            \Pd^{-1}\left(\Acltop\Pd + \Pd\Acl + \lambda\Pd\right)\Pd^{-1}&\Pd^{-1}\left(\Pd E \w^0\right)\\
            \left(\left(E\w^0\right)^\top \Pd\right)\Pd^{-1}&-\lambda\delta^2
        \end{bmatrix} &\preceq 0,\\
        \begin{bmatrix}
            \Pd^{-1}\Acltop + \Acl\Pd^{-1} + \lambda\Pd^{-1}&E \w^0\\
            \left(E\w^0\right)^\top&-\lambda\delta^2
        \end{bmatrix} &\preceq 0.
    \end{align}
\end{subequations}
Filling in the definition of $\Acl$ gives
\begin{subequations}
    \begin{align}
        \begin{bmatrix}
            \Pd^{-1}\left(\Aarg+\Barg\Kdarg\right)^\top + \left(\Aarg+\Barg\Kdarg\right)\Pd^{-1} + \lambda\Pd^{-1}&E \w^0\\
            \left(E\w^0\right)^\top&-\lambda\delta^2
        \end{bmatrix} &\preceq 0,\\
        \begin{bmatrix}
            \Pd^{-1}\Aarg^\top+\Pd^{-1}\Kdarg^\top\Barg^\top + \Aarg\Pd^{-1}+\Barg\Kdarg\Pd^{-1} + \lambda\Pd^{-1}&E \w^0\\
            \left(E\w^0\right)^\top&-\lambda\delta^2
        \end{bmatrix} &\preceq 0,
    \end{align}
\end{subequations}
which, under the same coordinate transformation as in Section~\ref{sec:rmpc_properties_design} with state-independent $\Xd=\Pd^{-1}$ and state-dependent $\Yd(\x)=\Kd(\x)\Pd$, and leveraging the symmetry of $\Pd$, gives the following LMI:
\begin{subequations}
    \begin{align}
        \begin{bmatrix}
            \Xd\Aarg^\top+\Ydarg^\top\Barg^\top + \Aarg\Xd+\Barg\Ydarg + \lambda\Xd&E \w^0\\
            \left(E\w^0\right)^\top&-\lambda\delta^2
        \end{bmatrix} &\preceq 0,\\
        \begin{bmatrix}
            \Aarg\Xd+\Barg\Ydarg + \left(\Aarg\Xd+\Barg\Ydarg\right)^\top + \lambda\Xd&E \w^0\\
            \left(E\w^0\right)^\top&-\lambda\delta^2
        \end{bmatrix} &\preceq 0.
    \end{align}
\end{subequations}
\qed

\section{Proof of LMI for Lipschitz continuity obstacle avoidance constraints \eqref{eq:rmpc_sdp_lmi_obs}}\label{app:lmi_obs}
The goal of this appendix is to derive the LMI ensuring Lipschitz continuity of the obstacle avoidance constraints using tightening constant $\co$ and incremental Lyapunov function $\Vdxz$ according to \eqref{eq:ass_inc_stab_co}. Following similar steps as in \cite{nubert2019learning}, we first derive a Lipschitz bound on general constraints $\gj(\xu) \leq 0$, i.e., a bound of the following form for $j \in \Nn$ where $n$ is the number of constraints:
\begin{equation}
    \gj(\xu) - \gj(\z,\bv) \leq \cj \norm{\x-\z}_\Pd.
\end{equation}
The linearization of these constraints with respect to states $\x$ and inputs $\bu$ is given by
\begin{equation}
    \frac{\partial \gj}{\partial \x}\rvert_\x + \frac{\partial \gj}{\partial \bu}\frac{\partial \bu}{\partial \x}\rvert_\x \bdelta,
\end{equation}
with $\bdelta$ defined in \eqref{eq:rmpc_bdelta}. This gives the following Lipschitz bound:
\begin{equation}
    \frac{\partial \gj}{\partial \x}\rvert_\x + \frac{\partial \gj}{\partial \bu}\frac{\partial \bu}{\partial \x}\rvert_\x \leq \cj \norm{\bdelta}_\Pd,
\end{equation}
in which we can rewrite the expression for $\norm{\bdelta}_\Pd$ as
\begin{equation}
    \norm{\x-\z}_\Pd = \sqrt{\bdelta^\top \Pd \bdelta} = \sqrt{\bdelta^\top \Pd^\frac{1}{2} \Pd^\frac{1}{2} \bdelta} = \sqrt{\left(\Pd^\frac{1}{2} \bdelta\right)^\top \left(\Pd^\frac{1}{2} \bdelta\right)} = \lrnorm{\Pd^\frac{1}{2} \bdelta},
\end{equation}
to obtain
\begin{equation}
    \left(\frac{\partial \gj}{\partial \x}\rvert_\x + \frac{\partial \gj}{\partial \bu}\frac{\partial \bu}{\partial \x}\rvert_\x\right) \bdelta \leq \cj \lrnorm{\Pd^\frac{1}{2} \bdelta}.
\end{equation}
Defining $\tbdelta \coloneqq \Pd^\frac{1}{2} \bdelta$ gives the following inequality:
\begin{equation}\label{eq:g_lin_tdelta_ub}
    \left(\frac{\partial \gj}{\partial \x}\rvert_\x + \frac{\partial \gj}{\partial \bu}\frac{\partial \bu}{\partial \x}\rvert_\x\right) \Pd^{-\frac{1}{2}} \tbdelta \leq \cj \lrnorm{\tbdelta}.
\end{equation}
An upper bound on the left-hand side of this equation is given by
\begin{subequations}
    \begin{align}
        \left(\frac{\partial \gj}{\partial \x}\rvert_\x + \frac{\partial \gj}{\partial \bu}\frac{\partial \bu}{\partial \x}\rvert_\x\right) \Pd^{-\frac{1}{2}} \tbdelta
        &\leq \lrnorm{\left(\frac{\partial \gj}{\partial \x}\rvert_\x + \frac{\partial \gj}{\partial \bu}\frac{\partial \bu}{\partial \x}\rvert_\x\right) \Pd^{-\frac{1}{2}} \tbdelta}\\
        &\leq \lrnorm{\left(\frac{\partial \gj}{\partial \x}\rvert_\x + \frac{\partial \gj}{\partial \bu}\frac{\partial \bu}{\partial \x}\rvert_\x\right) \Pd^{-\frac{1}{2}}} \lrnorm{\tbdelta}.
    \end{align}
\end{subequations}
Thus, a sufficient condition for \eqref{eq:g_lin_tdelta_ub} to hold is
\begin{subequations}\label{eq:g_lin_ub}
    \begin{align}
        \lrnorm{\left(\frac{\partial \gj}{\partial \x}\rvert_\x + \frac{\partial \gj}{\partial \bu}\frac{\partial \bu}{\partial \x}\rvert_\x\right) \Pd^{-\frac{1}{2}}} \lrnorm{\tbdelta}
        &\leq \cj \lrnorm{\tbdelta}\\
        \lrnorm{\left(\frac{\partial \gj}{\partial \x}\rvert_\x + \frac{\partial \gj}{\partial \bu}\frac{\partial \bu}{\partial \x}\rvert_\x\right) \Pd^{-\frac{1}{2}}} &\leq \cj.
    \end{align}
\end{subequations}

In the following, we derive the LMI that should hold for the obstacle avoidance constraints specifically. first note that obstacle avoidance constraints \eqref{eq:con_obs} are already linear and do not depend on inputs $\bu$. Therefore, we can write the following equality:
\begin{equation}
    \frac{\partial \go}{\partial \x}\rvert_\x + \frac{\partial \go}{\partial \bu}\frac{\partial \bu}{\partial \x}\rvert_\x = \Lo M.
\end{equation}
Filling this expression in in \eqref{eq:g_lin_ub} gives the following inequality:
\begin{equation}
    \lrnorm{\Lo M \Pd^{-\frac{1}{2}}} \leq \co.
\end{equation}
As commented under \eqref{eq:con_obs}, the polytopic obstacle avoidance constraints can always be normalized such that $\lrnorm{\Lo} = 1$. Therefore, we can write
\begin{equation}
    \lrnorm{\Lo M \Pd^{-\frac{1}{2}}} \leq \lrnorm{\Lo} \lrnorm{M \Pd^{-\frac{1}{2}}} = \lrnorm{M \Pd^{-\frac{1}{2}}}.
\end{equation}
Thus, the following implication holds:
\begin{equation}
    \lrnorm{M \Pd^{-\frac{1}{2}}} \leq \co \Inp \implies \lrnorm{\Lo M \Pd^{-\frac{1}{2}}} \leq \co.
\end{equation}
Working out the left-hand side of this implication and applying coordinate change $\Xd \coloneqq \Pd^{-1}$ yields
\begin{subequations}
    \begin{align}
        \lrnorm{M \Pd^{-\frac{1}{2}}} &\preceq \co \Inp\\
        \lrnorm{M \Pd^{-\frac{1}{2}}}^2 &\preceq \co^2 \Inp\\
        \lrnorm{\left(M \Pd^{-\frac{1}{2}}\right)^\top}^2 &\preceq \co^2 \Inp\\
        \lrnorm{\Pd^{-\frac{1}{2}} M^\top}^2 &\preceq \co^2 \Inp\\
        \left(\Pd^{-\frac{1}{2}} M^\top\right)^\top\left(\Pd^{-\frac{1}{2}} M\right) &\preceq \co^2 \Inp\\
        M \Pd^{-\frac{1}{2}} \Pd^{-\frac{1}{2}} M^\top &\preceq \co^2 \Inp\\
        M \Pd^{-1} M^\top &\preceq \co^2 \Inp\\
        \co^2 \Inp - M \Pd^{-1} M^\top &\succeq 0\\
        \co^2 \Inp - M \Pd^{-1} \Pd \Pd^{-1} M^\top &\succeq 0\\
        \co^2 \Inp - M \Xd \Xd^{-1} \Xd M^\top &\succeq 0.
    \end{align}
\end{subequations}
Taking the Schur complement of this LMI gives
\begin{equation}
    \begin{bmatrix}
        \co^2 \Inp&M\Xd\\
        \left(M\Xd\right)^\top&\Xd
    \end{bmatrix} \succeq 0.
\end{equation}
\qed

\section{Proof of terminal set \eqref{eq:rompc_xf} invariance}\label{app:rompc_xf_inv}
The proof in this appendix follows similar steps as the one in Appendix~\ref{app:rmpc_xf_inv}. To prove that terminal set \eqref{eq:rompc_xf} is invariant for $\tau \geq 0$, we need to show that if the terminal set constraint is satisfied at prediction time $T$ at time $t$, i.e.,
\begin{equation}\label{eq:rompc_term_t}
    \sqrt{\Vd(\z_{\Tt}^*,\x_{\tpT}\tr)} + s_T + \epsilon \leq \alpha,
\end{equation}
the terminal set constraint should also be satisfied for $\Tpt, \tau \geq 0$ at time $t$, i.e.,
\begin{equation}\label{eq:rompc_term_tTs}
    \sqrt{\Vd(\z_{\Tpt|t}^*,\x_{\tpTt}\tr)} + s_{\Tpt} + \epsilon \leq \alpha.
\end{equation}

Since there are no disturbances in this part of the prediction horizon, we know that terminal control law $\kappaf(\z_{\Tpt|t}^*,\x_{\tpTt}\tr)$ ensures contraction of $\sqrt{\Vd(\z_{\Tpt|t}^*,\x_{\tpTt}\tr)}$ by at least $\rho$ as given by \eqref{eq:vd_sqrt_t}. Given terminal set satisfaction \eqref{eq:rompc_term_t}, this gives the following upper bound:
\begin{equation}
    \sqrt{\Vd(\z_{\Tpt|t}^*,\x_{\tpTt}\tr)} \leq e^{-\rho \tau} \left(\alpha - s_T - \epsilon\right).
\end{equation}
If the terminal set is invariant, the right-hand side of this equation is upper-bounded by $\alpha - s_{\Tpt} - \epsilon$ as derived from \eqref{eq:rompc_term_tTs}:
\begin{equation}
    e^{-\rho \tau} \left(\alpha - s_T - \epsilon\right) \leq \alpha - s_{\Tpt} - \epsilon,
\end{equation}
which is trivially satisfied for $\tau = 0$. Working out this inequality for $\tau>0$, similar to \eqref{eq:rmpc_term_inv_der_2}, gives
\begin{subequations}\label{eq:rompc_term_inv_der}
    \begin{align}
        e^{-\rho \tau} \left(\alpha - s_T - \epsilon\right) &\leq \alpha - s_{\Tpt} - \epsilon\\
        \left(1-e^{-\rho \tau}\right) \alpha &\geq -e^{-\rho\tau}s_T + s_{\Tpt} + \left(1-e^{-\rho\tau}\right)\epsilon\\
        \left(1-e^{-\rho \tau}\right) \alpha &\geq -e^{-\rho\tau}\left(1-e^{-\rho T}\right)\frac{\bwo}{\rho} + \left(1-e^{-\rho (\Tpt)}\right)\frac{\bwo}{\rho} + \left(1-e^{-\rho\tau}\right)\epsilon\\
        \left(1-e^{-\rho \tau}\right) \alpha &\geq \left(1-e^{-\rho\tau}+e^{-\rho(\Tpt)}+e^{-\rho(\Tpt)}\right)\frac{\bwo}{\rho} + \left(1-e^{-\rho\tau}\right)\epsilon\\
        \left(1-e^{-\rho \tau}\right) \alpha &\geq \left(1-e^{-\rho\tau}\right)\frac{\bwo}{\rho} + \left(1-e^{-\rho\tau}\right)\epsilon\\
        \alpha &\geq \frac{\bwo}{\rho} + \epsilon.
    \end{align}
\end{subequations}
Thus, terminal set \eqref{eq:rompc_xf} is invariant provided that $\alpha \geq \frac{\bwo}{\rho} + \epsilon$.
\qed

\section{Proof of $\epsilon$-RPI LMI \eqref{eq:lmi_rpi_eps}}\label{app:lmi_rpi_eps}
The goal of this appendix is to design an LMI ensuring that sublevel set $\Vdxhx \leq \epsilon^2$ is RPI for the error between estimated state dynamics $\dhx$, given by \eqref{eq:rompc_dhx} with $\y \in \Ry$ and $\bmeta\in\H$ defined in \eqref{eq:y_eta} and $\H$ defined in \eqref{eq:eta_bb}, and disturbed dynamics $\dx$, given by \eqref{eq:xdot_w} with $\w=\wb+\w^0\in\wb\oplus\W$ and $\W$ defined in \eqref{eq:w_bb}.

Given state-independent $\Pd$, $\Vdxhx$ is given by
\begin{equation}
    \Vdxhx \coloneqq (\x-\hx)^\top \Pd (\x-\hx).
\end{equation}
For the $\epsilon$-sublevel set of $\Vdxhx$ to be RPI, we want the following property to hold:
\begin{equation}\label{eq:lmi_rpi_eps_prop}
    \dVdxhx \leq 0, \quad \forall\beps^\top \Pd \beps \geq \epsilon^2, \quad \forall\w^0\in\W \quad \forall\bmeta\in\H,
\end{equation}
with $\beps$ defined in \eqref{eq:rompc_beps}. Using the following additional shorthand notations $\dbeps$ and $\Acl$:
\begin{align}
    \dbeps &= \dx-\dhx,\notag\\
    &= f(\x,\kappa(\hx,\z,\bv))+E\w-\left(f(\hx,\kappa(\hx,\z,\bv))+E\wb+L(F\bmeta+C(\x-\hx))\right)\notag\\
    &= f(\x,\kappa(\hx,\z,\bv))-f(\hx,\kappa(\hx,\z,\bv))+E\w^0-L(F\bmeta+C\beps),\\
    A &\coloneqq \Ag,
\end{align}
we can upper-bound the time-derivative of $\Vdxhx$ as
\begin{subequations}
    \begin{alignat}{2}
        \dVdxhx &= &&(\dbeps)^\top \Pd \beps + \beps^\top \Pd \dbeps\notag\\
        &= &&2\beps^\top \Pd \dbeps\\
        &= &&2\beps^\top \Pd \left(f(\x,\kappa(\hx,\z,\bv))-f(\hx,\kappa(\hx,\z,\bv))+E\w^0-L(F\bmeta+C\beps)\right)\\
        &= &&2\beps^\top \Pd \left(f(\x,\kappa(\hx,\z,\bv))-f(\hx,\kappa(\hx,\z,\bv))\right) - 2\beps^\top \Pd LC\beps \notag\\
        &&&+2\beps^\top \Pd \left(E\w^0-LF\bmeta\right)\\
        &\overset{\text{(a)}}{\leq} &&2\beps^\top \Pd A\beps - 2\beps^\top \Pd LC\beps + 2\beps^\top \Pd \left(E\w^0-LF\bmeta\right)\\
        &= &&2\beps^\top \Pd\left(A-LC\right)\beps + 2\beps^\top \Pd \left(E\w^0-LF\bmeta\right)\\
        &= &&\beps^\top \left(\left(A-LC\right)^\top \Pd + \Pd\left(A-LC\right)\right)\beps + \beps^\top \Pd \left(E\w^0-LF\bmeta\right) \notag\\
        &&&+\left(E\w^0-LF\bmeta\right)^\top \Pd \beps,
    \end{alignat}
\end{subequations}
where (a) is obtained by knowing that an $s \in [0,1]$ exists, such that this expression holds by the MVT.

Setting the desired upper bound from \eqref{eq:lmi_rpi_eps_prop} gives
\begin{align}
    \dVdxhx \leq &\beps^\top \left(\left(A-LC\right)^\top \Pd + \Pd\left(A-LC\right)\right)\beps + \beps^\top \Pd \left(E\w^0-LF\bmeta\right) + \left(E\w^0-LF\bmeta\right)^\top \Pd \beps\notag\\
    \leq &0,
\end{align}
or, alternatively,
\begin{equation}
    \begin{bmatrix}
        \beps\\
        1
    \end{bmatrix}^\top
    \begin{bmatrix}
        \left(A-LC\right)^\top \Pd + \Pd\left(A-LC\right)&\Pd\left(E\w^0-LF\bmeta\right)\\
        \left(E\w^0-LF\bmeta\right)^\top \Pd&0
    \end{bmatrix}
    \begin{bmatrix}
        \beps\\
        1
    \end{bmatrix} \leq 0.
\end{equation}
It suffices to enforce this inequality for $\w^0\in\opvert(\W)$ and $\bmeta \in \opvert(\H)$ since $\W$ and $\H$ are convex polytopic sets.

Following the S-procedure similar to the derivation in Appendix~\ref{app:lmi_rpi} results in the following LMI:
\begin{equation}
    \begin{bmatrix}
        \left(A-LC\right)^\top \Pd + \Pd\left(A-LC\right) + \lambdaeps\Pd&\Pd\left(E\w^0-LF\bmeta\right)\\
        \left(E\w^0-LF\bmeta\right)^\top \Pd&-\lambdaeps\epsilon^2
    \end{bmatrix} \preceq 0,
\end{equation}
with multiplier $\lambdaeps \geq 0$ and for $\w^0\in\opvert(\W)$ and $\bmeta \in \opvert(\H)$.

From here, following the same steps as presented in Appendix~\ref{app:lmi_rpi} results in the following $\epsilon$-RPI LMI:
\begin{equation}
    \begin{bmatrix}
        \Xd\left(\Aarg-LC\right)^\top + \left(\Aarg-LC\right)\Xd + \lambdaeps\Xd&E\w^0-LF\bmeta\\
        \left(E\w^0-LF\bmeta\right)^\top&-\lambdaeps\epsilon^2
    \end{bmatrix} \preceq 0,
\end{equation}
with $\lambdaeps \geq 0$ and for $\w^0\in\opvert(\W)$ and $\bmeta \in \opvert(\H)$.

Note that this LMI is similar to the RPI LMI derived in Appendix~\ref{app:lmi_rpi}, but with additional terms proportional to $L$ in the top-left block and with $E\w^0-LF\bmeta$ instead of $E\w^0$ in the top-right and bottom-left blocks.
\qed

\section{Proof of $\delta$-RPI LMI \eqref{eq:rompc_lmi_rpi_delta}}\label{app:rompc_lmi_rpi_delta}
The goal of this appendix is to derive the LMI ensuring that sublevel set $\Vdhxz \leq \delta^2$ is RPI for the error between estimated state dynamics $\dhx$, given by \eqref{eq:rompc_dhx} with $\y \in \Ry$ and $\bmeta\in\H$ defined in \eqref{eq:y_eta} and $\H$ defined in \eqref{eq:eta_bb}, and observer error $\beps$, defined in \eqref{eq:rompc_beps}, satisfying $\beps^\top\Pd\beps\leq\epsilon^2$, as enforced by LMI \eqref{eq:lmi_rpi_eps}.

For the $\delta$-sublevel set of $\Vdxhx$ to be RPI, we want the following property to hold:
\begin{equation}\label{eq:rompc_lmi_rpi_delta_prop}
    \dVdhxz\leq 0, \quad \forall\bdelta^\top\Pd\bdelta\geq\delta^2, \quad \forall\beps^\top\Pd\beps\leq\epsilon^2, \quad \forall\bmeta\in\H,
\end{equation}
with $\bdelta$ and $\beps$ defined in \eqref{eq:rompc_bdelta} and \eqref{eq:rompc_beps}, respectively. Using the following shorthand notation for $\Acl$:
\begin{equation}
    \Acl \coloneqq \Aarg+\Barg\Kdarg,
\end{equation}
we can upper-bound the time-derivative of $\Vdhxz$ as
\begin{subequations}
    \begin{alignat}{2}
        \dVdhxz &= &&\dbdelta^\top\Pd\bdelta + \bdelta^\top\Pd\dbdelta\\
        &= &&2\bdelta^\top\Pd\dbdelta\\
        &= &&2\bdelta^\top\Pd\left(f(\hx,\kappahxzv)+E\wb+LF\bmeta+LC\beps-f(\z,\bv)-E\wb\right)\\
        &= &&2\bdelta^\top\Pd\left(f(\hx,\kappahxzv)-f(\z,\bv)\right) + 2\bdelta^\top\Pd(LF\bmeta+LC\beps)\\
        &\overset{\text{(a)}}{\leq} &&2\bdelta^\top\Pd\Acl\bdelta + 2\bdelta^\top\Pd(LF\bmeta+LC\beps)\\
        &= &&\bdelta^\top\Acl^\top\Pd\bdelta + \bdelta^\top\Pd\Acl\bdelta + (LF\bmeta+LC\beps)^\top\Pd\bdelta + \bdelta^\top\Pd(LF\bmeta+LC\beps)\\
        &= &&\bdelta^\top\left(\Acl^\top\Pd + \Pd\Acl\right)\bdelta + (LC\beps)^\top\Pd\bdelta + \bdelta^\top\Pd LC\beps\notag\\
        &&&+ (LF\bmeta)^\top\Pd\bdelta + \bdelta^\top\Pd LF\bmeta,\\
        &= &&\begin{bmatrix}\bdelta\\\beps\\1\end{bmatrix} \begin{bmatrix}\Acl^\top\Pd+\Pd\Acl&\Pd LC&\Pd LF\bmeta\\(LC)^\top\Pd&0^{\nx}&\bm{0}\\\left(LF\bmeta\right)^\top\Pd&\bm{0}&0\end{bmatrix} \begin{bmatrix}\bdelta\\\beps\\1\end{bmatrix},
    \end{alignat}
\end{subequations}
where (a) is obtained by knowing that an $s \in [0,1]$ exists, such that this expression holds by the MVT.

Setting the desired upper bound from \eqref{eq:rompc_lmi_rpi_delta_prop} gives
\begin{equation}
    \dVdhxz \leq \begin{bmatrix}\bdelta\\\beps\\1\end{bmatrix} \begin{bmatrix}\Acl^\top\Pd+\Pd\Acl&\Pd LC&\Pd LF\bmeta\\(LC)^\top\Pd&0^{\nx}&\bm{0}\\\left(LF\bmeta\right)^\top\Pd&\bm{0}&0\end{bmatrix} \begin{bmatrix}\bdelta\\\beps\\1\end{bmatrix} \leq 0.
\end{equation}
It suffices to enforce this inequality for $\bmeta\in\H$ since $\H$ is a convex polytopic set.

Applying the S-procedure to enforce that the inequality holds $\forall\bdelta^\top\Pd\bdelta\geq\delta^2$ with
\begin{subequations}
    \begin{align}
        \x &\coloneqq \begin{bmatrix}\bdelta\\\beps\\1\end{bmatrix},\\
        F_1 &\coloneqq \begin{bmatrix}
            -\Pd&0^{\nx}&\bm{0}\\
            0^{\nx}&0^{\nx}&\bm{0}\\
            \bm{0}&\bm{0}&\delta^2
        \end{bmatrix},\\
        F_2 &\coloneqq \begin{bmatrix}
            \Acl^\top\Pd+\Pd\Acl&\Pd LC&\Pd LF\bmeta\\
            (LC)^\top\Pd&0^{\nx}&\bm{0}\\
            \left(LF\bmeta\right)^\top\Pd&\bm{0}&0
        \end{bmatrix},
    \end{align}
\end{subequations}
results in the following LMI:
\begin{subequations}
    \begin{align}
        \lambdadelta \begin{bmatrix}
            -\Pd&0^{\nx}&\bm{0}\\
            0^{\nx}&0^{\nx}&\bm{0}\\
            \bm{0}&\bm{0}&\delta^2
        \end{bmatrix}-
        \begin{bmatrix}
            \Acl^\top\Pd+\Pd\Acl&\Pd LC&\Pd LF\bmeta\\
            (LC)^\top\Pd&0^{\nx}&\bm{0}\\
            \left(LF\bmeta\right)^\top\Pd&\bm{0}&0
        \end{bmatrix} &\succeq 0\\
        \begin{bmatrix}
            \Acl^\top\Pd+\Pd\Acl&\Pd LC&\Pd LF\bmeta\\
            (LC)^\top\Pd&0^{\nx}&\bm{0}\\
            \left(LF\bmeta\right)^\top\Pd&\bm{0}&0
        \end{bmatrix}-
        \lambdadelta \begin{bmatrix}
            -\Pd&0^{\nx}&\bm{0}\\
            0^{\nx}&0^{\nx}&\bm{0}\\
            \bm{0}&\bm{0}&\delta^2
        \end{bmatrix} &\preceq 0\\
        \begin{bmatrix}
            \Acl^\top\Pd+\Pd\Acl+\lambdadelta\Pd&\Pd LC&\Pd LF\bmeta\\
            (LC)^\top\Pd&0^{\nx}&\bm{0}\\
            \left(LF\bmeta\right)^\top\Pd&\bm{0}&-\lambdadelta\delta^2
        \end{bmatrix} &\preceq 0,
    \end{align}
\end{subequations}
with $\lambdadelta\geq 0$.

We want this LMI to hold $\forall\beps^\top\Pd\beps\leq\epsilon^2$. Applying the S-procedure again, this time with
\begin{subequations}
    \begin{align}
        \x &\coloneqq \begin{bmatrix}\bdelta\\\beps\\1\end{bmatrix},\\
        F_1 &\coloneqq \begin{bmatrix}
            0^{\nx}&0^{\nx}&\bm{0}\\
            0^{\nx}&\Pd&\bm{0}\\
            \bm{0}&\bm{0}&-\epsilon^2
        \end{bmatrix},\\
        F_2 &\coloneqq \begin{bmatrix}
            \Acl^\top\Pd+\Pd\Acl+\lambdadelta\Pd&\Pd LC&\Pd LF\bmeta\\
            (LC)^\top\Pd&0&\bm{0}\\
            \left(LF\bmeta\right)^\top\Pd&\bm{0}&-\lambdadelta\delta^2
        \end{bmatrix},
    \end{align}
\end{subequations}
results in the following LMI:
\begin{subequations}
    \begin{align}
        \lambdadeltaeps \begin{bmatrix}
            0^{\nx}&0^{\nx}&\bm{0}\\
            0^{\nx}&\Pd&\bm{0}\\
            \bm{0}&\bm{0}&-\epsilon^2
        \end{bmatrix}-
        \begin{bmatrix}
            \Acl^\top\Pd+\Pd\Acl+\lambdadelta\Pd&\Pd LC&\Pd LF\bmeta\\
            (LC)^\top\Pd&0^{\nx}&\bm{0}\\
            \left(LF\bmeta\right)^\top\Pd&\bm{0}&-\lambdadelta\delta^2
        \end{bmatrix} &\succeq 0\\
        \begin{bmatrix}
            \Acl^\top\Pd+\Pd\Acl+\lambdadelta\Pd&\Pd LC&\Pd LF\bmeta\\
            (LC)^\top\Pd&0^{\nx}&\bm{0}\\
            \left(LF\bmeta\right)^\top\Pd&\bm{0}&-\lambdadelta\delta^2
        \end{bmatrix}-
        \lambdadeltaeps \begin{bmatrix}
            0^{\nx}&0^{\nx}&\bm{0}\\
            0^{\nx}&\Pd&\bm{0}\\
            \bm{0}&\bm{0}&-\epsilon^2
        \end{bmatrix} &\preceq 0\\
        \begin{bmatrix}
            \Acl^\top\Pd+\Pd\Acl+\lambdadelta\Pd&\Pd LC&\Pd LF\bmeta\\
            (LC)^\top\Pd&-\lambdadeltaeps\Pd&\bm{0}\\
            \left(LF\bmeta\right)^\top\Pd&\bm{0}&\lambdadeltaeps\epsilon^2-\lambdadelta\delta^2
        \end{bmatrix} &\preceq 0,
    \end{align}
\end{subequations}
with $\lambdadelta,\lambdadeltaeps\geq 0$. Pre- and post-multiplying by $\begin{bmatrix}\Pd^{-1}&0^{\nx}&\bm{0}\\0^{\nx}&\Pd^{-1}&\bm{0}\\\bm{0}&\bm{0}&1\end{bmatrix}$ yields
\begin{subequations}
    \begin{align}
        \begin{bmatrix}
            \Pd^{-1}&0^{\nx}&\bm{0}\\
            0^{\nx}&\Pd^{-1}&\bm{0}\\
            \bm{0}&\bm{0}&1
        \end{bmatrix}
        \begin{bmatrix}
            \Acl^\top\Pd+\Pd\Acl+\lambdadelta\Pd&\Pd LC&\Pd LF\bmeta\\
            (LC)^\top\Pd&-\lambdadeltaeps\Pd&\bm{0}\\
            \left(LF\bmeta\right)^\top\Pd&\bm{0}&\lambdadeltaeps\epsilon^2-\lambdadelta\delta^2
        \end{bmatrix}
        \begin{bmatrix}
            \Pd^{-1}&0^{\nx}&\bm{0}\\
            0^{\nx}&\Pd^{-1}&\bm{0}\\
            \bm{0}&\bm{0}&1
        \end{bmatrix} &\preceq 0\\
        \begin{bmatrix}
            \Pd^{-1}\Acl^\top\Pd+\Acl+\lambdadelta&LC&LF\bmeta\\
            \Pd^{-1}(LC)^\top&-\lambdadeltaeps&\bm{0}\\
            \left(LF\bmeta\right)^\top\Pd&\bm{0}&\lambdadeltaeps\epsilon^2-\lambdadelta\delta^2
        \end{bmatrix}
        \begin{bmatrix}
            \Pd^{-1}&0^{\nx}&\bm{0}\\
            0^{\nx}&\Pd^{-1}&\bm{0}\\
            \bm{0}&\bm{0}&1
        \end{bmatrix} &\preceq 0\\
        \begin{bmatrix}
            \Pd^{-1}\Acl^\top+\Acl\Pd^{-1}+\lambdadelta\Pd^{-1}&LC\Pd^{-1}&LF\bmeta\\
            \Pd^{-1}(LC)^\top&-\lambdadeltaeps\Pd^{-1}&\bm{0}\\
            \left(LF\bmeta\right)^\top&\bm{0}&\lambdadeltaeps\epsilon^2-\lambdadelta\delta^2
        \end{bmatrix} &\preceq 0.
    \end{align}
\end{subequations}
Filling in the definition of $\Acl$ gives
\begin{equation}
    \begin{bmatrix}
        \Pd^{-1}\left(\Aarg+\Barg\Kdarg\right)^\top+\left(\Aarg+\Barg\Kdarg\right)\Pd^{-1}+\lambdadelta\Pd^{-1}&LC\Pd^{-1}&LF\bmeta\\
        \Pd^{-1}(LC)^\top&-\lambdadeltaeps\Pd^{-1}&\bm{0}\\
        \left(LF\bmeta\right)^\top&\bm{0}&\lambdadeltaeps\epsilon^2-\lambdadelta\delta^2
    \end{bmatrix} \preceq 0,
\end{equation}
which, under the coordinate change $\Xd\coloneqq\Pd^{-1}, \Ydarg\coloneqq\Kdarg\Xd$ yields the following LMI:
\begin{equation}
    \begin{bmatrix}
        \Aarg\Xd+\Barg\Ydarg+\left(\Aarg\Xd+\Barg\Ydarg\right)^\top+\lambdadelta\Xd&LC\Xd&LF\bmeta\\
        \left(LC\Xd\right)^\top&-\lambdadeltaeps\Xd&\bm{0}\\
        \left(LF\bmeta\right)^\top&\bm{0}&\lambdadeltaeps\epsilon^2-\lambdadelta\delta^2
    \end{bmatrix} \preceq 0.
\end{equation}
\qed

\section{Proof of norm LMI \eqref{eq:L_sdp_lmi_norm_bound}}\label{app:L_lmi_norm_bound}
The goal of this appendix is to show that \eqref{eq:L_sdp_l_ub} is enforced by LMI \eqref{eq:L_sdp_lmi_norm_bound}.

Squaring \eqref{eq:L_sdp_l_ub} yields
\begin{equation}\label{eq:L_sdp_l_ub_sq}
    \lrnorm{\Pd^\frac{1}{2}LC\Pd^{-\frac{1}{2}}}^2 \leq l^2.
\end{equation}

Working out this inequality in matrix form gives
\begin{subequations}
    \begin{align}
        \lrnorm{\Pd^\frac{1}{2}LC\Pd^{-\frac{1}{2}}}^2 &= \left(\Pd^\frac{1}{2}LC\Pd^{-\frac{1}{2}}\right)^\top \left(\Pd^\frac{1}{2}LC\Pd^{-\frac{1}{2}}\right)\\
        &\overset{\text{(a)}}{=} \Pd^{-\frac{1}{2}}(LC)^\top \Pd^\frac{1}{2}\Pd^\frac{1}{2}LC\Pd^{-\frac{1}{2}}\\
        &= \Pd^{-\frac{1}{2}}(LC)^\top\Pd LC\Pd^{-\frac{1}{2}}\\
        &\leq l^2 \Inx,
    \end{align}
\end{subequations}
where (a) holds by symmetry of $\Pd$. Multiplying both sides with $\Pd^\frac{1}{2}$ from left and right yields
\begin{equation}
    (LC)^\top \Pd LC \preceq l^2 \Pd,
\end{equation}
or, equivalently,
\begin{equation}
    l^2 \Pd - (LC)^\top \Pd \Pd^{-1} \Pd LC \succeq 0.
\end{equation}

Taking the Schur complement of this LMI by leveraging $\Pd\succeq0$ gives
\begin{equation}
    \begin{bmatrix}
        \Pd&\Pd LC\\
        \left(\Pd LC\right)^\top&l^2 \Pd
    \end{bmatrix}
    \succeq 0,
\end{equation}
which is equivalent to LMI \eqref{eq:L_sdp_lmi_norm_bound}.
\qed

\clearpage
\bibliographystyle{IEEEtran}
\bibliography{IEEEabrv,references.bib}

\end{document}